\documentclass{article}
\usepackage{ijcai22}

\usepackage{times}
\usepackage{soul}
\usepackage{url}
\usepackage[hidelinks]{hyperref}
\usepackage[utf8]{inputenc}
\usepackage[small]{caption}
\usepackage{graphicx}
\usepackage{amsmath}
\usepackage{amsthm}
\usepackage{booktabs}
\usepackage{algorithm}
\usepackage{algpseudocode}
\usepackage{amssymb}

\usepackage{pifont}

\usepackage{enumitem}

\usepackage{xspace}
\usepackage{misc}

\usepackage{stmaryrd}

\usepackage{thm-restate}

\usepackage{tikz}
\usetikzlibrary{positioning}

\newtheorem{definition}{Definition}
\newtheorem{theorem}{Theorem}
\newtheorem{lemma}{Lemma}
\newtheorem{exmp}{Example}

\title{Frontiers and Exact Learning of \texorpdfstring{\ELI}{ELI} Queries under
  DL-Lite Ontologies}

 \author{Maurice Funk$^1$ \and Jean Christoph Jung$^2$ \And Carsten Lutz$^1$
     \affiliations
     $^1$ Leipzig University, Faculty of Mathematics and Computer
     Science, Germany\\
     $^2$ University of Hildesheim, Institute of Computer Science, Germany
     \emails
     mfunk@informatk.uni-leipzig.de,
     jungj@uni-hildesheim.de,
     clu@informatik.uni-leipzig.de
 }

\begin{document}

\maketitle

\begin{abstract}
  We study \ELI queries (ELIQs) in the presence of ontologies
  formulated in the description logic \DL. For the dialect
  \DLR, we show that ELIQs have a frontier (set of least general
  generalizations) that is of polynomial size and can be computed in
  polynomial time. In the dialect \DLF, in contrast, frontiers may be
  infinite. We identify a natural syntactic restriction that enables 
  the same positive results as for \DLR. We use our results on
  frontiers to show that ELIQs are learnable in polynomial time in the
  presence of a \DLR / $\!\!$restricted \DLF ontology in Angluin's framework
  of exact learning with only membership queries.
\end{abstract}

\section{Introduction}

In the widely studied paradigm of ontology-mediated querying, a
database query is enriched with an ontology that provides domain
knowledge as well as additional vocabulary for query
formulation~\cite{DBLP:conf/pods/BienvenuCLW13,DBLP:conf/rweb/CalvaneseGLLPRR09}. We
consider ontologies formulated in description logics (DLs) of the \DL
family and queries that are \ELI queries (ELIQs) or, in other words,
tree-shaped unary conjunctive queries (CQs). \DL is a prominent choice
for the ontology language as it underpins the OWL 2 QL profile of the
OWL ontology language \cite{owl2-overview}. Likewise, ELIQs are a
prominent choice for the query language as they are computationally
very well-behaved: without an ontology, they can be evaluated in
polynomial time in combined complexity, in contrast to
\NPclass-completeness for unrestricted~CQs. Moreover, in the form of
\ELI concepts they are a central building block of ontologies in
several dialects of \DL and beyond.

The aim of this paper is to study the related topics of computing
least general generalizations (LGGs) of ELIQs under \DL ontologies and
learning ELIQs under \DL ontologies in Angluin's framework of exact
learning
\cite{DBLP:journals/iandc/Angluin87,DBLP:journals/ml/Angluin87}.
Computing generalizations is a natural operation in query engineering
that plays a crucial role in learning logical formulas
\cite{Plotkin,DBLP:journals/ngc/Muggleton91}, in particular in exact learning
\cite{DBLP:conf/icdt/CateD21}. Exact learning, in turn, is concerned
with constructing queries and ontologies. This can be challenging and
costly, especially when logic expertise and domain knowledge are not
in the same hands. Aiming at such cases, exact learning provides a
systematic protocol for query engineering in which a learner interacts
in a game-like fashion with an oracle, which may be a domain expert.

Our results on LGGs concern the notion of a \emph{frontier} of an
ELIQ~$q$ w.r.t.\ an ontology~\Omc. Such a frontier is a set \Fmc of
ELIQs that \emph{generalize} $q$, that is, $q \subseteq_\Omc q_F$ and
$q_F \not\subseteq_\Omc q$ for all $q_F\in \Fmc$, where
`$\subseteq_\Omc$' denotes query containment w.r.t.~\Omc. Moreover,
\Fmc must be \emph{complete} in that for all ELIQs $q'$ with
$q \subseteq_\Omc q'$ and $q' \not\subseteq_\Omc q$, there is a
$q_F \in \Fmc$ such that \mbox{$q_F \subseteq_\Omc q'$}. 
We are interested in computing a frontier that contains only
polynomially many ELIQs of polynomial size, in polynomial time. This
is possible in the case of ELIQs without
ontologies as shown in~\cite{DBLP:conf/icdt/CateD21}; for the simpler \EL queries, the
same had been observed earlier (also without ontologies) in
\cite{DBLP:conf/kr/BaaderKNP18,KriegelPhD}. In contrast, unrestricted
  CQs do not even admit finite frontiers~\cite{DBLP:journals/jct/NesetrilT00}.

In exact learning, the learner and the oracle know and agree on the
ontology~\Omc, and they also agree on the target query $q_T$ to use only
concept and role names from \Omc. 
The learner may ask \emph{membership queries} where they
produce an ABox~\Amc and a candidate answer $a$ and ask whether
$\Amc,\Omc \models q_T(a)$, that is, whether $a$ is an answer to $q_T$
w.r.t.\ \Omc on \Amc. The oracle faithfully answers ``yes'' or
``no''. \emph{Polynomial time learnability} then means that the
learner has an algorithm for constructing $q_T$, up to equivalence
w.r.t.~\Omc, with running time bounded by a polynomial in the
sizes of $q_T$ and~$\Omc$.

Learning with only membership queries, as described above and studied
in this article, is a strong form of exact learning. In fact, there
are not many cases where polynomial time learning with only membership
queries is possible, ELIQs without ontologies being an important
example~\cite{DBLP:conf/icdt/CateD21}. Often, one would therefore also
admit \emph{equivalence queries} where the learner provides a
hypothesis ELIQ~$q_H$ and asks whether $q_H$ is equivalent to $q_T$
under~\Omc; the oracle answers ``yes'' or provides a counterexample,
that is, an ABox \Amc and answer~$a$ such that
$\Amc,\Omc \models q_T(a)$ and $\Amc,\Omc \not\models q_H(a)$ or vice
versa. This is done, for instance, in \cite{KLOW-JMLR18,FJL-IJCAI21}.

We consider as ontology languages the DLs \DLR and \DLF, equipped with
\emph{role inclusions} (also known as role hierarchies) and
\emph{functional roles}, respectively.  Both dialects admit concept
and role disjointness constraints and \ELI concepts on the right-hand
side of concept inclusions~\cite{Romans,DBLP:conf/dlog/KikotKZ11}. We
show that \DLR admits polynomial frontiers that can be computed in
polynomial time, and that \DLF does not even admit finite
frontiers. We then introduce a fragment \DLFsynr of \DLF that
restricts the use of inverse functional roles on the right-hand side
of concept inclusions and show that it is as well-behaved as
\DLR. Both frontier constructions require a rather subtle
analysis. We also note that adding conjunction results in frontiers of
exponential size, even for very simple fragments of \DL.
One application of our results is to show that every ELIQ $q$ can be
characterized up to equivalence w.r.t.\ ontologies formulated in \DLR
or \DLFsynr by only polynomially many data examples of the form
$(\Amc,a)$, labeled as positive if $\Amc,\Omc \models q(a)$ and as
negative otherwise.

We then consider in detail the application of our results in the
context of exact learning and show that ELIQs can be learned in
polynomial time w.r.t.\ ontologies \Omc formulated in \DLR or
\DLFsynr. The learning algorithm uses only membership queries provided
that a seed query is available, that is, an ELIQ $q_0$ such that
$q_0 \subseteq_\Omc q_T$. Such a seed query can be constructed using
membership queries if \Omc contains no concept disjointness
constraints and obtained by a single initial equivalence query
otherwise. We also show that ELIQs cannot be learned at all w.r.t.\
unrestricted \DLF ontologies using only membership queries, and that
they cannot be learned with only polynomially many membership queries
when conjunction is admitted.
%

Proof details are in the appendix.

\smallskip

{\bf Related Work.} Exact learning of queries in the context of
description logics has been studied in \cite{FJL-IJCAI21} while
\cite{KLOW-JMLR18} considers learning entire ontologies, see also
\cite{DBLP:conf/aaai/OzakiPM20,OzakiSurvey}. It is shown in \cite{FJL-IJCAI21}
that a restricted form of CQs (that do not encompass all
ELIQs) can be learned in polynomial time under \EL ontologies using
both membership and equivalence queries. The results from that paper
indicate that inverse roles provide a challenge for exact learning
under ontologies and thus it is remarkable that we can handle them
without any restrictions in our context.
Related
forms of learning are the construction of the least common subsumer
(LCS) and the most specific concept (MSC)
\cite{DBLP:conf/ijcai/Baader03,DBLP:conf/ijcai/BaaderKM99,DBLP:journals/japll/BaaderST07,aaaithis,DBLP:conf/ijcai/ZarriessT13}
which may both be viewed as a form of query generalization.
There is also a more loosely related research thread on learning DL
concepts from labeled data examples
\cite{DBLP:conf/ijcai/FunkJLPW19,DBLP:conf/kr/JungLPW20,DBLP:journals/ml/LehmannH10,DBLP:series/ssw/LehmannV14,DBLP:conf/aaai/SarkerH19}.

\section{Preliminaries}\label{sec:prelims}

\paragraph{\bf Ontologies and ABoxes.}
Let $\NC$, $\NR$, and $\NI$ be countably infinite sets of
\emph{concept}, \emph{role}, and \emph{individual names}.  A
\emph{role} $R$ is a role name $r \in \NR$ or the inverse $r^-$ of a role
name $r$. An \emph{\ELI concept} is formed according to the syntax rule
$C,D ::= \top \mid A \mid C \sqcap D \mid \exists R . C$ where $A$
ranges over concept names and $R$ over roles.  A \emph{basic concept}
$B$ is an \ELI concept of the form $\top$, $A$, or $\exists R .\top$.
When dealing with basic concepts, for brevity we may write $\exists R$
in place of $\exists R. \top$. 

A $\DLRF$ ontology \Omc is a finite set of \emph{concept inclusions
  (CIs)} $B \sqsubseteq C$, \emph{role inclusions (RIs)}
$R_1 \sqsubseteq R_2$, \emph{concept disjointness constraints}
$B_1 \sqcap B_2 \sqsubseteq \bot$, 
\emph{role
  disjointness constraints} $R_1 \sqcap R_2 \sqsubseteq \bot$, and
\emph{functionality assertions} $\mn{func}(R)$.  Here, $B$, $B_1$, and
$B_2$ range over basic concepts, $C$ over \ELI concepts, and
$R_1,R_2,R$ over roles. Superscript $\cdot^\Hmc$ indicates the
presence of role inclusions (also called role hierarchies) and
superscript $\cdot^\Fmc$ indicates functionality assertions, and thus
it should be clear what we mean with a $\DLR$ ontology and with a
$\DLF$ ontology. In fact, we are mainly interested in these two
fragments of \DLRF. 

A $\DLRF$ ontology is in \emph{normal form} if all concept inclusions
in it are of one of the forms $A \sqsubseteq B$, $B \sqsubseteq A$,
and $A \sqsubseteq \exists R . A'$ with $A,A'$
concept names or $\top$ and $B$ a basic concept. Note that
CIs of the form $\exists R \sqsubseteq \exists S$ are not admitted
and neither are CIs of the form $A \sqsubseteq \exists R . C$ with $C$
a compound concept.
%
An ABox \Amc is a finite set of concept assertions $A(a)$ and role
assertions $r(a, b)$ with $A$ a concept name or $\top$, $r$ a role
name, and $a, b$ individual names. We use $\mn{ind}(\Amc)$ to denote
the set of individual names used in \Amc.

As usual, the semantics is given in terms of
\emph{interpretations}~$\Imc$, which we define to be a (possibly
infinite and) non-empty set of concept and role assertions.  We use
$\Delta^\Imc$ to denote the set of individual names in $\Imc$, define
$A^\Imc = \{ a \mid A(a) \in \Imc \}$ for all $A \in \NC$, and
$r^\Imc = \{ (a, b) \mid r(a, b) \in \Imc\}$ and
$(r^-)^\Imc = \{ ( b,a) \mid r(a, b) \in \Imc\}$ for all $r \in
\NR$. 
This definition of interpretation is slightly different from the usual
one, but equivalent;\footnote{This depends on admitting assertions
  $\top(a)$ in ABoxes.} its virtue is uniformity as every ABox is a
finite interpretation. The interpretation function $\cdot^\Imc$ can be
extended from concept names to \ELI concepts in the standard way
\cite{DL-Textbook}. An interpretation $\Imc$ \emph{satisfies} a
concept or role inclusion $\alpha_1 \sqsubseteq \alpha_2$ if
$\alpha_1^\Imc \subseteq \alpha_2^\Imc$, a concept or role
disjointness constraint $\alpha_1 \sqcap \alpha_2 \sqsubseteq \bot$ if
$\alpha_1^\Imc \cap \alpha_2^\Imc = \emptyset$, and a functionality
assertion $\mn{func}(R)$ if $R^\Imc$ is a partial function. It
\emph{satisfies} a concept or role assertion $\alpha$ if $\alpha \in
\Imc$. Note that, as usual, we thus make the standard names
assumption, implying the unique name assumption.

An interpretation is a \emph{model} of an ontology or an ABox if it
satisfies all inclusions, disjointness constraints, and assertions in
it.  We write $\Omc \models \alpha_1 \sqsubseteq \alpha_2$ if every
model of the ontology $\Omc$ satisfies the concept or role inclusion
$\alpha_1 \sqsubseteq \alpha_2$ and $\Omc \models \alpha_1 \equiv
\alpha_2$ if $\Omc \models \alpha_1 \sqsubseteq \alpha_2$ and $\Omc
\models \alpha_2 \sqsubseteq \alpha_1$. If $\alpha_1$ and $\alpha_2$
are basic concepts or roles, then such consequences are decidable in
\PTime both in $\DLR$ and in $\DLF$~\cite{dllite-jair09}.
An ABox \Amc is \emph{satisfiable} w.r.t.\ an ontology $\Omc$ if \Amc
and \Omc have a common model. Deciding ABox satisfiability is also in \PTime in
both $\DLR$ and $\DLF$.

A \emph{signature} is a set of concept and role names, uniformly
referred to as symbols. For any syntactic object $O$ such as an
ontology or an ABox, we use $\mn{sig}(O)$ to denote the symbols used
in $O$ and $||O||$ to denote the \emph{size} of $O$, that is, the
length of a representation of $O$ as a word in a suitable alphabet.

\paragraph{Queries.}

An \ELI concept $C$ can be viewed as an \emph{\ELI
    query (ELIQ)}. An individual $a \in \mn{ind}(\Amc)$ is an
  \emph{answer} to $C$ on an ABox \Amc w.r.t.\ an ontology \Omc, written
  $\Amc,\Omc\models C(a)$, if $a\in C^\Imc$ for all models $\Imc$ of
  $\Amc$ and $\Omc$. We shall often view ELIQs as unary
  \emph{conjunctive queries (CQs)} and also consider CQs that are not
  ELIQs. In this paper, CQs are always unary. A CQ thus takes the form
  $q(x_0)=\exists \bar y \, \phi(x_0, \bar y)$ with $\phi$ a
  conjunction of \emph{concept atoms} $A(x)$ and \emph{role atoms}
  $r(x, y)$ where $A \in \NC$ and $r \in \NR$. We use $\mn{var}(q)$ to
  denote the set of variables that occur in $q$.
  We may view $q$ as a set of atoms 
  and may write $r^-(x, y)$ in place of $r(y, x)$. 
  We call $x_0$ the \emph{answer variable} and use the notion of an
  answer and the notation $\Amc,\Omc \models q(a)$ also for CQs. The
  formal definition is in terms of homomorphisms as usual, details are
  in the appendix.  ELIQs are in 1-to-1 correspondence with CQs whose
  Gaifman graph is a tree and that contain no self-loops and
  multi-edges.  For example, the ELIQ
  $C=A\sqcap \exists r^-.(\exists s.B\sqcap \exists r.A)$ is the CQ
  $q(x_0) = \{A(x),r(y,x),s(y,z),B(z),r(y,z'),A(z')\}$.  We use
  $\Amc_q$ to denote the ABox obtained from CQ $q$ by viewing
  variables as individuals and atoms as assertions.  A CQ $q$ is
  \emph{satisfiable} w.r.t.\ ontology \Omc if $\Amc_q$ is.
  
  For CQs $q_1$ and $q_2$ and an ontology $\Omc$, we say that $q_1$ is
  \emph{contained in} $q_2$ w.r.t.\ $\Omc$, written
  $q_1 \subseteq_\Omc q_2$ if for all ABoxes $\Amc$ and
  $a \in \mn{ind}(\Amc)$, $\Amc, \Omc \models q_1(a)$ implies
  $\Amc, \Omc \models q_2(a)$. If $q_1,q_2$ are ELIQs, then this
  coincides with $q_1$ viewed as \ELI concept being subsumed w.r.t.\
  \Omc by $q_2$ viewed as an \ELI concept \cite{DL-Textbook}.  We call
  $q_1$ and $q_2$ equivalent w.r.t.\ $\Omc$, written
  $q_1 \equiv_\Omc q_2$, if $q_1 \subseteq_\Omc q_2$ and
  $q_2 \subseteq_\Omc q_1$.

\paragraph{\bf \Omc-saturatedness and \Omc-minimality.}

A
CQ $q$ is \emph{$\Omc$-saturated}, with \Omc an ontology, if
$\Amc_q,\Omc \models A(y)$ implies $A(y) \in q$ for all
$y \in \mn{var}(q)$ and $A \in \NC$. It is \emph{\Omc-minimal} if
there is no $x \in \mn{var}(q)$ such that
$q \equiv_\Omc q|_{\mn{var}(q) \setminus \{x\}}$ with $q|_S$ the
restriction of $q$ to the atoms that only contain variables in~$S$.
For a CQ $q$ and an ontology \Omc formulated in \DLR or \DLF, one can
easily find in polynomial time an \Omc-saturated CQ $q'$ with
$q \equiv_\Omc q'$. To achieve \Omc-minimality, we may repeatedly
choose variables $x \in \mn{var}(q)$, check whether
$\Amc_{q|_{\mn{var}(q) \setminus \{x\}}},\Omc \models q$, and if so
replace $q$ with $q|_{\mn{var}(q) \setminus \{x\}}$. For ELIQs, the
required checks can be carried out in \PTime in \DLF
\cite{DBLP:conf/ijcai/BienvenuOSX13}, but are \NPclass-complete in
\DLR \cite{DBLP:conf/dlog/KikotKZ11}. We conjecture that in \DLR, it
is not possible to construct equivalent \Omc-minimal ELIQs in
polynomial time.

\section{Frontiers in \texorpdfstring{\DLR}{DL-LiteH}}
\label{sect:frontier}

We show that for every ELIQ $q$ and \DLR ontology \Omc such that $q$
is satisfiable w.r.t.\ \Omc, there is a frontier of polynomial size
that can be computed in polynomial time. 
We also observe that this fails when \DLR is extended
with conjunction, even in very restricted cases. 
%
\begin{definition}\label{def:frontier} A \emph{frontier} of an ELIQ $q$ w.r.t.\ a \DLRF
  ontology~$\Omc$ is a 
  set of ELIQs $\Fmc$ such that 
    \begin{enumerate} \item $q \subseteq_\Omc q_F$ for all $q_F \in
	\Fmc$; \item $q_F \not\subseteq_\Omc q$ for all $q_F \in
	\Fmc$; \item for all ELIQs $q'$ with $q \subseteq_\Omc q'
	  \not\subseteq_\Omc q$, there is a $q_F \in \Fmc$ with $q_F
	  \subseteq_\Omc q'$.  \end{enumerate} \end{definition}
      It is not hard to see that finite frontiers that
      are minimal w.r.t.\ set inclusion are unique up to equivalence
      of the ELIQs in them, that is, if $\Fmc_1$ and $\Fmc_2$ are
      finite minimal frontiers of $q$ w.r.t.~\Omc, then for every
      $q_F \in \Fmc_1$ there is a $q'_F \in \Fmc_2$ such that
      $q_F \equiv_\Omc q'_F$ and vice versa. The following is the main
      result of this section.
\begin{theorem} \label{thm:frontiermain} Let $\Omc$ be a \DLR ontology
  and $q$ an ELIQ that is \Omc-minimal and satisfiable w.r.t.\ $\Omc$.
  Then a frontier of $q$ w.r.t.\ $\Omc$ can be computed in polynomial
  time.  \end{theorem}
%
We note that Theorem~\ref{thm:frontiermain} still holds when
\Omc-minimality is dropped as a precondition and Condition~2 of
frontiers is dropped as well.
For proving
Theorem~\ref{thm:frontiermain}, we first observe that we can
concentrate on ontologies that are in normal form.
\begin{restatable}{lemma}{lemfrontiernormalformR}
  \label{lem:frontier-normal-form-R} For every \DLR ontology \Omc, we
  can construct in polynomial time a \DLR ontology $\Omc'$ in normal
  form such that every \Omc-minimal ELIQ $q$ is also $\Omc'$-minimal
  and a frontier of $q$ w.r.t.\ $\Omc$ can be constructed in
  polynomial time given a frontier of $q$ w.r.t.\
  $\Omc'$.
\end{restatable}
%
%
We now prove Theorem~\ref{thm:frontiermain}, adapting and
generalizing a technique from
\cite{DBLP:conf/icdt/CateD21}.  Let \Omc and $q(x_0)$ be as in the
formulation of the theorem, with \Omc in normal form.  We may
assume w.l.o.g.\ that $q$ is \Omc-saturated.
To construct a frontier of $q$ w.r.t.~\Omc, we consider all ways to
generalize $q$ in a least general way where `generalizing' means to
construct from $q$ an ELIQ $q'$ such that $q \subseteq_\Omc q'$ and
$q' \not\subseteq_\Omc q$ and `least general way' that
  there is no ELIQ $\widehat q$ that generalizes $q$ and satisfies
  $\widehat q \subseteq_\Omc q'$ and
  $q' \not\subseteq_\Omc \widehat q$. We do this in two steps: the
actual generalization plus a compensation step, the latter being
needed to guarantee that we indeed arrive at a least general
generalization. 

For $x\in \mn{var}(q)$, we use $q_{x}$ to denote the ELIQ obtained from $q$ by
taking the subtree of $q$ rooted at $x$ and making $x$ the answer variable.
The construction that follows involves the
  introduction of fresh variables $x$, some of which are a `copy' of a
  variable from $\mn{var}(q)$. We then use
  $x^\downarrow$ to denote that original variable.

\paragraph{\textbf{Step~1: Generalize.}}
For each variable $x \in \mn{var}(q)$, 
define a set $\Fmc_0(x)$ that contains all ELIQs which can be obtained
by starting with $q_x(x)$ and then doing one of the following:

\smallskip
\noindent
(A)~\emph{Drop concept atom}: 
  \begin{enumerate}
  \item choose an atom $A(x) \in q$ such that 
    \begin{enumerate}
    \item there is no $B(x) \in q$ with $\Omc \models B \sqsubseteq
      A$ and $\Omc \not\models A \sqsubseteq B$ and 
    \item there is no $R(x, y) \in q$ with
      $\Omc \models \exists R \sqsubseteq A$;
    \end{enumerate}
    \item remove all $B(x) \in q$ with $\Omc \models A \equiv B$,
      including $A(x)$.


  \end{enumerate}

\noindent
(B)~\emph{Generalize subquery:}
  \begin{enumerate}
  \item choose an atom $R(x,y) \in q$ directed away from $x_0$;
  \item remove $R(x, y)$ and all atoms of $q_y$;
  

  \item  for each $q'(y) \in
    \Fmc_0(y)$, add a disjoint copy $\widehat q'$ of $q'$ and the role atom
    $R(x,y'')$ with $y''$ the copy of $y$ in $\widehat q'$;


  \item for every role $S$ with $\Omc \models R \sqsubseteq S$ and
    $\Omc \not\models S \sqsubseteq R$, add a disjoint copy $\widehat q_y$ of $q_y$ and the role
    atom $S(x,y')$ with $y'$ the copy of $y$ in $\widehat q_y$.


  \end{enumerate}
  The definition of $x^\downarrow$ should be clear in all cases. In
  Point~3 of Case~(B), for example, for every variable $z$ in $q'$
  that was renamed to $z'$ in $\widehat q'$ set
  ${z'}^\downarrow=z^\downarrow$.  Note that $z^\downarrow$ is defined
  for all variables $z$ that occur in queries in
  $\Fmc_0(x)$. Also note that, in Point~1b of (A), it is
    important to use $q$ rather than $q_x$ as $y$ could be a
    predecessor
    of $x$ in $q$.

\paragraph{\textbf{Step~2: Compensate.}}
We
construct a frontier $\Fmc$ of $q(x_0)$
by including, for each $p \in \Fmc_0(x_0)$, the ELIQ obtained from $p$
by the following two steps.
We write $x\rightsquigarrow_{q,\Omc}^R A$
if $\Amc_q,\Omc\models \exists R.A(x)$ and there is no
$S(x,y) \in q$ with $\Omc\models S\sqsubseteq R$ and $\Amc_q,\Omc \models A(y)$.

\medskip \emph{Step 2A.} Consider all $x \in \mn{var}(p)$, roles
$R,S$, and concept names $A$ such that
$x^\downarrow\rightsquigarrow_{q,\Omc}^R A$, $\Omc \models R \sqsubseteq S$, and
$\Omc \models \exists S \sqsubseteq B$ implies $B(x) \in p$ for all concept
names $B$. Add the atoms $S(x,z),A(z), R(x', z)$ where $z$ and $x'$ are fresh variables with $z^\downarrow$ undefined, $x'^\downarrow = x^\downarrow$, and
add a disjoint copy $\widehat q$ of $q$, glue the copy of $x^\downarrow$ in $\widehat q$ to $x'$.

\medskip 

\emph{Step 2B.} Consider every $S(x,y) \in p$ directed away from~$x_0$ that
was not added in Step~2A. Then $x^\downarrow$ and $y^\downarrow$ are
defined.
For every role $R$ with $\Amc_q,\Omc\models
R(x^\downarrow,y^\downarrow)$,
add an atom $R(z,y)$, $z$ a fresh variable with
$z^\downarrow=x^\downarrow$, as well as a disjoint copy $\widehat q$
of $q$ and glue the copy of $x^\downarrow$ in $\widehat q$ to~$z$.

\smallskip

This finishes the construction of the frontier \Fmc of $q$.

\begin{exmp}
  Consider the \DLR ontology
  $\Omc = \{ A \sqsubseteq \exists r,\ \exists r \sqsubseteq A,\ r
  \sqsubseteq s\}$ and the ELIQ $q(x_0)=A(x_0) \wedge B(x_0)$. Then
  \Fmc contains the ELIQs $p_1$ and $p_2$ shown below:
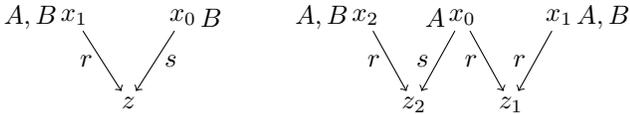
\begin{center}
\tikzstyle{answer} = [inner sep = 2pt]
\tikzstyle{exists} = [inner sep = 2pt]
\tikzstyle{role} = [->]
\begin{tikzpicture}


\node (b_1) [answer] {$x_0$};
\node (b_1A) [right= -0.15cm of b_1] {$B$};
\node (b_2) [exists, below left = 0.8cm and 0.3cm of b_1] {$z$};
\node (b_3) [exists, above left = 0.8cm and 0.3cm of b_2] {$x_1$};
\node (b_3A) [left= -0.15cm of b_3] {$A, B$};

\node (b_4) [exists, right= 3.2cm of b_1] {$x_0$};
\node (b_4A) [left= -0.15cm of b_4] {$A$};
\node (b_5) [exists, below right = 0.8cm and 0.15cm of b_4] {$z_1$};
\node (b_6) [exists, above right = 0.8cm and 0.15cm of b_5] {$x_1$};
\node (b_6A) [right= -0.15cm of b_6] {$A, B$};
\node (b_7) [exists, below left = 0.8cm and 0.15cm of b_4] {$z_2$};
\node (b_8) [exists, above left = 0.8cm and 0.15cm of b_7] {$x_2$};
\node (b_8A) [left= -0.15cm of b_8] {$A, B$};

\draw[role] (b_1) to node [right] {$s$} (b_2);
\draw[role] (b_3) to node [left] {$r$} (b_2);

\draw[role] (b_4) to node [left] {$r$} (b_5);
\draw[role] (b_6) to node [left] {$r$} (b_5);

\draw[role] (b_4) to node [left] {$s$} (b_7);
\draw[role] (b_8) to node [left] {$r$} (b_7);

\end{tikzpicture}
\end{center}
ELIQ $p_1$ is the result of dropping the concept atom $A(x_0)$ and
$p_2$ is the result of dropping the concept atom $B(x_0)$.
Step~2A adds an $r$-successor and an $s$-successor of $x_0$ in $p_2$ but
only an $s$-successor in  $p_1$ as $\Omc \models \exists r \sqsubseteq
A$, and then attaches copies of $q$.
Step~2B does nothing, as all role atoms have been added in
Step~2A.\footnote{Variables $x_2$ and $z_2$ can be dropped from $p_2$
resulting in an ELIQ that is equivalent w.r.t.\ \Omc. We did not include such optimizations in the
  compensation
  step to avoid making it more complicated.}
\end{exmp}

\begin{exmp}
Consider the \DLR ontology $\Omc = \{ r \sqsubseteq s\}$ and the ELIQ $q(x_0)$
shown on the left-hand side below:
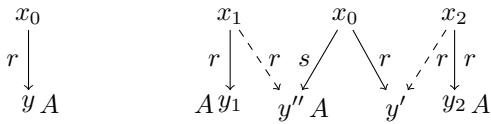
\begin{center}
\tikzstyle{answer} = [inner sep = 2pt]
\tikzstyle{exists} = [inner sep = 2pt]
\tikzstyle{role} = [->]
\begin{tikzpicture}

\node (a_1) [answer] {$x_0$};
\node (a_2) [exists, below = 0.8cm of a_1] {$y$};

\node (a_2A) [right= -0.15cm of a_2] {$A$};
\draw[role] (a_1) to node [left] {$r$} (a_2);

\node (b_1) [answer, right= 3.7cm of a_1] {$x_0$};
\node (b_2) [exists, below left= 0.8cm and 0.2cm of b_1] {$y''$};
\node (b_3) [exists, below right= 0.8cm and 0.2cm of b_1] {$y'$};
\node (b_2A) [right= -0.15cm of b_2] {$A$};

\node (b_4) [exists, above left = 0.8cm and 0.3cm of b_2] {$x_1$};
\node (b_5) [exists, below = 0.8cm of b_4] {$y_1$};
\node (b_5A) [left= -0.15cm of b_5] {$A$};

\node (b_7) [exists, above right= 0.8cm and 0.3cm of b_3] {$x_2$};
\node (b_8) [exists, below = 0.8cm of b_7] {$y_2$};
\node (b_8A) [right= -0.15cm of b_8] {$A$};

\draw[role] (b_1) to node [left] {$s$} (b_2);
\draw[role] (b_1) to node [right] {$r$} (b_3);

\draw[role, dashed] (b_4) to node [right] {$r$} (b_2);
\draw[role] (b_4) to node [left] {$r$} (b_5);

\draw[role, dashed] (b_7) to node [right] {$r$} (b_3);
\draw[role] (b_7) to node [right] {$r$} (b_8);

\end{tikzpicture}
\end{center}
Then \Fmc contains only the ELIQ $p$ shown on the right-hand side.
It is the result of
dropping the concept atom $A(y)$ in~$q_{y}$, then generalizing the
subquery $r(x_0,y)$
in $q_{x_0} = q$, and then compensating.  Step~2A of compensation adds nothing.
Step~2B adds the two dashed role atoms and attaches copies of $q$ to
$x_1$ and~$x_2$.
\end{exmp}
\begin{restatable}{lemma}{lemfrontiermainh}
  \label{lem:frontiermain-r}
  $\Fmc$ is a frontier of
  $q(x_0)$ w.r.t.\ $\Omc$.
\end{restatable}
We next show that the constructed frontier is of polynomial
size and that its computation takes
only polynomial time. 
%
\begin{restatable}{lemma}{lemfrontiersize}
  \label{lem:frontiersize}
  The construction of \Fmc runs in time polynomial in $||q||+||\Omc||$
  (and thus $\displaystyle \sum_{p \in \Fmc} ||p||$ is
  polynomial in $||q||+||\Omc||$).
\end{restatable}
We next observe that adding conjunction to \DL destroys polynomial
frontiers and thus Theorem~\ref{thm:frontiermain} does not apply to
\DLhorn ontologies \cite{dllite-jair09}. In fact, this already holds for very
simple queries and ontologies, implying that also for other DLs that
support conjunction such as \EL, polynomial frontiers are elusive.  A
\emph{conjunction of atomic queries (AQ$^\land$)} is a unary CQ of the form
$q(x_0) = A_1(x_0) \land \dots \land A_n(x_0)$ and a \emph{conjunctive
  ontology} is a set of CIs of the form
$A_1 \sqcap \cdots \sqcap A_n \sqsubseteq A$ where $A_1, \dots, A_n$ and $A$ are concept names.
\begin{restatable}{theorem}{thmlowerfrontier}
  \label{thm:lowerfrontier}
  There are families of AQ$^\land$s $q_1,q_2,\dots$ and conjunctive
  ontologies $\Omc_1,\Omc_2,\dots$ such that for all $n \geq 1$, any
  frontier of $q_n$ w.r.t.\ $\Omc_n$ has size at least~$2^n$.
\end{restatable}
%

\section{Frontiers in \texorpdfstring{\DLF}{DL-LiteF}}
\label{sect:dllitef}

We start by observing that frontiers of ELIQs w.r.t.~\DLF ontologies
may be infinite. This leads us to identifying a syntactic restriction
on \DLF ontologies that regains finite frontiers. In fact, we show that they
are of polynomial size and can be computed in polynomial time. 
%
\begin{restatable}{theorem}{thmnofinitefrontier}\label{thm:no-frontier}
  There is an ELIQ $q$ and a \DLF{} ontology $\Omc$ such that $q$ does
  not have a  finite frontier w.r.t.\ $\Omc$.
\end{restatable}
In the proof of Theorem~\ref{thm:no-frontier}, we use the ELIQ
$A(x)$ and 
    \[
        \Omc = \{\ A \sqsubseteq \exists r,\quad
                  \exists r^- \sqsubseteq \exists r,\quad
                  \exists r \sqsubseteq \exists s,\quad
                  \mn{func}(r^-)\ \}.
    \]
    The universal model $\Umc_{q, \Omc}$ of $\Amc_q$ and $\Omc$ is an
    infinite $r$-path on which every point has an $s$-successor.
    Now consider the following ELIQs $q_1,q_2,\dots$ that
    satisfy $q_i \not\subseteq_\Omc q \subseteq_\Omc q_i$:
    \begin{align*}
      q_i(x_1) ={ } & r(x_1, x_2), \dots, r(x_{n - 1}, x_n), 
                       s(x_n, y), s(x'_n, y), \\
                      & r(x_1', x_2'), \dots, r(x_{n - 1}', x_n'), A(x_1').
    \end{align*}
    Any frontier $\Fmc$ must contain a $p_i$ with
    $p_i \subseteq_\Omc q_i$ for all \mbox{$i \geq 1$}. We show that,
    consequently, there is no bound on the size of the queries in \Fmc.
    We invite the reader to apply the frontier construction from
    Section~\ref{sect:frontier} after dropping
    $\mn{func}(r^-)$.

The proof actually shows that there is no finite frontier even if we
admit the use of unrestricted CQs in the frontier in place of ELIQs.
To regain finite frontiers, we restrict our attention to \DLF
ontologies \Omc such that if $B \sqsubseteq C$ is a CI in~\Omc, then
$C$ contains no subconcept of the form $\exists R . D$ with
\mbox{$\mn{func}(R^-) \in \Omc$}. We call such an ontology a
$\DLFsynr$ ontology. We again concentrate on ontologies in normal
form.
\begin{restatable}{lemma}{lemfrontiernormalformF}
  \label{lem:frontier-normal-form-f} For every \DLFsynr
    ontology \Omc, we can construct in polynomial time a \DLFsynr
    ontology $\Omc'$ in normal form such that for every ELIQ $q$, a
    frontier of $q$ w.r.t.\ $\Omc$ can be constructed in polynomial
    time given a frontier of $q$ w.r.t.\ $\Omc'$.
\end{restatable}
The main result of this section is as follows. 
\begin{theorem} \label{thm:frontiermainF} Let $\Omc$ be a \DLFsynr
  ontology and $q$ an ELIQ that is satisfiable w.r.t.\ $\Omc$.  Then a
  frontier of $q$ w.r.t.\ $\Omc$ can be computed in polynomial time.
\end{theorem}
To prove Theorem~\ref{thm:frontiermainF}, let \Omc and $q$ be as in
the theorem, \Omc in normal form. We may
assume w.l.o.g.\ that $q$ is \Omc-minimal and 
\Omc-saturated. The
construction of a frontier follows the same general approach as for
\DLR, but the presence of functional roles significantly complicates
the compensation step. As before, we introduce
fresh variables and rely on the mapping $x^\downarrow$.

\paragraph{\textbf{Step~1: Generalize.}}
For each variable $x \in \mn{var}(q)$, 
define a set $\Fmc_0(x)$ that contains all ELIQs which can be obtained
by starting with $q_x(x)$ and then doing one of the following:

\smallskip
\noindent
(A)~\emph{Drop concept atom}: exactly as for \DLR.
%


\smallskip
\noindent
(B)~\emph{Generalize subquery:}
  \begin{enumerate}
  \item choose an atom $R(x,y) \in q$ directed away from $x_0$;
  \item remove $R(x, y)$ and all atoms of $q_y$;

  \item if $\mn{func}(R) \notin \Omc$, then for each $q'(y) \in
    \Fmc_0(y)$ add a disjoint copy $\widehat q'$ of $q'$ and the role atom
    $R(x,y')$ with $y'$ the copy of $y$ in $\widehat q'$;

  \item  if $\mn{func}(R) \in \Omc$ and $\Fmc_0(y) \neq \emptyset$,
    then choose and add a
      $q' \in~\Fmc_0(y)$ and the role atom $R(x,y)$.

  \end{enumerate}


  \paragraph{\textbf{Step~2: Compensate.}}
  We construct a frontier $\Fmc$ of
  $q(x_0)$ by including, for each $p \in \Fmc_0(x_0)$, the CQ obtained
  from $p$ by the following two steps. For $x\in\mn{var}(q)$, $R$ a
  role, and $M$ a set of concept names from $\Omc$, we write
  $x\rightsquigarrow_{q,\Omc}^R M$ if $M$ is maximal with
  $\Amc_q,\Omc\models \exists R.\bigsqcap M(x)$ and
  there is no $R(x,y) \in q$ with $\Amc_q,\Omc\models \bigsqcap M(y)$.

\medskip 

\emph{Step 2A.} Consider every $x \in \mn{var}(p)$, role $R$,
and set of concept names $M=\{A_1,\ldots,A_k\}$ with
$x^\downarrow\rightsquigarrow_{q,\Omc}^R M$.
If $\Omc \models \exists R \sqsubseteq B$ implies $B(x) \in p$
for all concept names~$B$,
add the atoms $R(x,z),A_1(z),\ldots,A_k(z)$ where $z$ is a fresh
variable, 
and leave $z^\downarrow$ undefined.

\medskip 

\emph{Step 2B.} This step is iterative. For bookkeeping, we mark
atoms $R(x,y) \in p$ to be processed in the next round of the
iteration. Marking is only applied to atoms $R(x,y)$ directed away
from $x_0$ such that $y^\downarrow$ is defined and if $x^\downarrow$
is undefined then $\mn{func}(R^-) \notin \Omc$ or $q$ contains no atom of
the form $R(y^\downarrow,z)$.

\smallskip
\emph{Start.} Consider every $R(x,y) \in p$ directed away
from $x_0$ with $\mn{func}(R^-) \notin \Omc$.  
Then $x^\downarrow$ is defined.
Extend $p$ with atom $R^-(y,x')$,
$x'$ a fresh variable with $x^{\prime\downarrow}=x^\downarrow$.
Mark the new atom.

\smallskip \emph{Step.}  Choose a marked atom $R(x,y)$ 
and unmark it. If $\mn{func}(R^-) \notin \Omc$ or $q$ contains no atom
of the form $R(y^\downarrow,z)$, then add a disjoint copy $\widehat q$ of $q$ and
glue the copy of $y^\downarrow$ in $\widehat q$ to~$y$.  Otherwise, do
the following:
%
\begin{enumerate}

\item[(i)] add $A(y)$ whenever $\Amc_q,\Omc\models A(y^\downarrow)$;
  
\item[(ii)] for all atoms $S(y^\downarrow,z) \in q$ with
  $S(y^\downarrow,z)\neq R^-(y^\downarrow,x^\downarrow)$, extend $p$
  with atom $S(y,z')$, $z'$~a fresh variable with
  \mbox{$z^{\prime\downarrow} = z$}. Mark $S(y,z')$.

\item[(iii)] For all roles $S$ and sets $M=\{A_1,\ldots,A_k\}$ such
  that $y^\downarrow\rightsquigarrow_{q,\Omc}^{S} M$, extend $p$ with atoms
  $S(y,u),$ $S^-(u,y'),A_1(u),\ldots,A_k(u)$ where $u$ and $y'$ are fresh
  variables. Set $y^{\prime\downarrow}=y^\downarrow$ and mark $S^-(u,y')$.

%
  \end{enumerate}
The step is repeated as long as possible.  Note that in Point~(iii),
the role $S$ must occur on the right-hand side of some CI in the
\DLFsynr ontology \Omc. Consequently, $\mn{func}(S^-) \notin \Omc$
and it is not a problem that $u$ receives two $S$-predecessors.
  Also in Point~(iii), $\mn{func}(S) \in \Omc$ implies that $q$ cannot contain an atom $S(y^\downarrow,z)$
  due to the definition of `$\rightsquigarrow$' and
thus we may leave $u^\downarrow$ undefined.
 
%
This finishes the construction of the frontier \Fmc of $q$. 
\begin{exmp}
  Consider the ontology $\Omc = \{ \mn{func}(s) \}$ and ELIQ~$q(x_0)$
  shown on the left-hand side below:
%
%
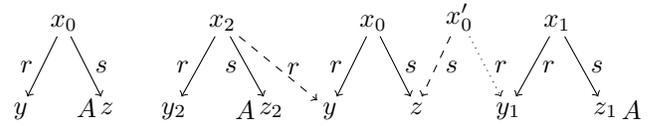
\begin{center}
\tikzstyle{answer} = [inner sep = 2pt]
\tikzstyle{exists} = [inner sep = 2pt]
\tikzstyle{role} = [->]
\begin{tikzpicture}

\node (a_1) [answer] {$x_0$};
\node (a_2) [exists, below left= 0.8cm and 0.15cm of a_1] {$y$};
\node (a_3) [exists, below right= 0.8cm and 0.15cm of a_1] {$z$};

\node (a_3A) [left= -0.15cm of a_3] {$A$};
\draw[role] (a_1) to node [left] {$r$} (a_2);
\draw[role] (a_1) to node [right] {$s$} (a_3);

\node (b_1) [answer, right= 3.6cm of a_1] {$x_0$};
\node (b_2) [exists, below left= 0.8cm and 0.15cm of b_1] {$y$};
\node (b_3) [exists, below right= 0.8cm and 0.15cm of b_1] {$z$};

\node (b_4) [exists, above right = 0.8cm and 0.15cm of b_3] {$x_0'$};

\node (b_6) [exists, below right= 0.8cm and 0.15cm of b_4] {$y_1$};
\node (b_5) [exists, above right= 0.8cm and 0.15cm of b_6] {$x_1$};
\node (b_7) [exists, below right= 0.8cm and 0.15cm of b_5] {$z_1$};
\node (b_7A) [right= -0.15cm of b_7] {$A$};

\node (b_8) [exists, above left= 0.8cm and 1cm of b_2] {$x_2$};
\node (b_9) [exists, below right= 0.8cm and 0.15cm of b_8] {$z_2$};
\node (b_10) [exists, below left= 0.8cm and 0.15cm of b_8] {$y_2$};
\node (b_9A) [left= -0.15cm of b_9] {$A$};

\draw[role] (b_1) to node [left] {$r$} (b_2);
\draw[role] (b_1) to node [right] {$s$} (b_3);
\draw[role, dashed] (b_4) to node [right] {$s$} (b_3);
\draw[role, dotted] (b_4) to node [right] {$r$} (b_6);

\draw[role] (b_5) to node [right] {$r$} (b_6);
\draw[role] (b_5) to node [right] {$s$} (b_7);

\draw[role, dashed] (b_8) to node [right] {$r$} (b_2);
\draw[role] (b_8) to node [left] {$s$} (b_9);
\draw[role] (b_8) to node [left] {$r$} (b_10);
\end{tikzpicture}
\end{center}
The ELIQ $p \in \Fmc$ shown on the right-hand side is the result of
dropping the concept atom $A(z)$ in $q_{z}$, then generalizing the
subquery $s(x_0,z)$ in $q_{x_0} = q$, and then compensating.  Step~2A of
compensation adds nothing.  The start of Step~2B adds the two dashed
role atoms and marks them.  The step of Step~2B adds the dotted role
atom via Point~(ii) and marks it. When the step of Step~2B
processes role atoms $r^-(y,x_2)$ and $r(x_0',y)$, it attaches copies of
$q$ to $x_2$ and $y_1$.  Note that directly attaching a copy
of $q$ to $x_0'$ would violate $\mn{func}(s)$.
\end{exmp}
\begin{restatable}{lemma}{lemfrontiermainf}
  \label{lem:frontiermain2}
  $\Fmc$ is a frontier of
  $q(x_0)$ w.r.t.\ $\Omc$.
\end{restatable}
As for \DLR, the constructed frontier is of polynomial size and its
computation takes only polynomial time. Crucially, the iterative
process in Point~2B terminates since in Step~(ii) a (copy of a)
subquery of $q$ is added and the process stops at atoms added in
Step~(iii).
\begin{restatable}{lemma}{lemfrontiersizeF}
  \label{lem:frontiersizeF}
  The construction of \Fmc runs in time polynomial in $||q||+||\Omc||$
  (and thus $\displaystyle \sum_{p \in \Fmc} ||p||$ is
  polynomial in $||q||+||\Omc||$).
\end{restatable}
%

\section{Uniquely Characterizing ELIQs}
\label{sect:uniquechar}

As a first application of our results on frontiers, we consider the
unique characterization of ELIQs in terms of polynomially many data
examples. One area where this is relevant is reverse query
engineering, also known as \emph{query-by-example (QBE)}, which in a
DL context was studied
in~\cite{GuJuSa-IJCAI18,DBLP:conf/ijcai/FunkJLPW19,DBLP:conf/gcai/Ortiz19}.
The idea of QBE is that a query is not formulated directly, but
derived from data examples that describe its behavior. The results in
this section imply that every ELIQ can be described up to equivalence by
such examples this is always possible and that a
reasonable number of examples of reasonable size suffices.

Formally, a \emph{data example} takes the form $(\Amc,a)$ where \Amc
is an ABox and $a \in \mn{ind}(\Amc)$. Let $E^+$, $E^-$ be finite sets
of data examples. We say that an ELIQ $q$ \emph{fits $(E^+,E^-)$
  w.r.t.\ a \DLRF$\!\!\!\!\!$ ontology} \Omc if $(\Amc,a) \in E^+$ implies
$\Amc,\Omc \models q(a)$ and $(\Amc,a) \in E^-$ implies
$\Amc,\Omc \not\models q(a)$.  Then $(E^+,E^-)$ \emph{uniquely
  characterizes} $q$ w.r.t.\ \Omc if $q$ fits $(E^+,E^-)$ and every
ELIQ $q'$ that also fits $(E^+,E^-)$ satisfies $q \equiv_\Omc q'$. The
following is a consequence of Theorems~\ref{thm:frontiermain}
and~\ref{thm:frontiermainF}, see also
\cite{DBLP:conf/icdt/CateD21}.
\begin{restatable}{theorem}{thmunique}
  \label{thm:unique}
  Let \Omc be an ontology formulated in \DLR or \DLFsynr. Then for
  every ELIQ $q$ that is satisfiable w.r.t.~\Omc, there are sets of
  data examples $(E^+,E^-)$ that uniquely characterize $q$ w.r.t.\
  \Omc and such that $||(E^+,E^-)||$ is polynomial in
  $||q||+||\Omc||$.  If \Omc is a \DLFsynr ontology, then $(E^+,E^-)$
  can be computed in polynomial time and the same holds for \DLR if $q$
  is \Omc-minimal.
\end{restatable}

\section{Learning ELIQs under Ontologies}\label{sec:learning}

We use our results on frontiers to show that ELIQs are polynomial
time learnable under ontologies formulated in \DLR and \DLFsynr, using
only membership queries.  We also present two results on
non-learnability.  
\begin{theorem}\mbox{}
  \label{thm:learnable}
  ELIQs are polynomial time learnable under $\DLR$ ontologies and
  under \DLFsynr ontologies using only membership queries.

  If the ontology contains concept disjointness constraints, then this
  only holds true if the learner is provided with a seed~CQ
  (definition given below).
\end{theorem}
%
For proving Theorem~\ref{thm:learnable}, let \Omc be an ontology
formulated in \DLR or \DLFsynr and $q_T(x_0)$ the target ELIQ known to
the oracle. We may again assume \Omc to be in normal form.
\begin{restatable}{lemma}{lemlearningnormalform}
  \label{lem:learning-normal-form}\mbox{}
  In \DLR and \DLFsynr, every polynomial time learning algorithm for
  ELIQs under ontologies in normal form that uses only membership
  queries can be transformed into a learning algorithm with the same
  properties for ELIQs under unrestricted
  ontologies. 
%
%
\end{restatable}
%

The learning
algorithm is displayed as Algorithm~\ref{alg:learn}.
\begin{algorithm}[t]
\caption{Learning ELIQs under \DL ontologies}
\label{alg:learn}
\begin{algorithmic}
  \State \textbf{Input} An 
  ontology \Omc in normal form and a CQ $q^0_H$ satisfiable w.r.t.\ \Omc such that $q^0_H\subseteq_\Omc q_T$
    \State \textbf{Output} An ELIQ $q_H$ such that $q_H \equiv_\Omc q_T$
    \medskip 
    \State $q_H := \mn{treeify}(q^0_H)$
    \While{there is a $q_F \in \Fmc_{q_H}$ with $q_F \subseteq_\Omc q_T$}
        \State $q_H := \mn{minimize}(q_F)$ 
    \EndWhile 
    \State \Return $q_H$
\end{algorithmic}
\end{algorithm}
It assumes a \emph{seed CQ} $q^0_H$, that is, a CQ $q^0_H$ such that
$q^0_H \subseteq_\Omc q_T$ and $q^0_H$ is satisfiable w.r.t.\ \Omc. If
\Omc contains no disjointness constraints, then for
$\Sigma=\mn{sig}(\Omc)$ we can use as the seed CQ
 \begin{equation*}\label{eq:qbot}
 q^0_H(x_0) = \{A(x_0)\mid A\in \Sigma\cap\mn{N}_\mn{C}\}\cup
 \{r(x_0,x_0)\mid r\in \Sigma\cap \mn{N}_\mn{R}\}.
\end{equation*}
We can still construct a seed CQ $q^0_H$ in time polynomial in
$||\Omc||$ if \Omc contains no disjoint constraints on concepts (but
potentially on roles); details are in the appendix. In the presence
of concept disjointness constraints, a seed CQ can be obtained through
an initial equivalence query.


The algorithm constructs and
repeatedly updates a hypothesis ELIQ $q_H$ while maintaining the
invariant $q_H
\subseteq_\Omc
q_T$.  The initial call to subroutine \mn{treeify} yields an ELIQ
$q_H$ with $q^0_H \subseteq_\Omc q_H \subseteq_\Omc
q_T$ to be used as the first hypothesis. The algorithm then
iteratively generalizes
$q_H$ by constructing the frontier $\Fmc_{q_H}$ of
$q_H$ w.r.t.\ \Omc in polynomial time and choosing from it a new ELIQ
$q_H$ with $q_H \subseteq_\Omc
q_T$. In between, the algorithm applies the
$\mn{minimize}$ subroutine to ensure that the new
$q_H$ is
$\Omc$-minimal and to avoid an excessive blowup while iterating in the
while loop.


We next detail the subroutines \mn{treeify} and \mn{minimize}. We
define \mn{minimize} on unrestricted CQs since it is applied to
non-ELIQs as part of the \mn{treeify} subroutine.

\paragraph{\bf The \mn{minimize} subroutine.} The subroutine takes as
input a unary CQ $q(x_0)$ that is 
satisfiable w.r.t.\ $\Omc$ and satisfies
\mbox{$q \subseteq_\Omc q_T$}. It computes a unary CQ $q'$ with
$q \subseteq_\Omc q' \subseteq_\Omc q_T$ using membership queries
that is minimal in a strong sense. Formally, 
%
\mn{minimize} first makes sure that $q$ is \Omc-saturated and
then exhaustively applies the following operation:

\smallskip
\noindent
\emph{Remove atom.} Choose a role atom $r(x, y) \in q$ and let
$q^-$ be the maximal connected component of $q \setminus \{r(x, y)\}$ that contains $x_0$.
Pose
the membership query $\Amc_{q^-}, \Omc \models q_T(x_0)$. If the response is
positive, continue with $q^-$ in place of $q$.

\smallskip
Clearly, the result of \mn{minimize} is \Omc-minimal.

\paragraph{\bf The \mn{treeify} subroutine.} The subroutine takes as input a
unary CQ $q(x_0)$ that is satisfiable w.r.t.\ $\Omc$,
and satisfies \mbox{$q \subseteq_\Omc q_T$}.  It computes an ELIQ
$q'$ with $q \subseteq_\Omc q' \subseteq_\Omc q_T$
by repeatedly increasing the length of cycles in $q$ and 
minimizing the obtained query; a similar construction is used
in~\cite{DBLP:conf/icdt/CateD21}. The resulting ELIQ
is \Omc-minimal.

Formally, \mn{treeify} first makes sure that $q(x_0)$ is \Omc-saturated
and then constructs a sequence of CQs $p_1, p_2, \dots$ starting with
$p_1 = \mn{minimize}(q)$ and then taking
$p_{i + 1} = \mn{minimize}(p_i')$ where $p'_i$ is obtained from $p_i$
by doubling the length of some cycle. Here, a \emph{cycle} in a CQ $q$
is a sequence $R_1(x_1, x_2), \dots, R_n(x_n, x_1)$ of distinct role
atoms in $q$ such that $x_1, \dots x_n$ are distinct.  More precisely,
$p_i'$ is the result of the following operation.

\smallskip
\noindent
\emph{Double cycle.} Choose a role atom $r(x, y) \in p_i$ that is part of a cycle in $p_i$ and let $p$ be $p_i \setminus \{ r(x, y) \}$.
The CQ $p_i'$ is then obtained by starting with $p$, adding a disjoint
copy $p'$ of $p$ where $x'$ refers to the copy of $x \in
\mn{var}(p)$ in $p'$ and adding the role atoms $r(x, y'), r(x', y)$.

\smallskip

If $p_i$ contains no more cycles, \mn{treeify} stops and returns~$p_i$.

\smallskip

Returning to Algorithm~\ref{alg:learn}, let $q_1, q_2, \dots$ be the
sequence of ELIQs that are assigned to $q_H$ during a run of the
learning algorithm. We show in the appendix that for all $i \geq 1$,
it holds that $q_i \subseteq_\Omc q_T$, $q_i \subseteq_\Omc q_{i + 1}$
while $q_{i + 1} \not\subseteq_\Omc q_i$, and
$|\mn{var}(q_{i+1})| \geq |\mn{var}(q_{i})|$.  This can be used to
prove that the while loop in Algorithm~\ref{alg:learn} terminates after
a polynomial number of iterations, arriving at a hypothesis $q_H $
with $q_H \equiv_\Omc q_T$.


\smallskip

We now turn to non-learnability results. Without a
seed CQ, ontologies with concept disjointness constraints are not
learnable using only polynomially many membership queries.
A \emph{disjointness ontology} is a \DLRF ontology that only consists
of concept disjointness constraints.
\begin{restatable}{theorem}{thmlower} \label{thm:lower}
  AQ$^\land$s are not learnable under disjointness ontologies using only
  polynomially many membership queries.
\end{restatable}
%
We next show that when we drop the syntactic restriction from
\DLFsynr, then ELIQs are no longer learnable at all using only
membership queries. Note that this is not a direct consequence of
Theorem~\ref{thm:no-frontier} as there could be an alternative
approach that does not use frontiers.
\begin{restatable}{theorem}{thmnolearn}\label{thm:no-learn}
  ELIQs are not learnable under \DLF ontologies using only membership queries. 
\end{restatable}

\section{Outlook}

A natural next step for future work is to generalize the results
presented in this paper to \DLRF, adopting the same syntactic
restriction that we have adopted for \DLF, and additionally requiring
that functional roles have no proper subroles. The latter serves to
control the interaction between functional roles and role
inclusions. Even with this restriction, however, that interaction is
very subtle and the frontier construction becomes significantly
more complex.
%
Other interesting questions are whether ELIQs can be learned in
polynomial time w.r.t.\ \DLhorn ontologies and whether CQs can
be learned w.r.t.\ \DLcore ontologies when both membership and
equivalence queries are admitted.


\noindent
\section*{Acknowledgements}
Carsten Lutz was supported by DFG CRC 1320 EASE.

\begin{small}
\bibliographystyle{named}
\bibliography{local}

\end{small}
\appendix

\newpage

\section{Additional Preliminaries}
\label{app:addprelim}

We introduce some additional preliminaries that are needed for the lemmas and
proofs in the appendix.

We start with defining the semantics of conjunctive queries in full
detail.  A \emph{homomorphism} $h$ from interpretation $\Imc_1$ to
interpretation $\Imc_2$ is a mapping from $\Delta^{\Imc_1}$ to
$\Delta^{\Imc_2}$ such that $d \in A^{\Imc_1}$ implies
$h(d) \in A^{\Imc_2}$ and $(d, e) \in r^{\Imc_1}$ implies
$(h(d), h(e)) \in r^{\Imc_2}$.  We use $\mn{img}(h)$ to denote the set
$\{ e \in \Delta^{\Imc_2} \mid \exists d \in \Delta^{\Imc_1}\colon
h(d) = e\}$.  For $d_i \in \Delta^{\Imc_i}$, $i \in \{1, 2\}$, we
write $\Imc_1, d_1 \to \Imc_2, d_2$ if there is a homomorphism $h$
from $\Imc_1$ to $\Imc_2$ with $h(d_1) = d_2$. With a homomorphism
from a CQ $q$ to an interpretation $\Imc$, we mean a homomorphism from
$\Amc_q$ to $\Imc$.  For a CQ $q(x_0)$, we write
$q(x_0) \to (\Imc, d)$ if there is a homomorphism $h$ from $q$ to $\Imc$
with $h(x_0) = d$.  Let $q(x_0)$ be a CQ and $\Imc$ an
interpretation. An element $d \in \Delta^\Imc$ is an \emph{answer to
  $q$ in $\Imc$}, written $\Imc \models q(d)$, if
$q(x_0) \to (\Imc, d)$. Now let $\Omc$ be an ontology and $\Amc$ an
ABox. An individual $a \in \mn{ind}(\Amc)$ is an \emph{answer to $q$
  on $\Amc$ w.r.t.~$\Omc$}, written $\Amc, \Omc \models q(a)$, if $a$
is an answer to $q$ in every model of $\Omc$ and $\Amc$. 

Let $q(x_0)$ be an ELIQ. The \emph{codepth} of a variable $x \in \mn{var}(q)$
is $0$ if there is no $R(x, x') \in q$ directed away from $x_0$ and is $k + 1$
if $k$ is the maximum of the codepths of all $x' \in \mn{var}(q)$ with $R(x,
x') \in q$ directed away from $x_0$.


\smallskip

We next define universal models, first for \DLR and then for \DLF.
Let $\Omc$ be a \DLR ontology and let $\Amc$ be an ABox that is
satisfiable w.r.t.\ \Omc. For $a \in \mn{ind}(\Amc)$, concept names
$A$, and $R$ a role, we write $a\rightsquigarrow_{\Amc,\Omc}^R A$
if $\Amc,\Omc\models \exists R.A(a)$ and there is no $S(a,b)\in\Amc$
such that $\Omc\models S\sqsubseteq R$ and $\Amc,\Omc \models A(b)$. Note that this is
identical to the definition of `$\rightsquigarrow$' given in
Section~\ref{sect:frontier}, but is formulated for ABoxes in place of
ELIQs.

A \emph{trace} for $\Amc$ and $\Omc$ is a sequence
$t = aR_1A_1R_2A_2\dots R_nA_n$, $n \geq 0$ where
$a \in \mn{ind}(\Amc)$, $R_1,\dots,R_n$ are roles that occur in \Omc,
and $A_1,\dots,A_n$ are sets of concept names that occur in \Omc, such
that $a\rightsquigarrow_{\Amc,\Omc}^{R_1}A_1$ and
$\Omc \models A_i\sqsubseteq \exists R_{i+1} . A_{i+1}$ for $1 \leq i < n$.  Let
$\Tbf$ denote the set of all traces for $\Amc$ and $\Omc$. Then the
\emph{universal model} of \Amc and \Omc is
\begin{align*}
    \Umc_{\Amc, \Omc} = {} & \Amc \cup \{ A(a) \mid \Amc, \Omc \models
                             A(a)\} \cup {}\\
                           & \{ S(a,b) \mid R(a,b) \in \Amc \text{ and
			   } \Omc \models R \sqsubseteq S\}\cup {}\\
                           & \{B(tRA) \mid tRA \in \Tbf \text{ and }
                             \Omc \models A \sqsubseteq B \}
   \cup {}\\
   &\{S(t, tRA) \mid tRA \in \Tbf \text{ and }\Omc\models R\sqsubseteq
 S\}.
\end{align*}
For brevity, we write $\Umc_{q, \Omc}$ instead of $\Umc_{\Amc_q, \Omc}$ and $x
\rightsquigarrow^{R}_{q, \Omc} A$ instead of $x \rightsquigarrow^R_{\Amc_q,
\Omc} A$ for any conjunctive query $q$.

\medskip

Now let $\Omc$ be a \DLF ontology and let $\Amc$ be an ABox that is
satisfiable w.r.t.\ \Omc. For a set $M$ of concept names, we write
$\bigsqcap M$ as a shorthand for $\bigsqcap_{A\in M}A$. For
$a \in \mn{ind}(\Amc)$, $M,M'$ sets of concept names, and $R$ a role,
we write
\begin{itemize}

\item $a\rightsquigarrow_{\Amc,\Omc}^R M$ if
  $\Amc,\Omc\models \exists R.\bigsqcap M(a)$, $M$ is maximal with
  this condition, and there is no $R(a,b)\in\Amc$ such that $\Amc,\Omc
  \models \bigsqcap M(b)$;

  \item $M\rightsquigarrow_{\Omc}^R M'$ if $\Omc\models
    \bigsqcap M\sqsubseteq \exists R.\bigsqcap M'(a)$ and $M'$ is maximal with
    this condition.

\end{itemize}
The definition of `$\rightsquigarrow$' in the first item is identical
to the definition of `$\rightsquigarrow$' given in
Section~\ref{sect:dllitef}, but is formulated for ABoxes in place of
ELIQs.
The maximality of $M$ is important when dealing with functional roles.
If, for example, \Omc contains $A \sqsubseteq \exists r.B_1$,
$A \sqsubseteq \exists r.B_2$, and $\mn{func}(r)$ and $\Amc= \{ A(a) \}$, then
we have $a \rightsquigarrow_{\Amc,\Omc}^r \{ B_1,B_2\}$, but not
$a \rightsquigarrow_{\Amc,\Omc}^r \{ B_i\}$ for an $i \in
\{1,2\}$. This helps to ensure that $a$ gets only a single $r$-successor in
the universal model.

A \emph{trace} for $\Amc$ and $\Omc$ is a
sequence $t = aR_1M_1R_2M_2\dots R_nM_n$, $n \geq 0$ where
$a \in \mn{ind}(\Amc)$, $R_1,\dots,R_n$ are roles that occur in \Omc,
and $M_1,\dots,M_n$ are sets of concept names that occur in \Omc, such
that
\begin{enumerate}[label=(\roman*)]

  \item $a\rightsquigarrow_{\Amc,\Omc}^{R_1}M_1$ and 

  \item for $1 \leq i < n$, we have
    $M_i\rightsquigarrow_{\Omc}^{R_{i+1}}M_{i+1}$ 
 and if $\mn{func}(R_i^-)\in
\Omc$, then $R_{i+1}\neq R_i^-$.

\end{enumerate}
Let $\Tbf$ denote the set of all traces for $\Amc$ and $\Omc$. Then
the \emph{universal model} of \Amc and \Omc is defined as
\begin{align*}
    \Umc_{\Amc, \Omc} = {} & \Amc \cup \{ A(a) \mid \Amc, \Omc \models
                             A(a)\} \cup {}\\
                           & \{A(tRM) \mid tRM \in \Tbf \text{ and }
                             A\in M \} \cup {}\\
			     &\{R(t, tRM) \mid tRM \in \Tbf \}.
\end{align*}

\medskip
In the following, we give three elementary lemmas that pertain
to universal models and are used
throughout the appendix. Their proof is entirely standard and omitted.
The most important properties of universal models are as follows.
\begin{lemma}
    \label{lem:universal-model}
    Let \Omc be an ontology formulated in \DLR or \DLF and let $\Amc$ an ABox that is satisfiable w.r.t.\ \Omc. Then 
    \begin{enumerate}
        \item $\Umc_{\Amc, \Omc}$ is a model of \Amc and \Omc; 
        \item $\Amc, \Omc \models q(\bar a)$ iff $q(\bar x) \to (\Umc_{\Amc, \Omc}, \bar a)$
            for all CQs $q(\bar x)$ and all $\bar a \in 
            \mn{ind}(\Amc)^{|\bar x|}$. 
    \end{enumerate}
\end{lemma}
We note that Lemma~\ref{lem:universal-model} ceases to hold when
\Omc is formulated in \DLRF due to subtle interactions between role
inclusions and functionality assertions. 


The next lemma links query containment to the existence of
homomorphisms into the universal model.
\begin{lemma}
    \label{lem:hom-inclusion}
    Let \Omc be an ontology formulated in \DLR or \DLF, and let $q_1(\bar x)$, $q_2(\bar y)$ be CQs that are satisfiable w.r.t.\ \Omc.
    Then $q_1 \subseteq_\Omc q_2$ iff $q_2(\bar y) \to (\Umc_{q_1, \Omc}, \bar x)$.
\end{lemma}
The following lemma states that any homomorphism from a CQ~$q$ to some
universal model $\Umc_{\Amc, \Omc}$ can be extended to a homomorphism
from $\Umc_{q, \Omc}$ to $\Umc_{\Amc, \Omc}$.
\begin{lemma}
    \label{lem:extend-hom}
    Let $\Omc$ be an ontology formulated in \DLR or \DLF, $\Amc$ an
    ABox, and $q(x_0)$ a unary CQ, such that $\Amc$ and $q$ are both
    satisfiable w.r.t.\ \Omc.  If $h$ is a homomorphism from $q$ to
    $\Umc_{\Amc, \Omc}$ with $h(x_0) = a$ for some
    $a \in \mn{ind}(\Amc)$, then 
    $h$ can be extended to a homomorphism $h'$ from
    $\Umc_{q, \Omc}$ to $\Umc_{\Amc, \Omc}$ with $h'(x_0) = a$.
\end{lemma}

And finally, we show a property of \Omc-minimal and \Omc-saturated queries
that we will use to show that the constructions indeed yield frontiers.  Note
that the property implies that every homomorphism $h$ from an ELIQ $q(x_0)$
to $\Umc_{q, \Omc}$ with $h(x_0) = x_0$ is injective.

\begin{lemma}
    \label{lem:minimal-injective-hom}
    Let \Omc be an ontology in normal form formulated in \DLR or \DLF, and $q(x_0)$ an ELIQ that is
    \Omc-minimal, \Omc-saturated and satisfiable w.r.t.\
    \Omc. Then $\mn{var}(q) \subseteq \mn{img}(h)$
    for every homomorphism $h$ from $q$ to $\Umc_{q, \Omc}$ with
    $h(x_0) = x_0$.
\end{lemma}
\noindent
\begin{proof}\
    Assume for contradiction that there is a variable 
    $x \in \mn{var}(q)$ with $x \notin \mn{img}(h)$. Let $q'$
    be the restriction of $q$ to $\mn{var}(q) \setminus \mn{var}(q_x)$.  We
    show that $h$ is also a homomorphism from $q$ to
    $\Umc_{q', \Omc}$, and thus that $q \equiv_\Omc q'$. This contradicts the
    minimality of $q$.

    First, observe that for all $y \in \mn{var}(q)$, $h(y) \notin
    \mn{var}(q_x)$, since $q$ is connected and there is no $y' \in
    \mn{var}(q)$ with $h(y') = x$.
    
    Next let $y R_1 M_1 \dots R_n M_n \in \Delta^{\Umc_{q, \Omc}}$ be a trace
    starting with some variable $y \in \mn{var}(q')$. Then $y
    \rightsquigarrow^{R_1}_{q, \Omc} M_1$ and therefore $\Amc_q, \Omc \models
    \exists R_1. \bigsqcap M_1(y)$ and there is no $R_1(y, y') \in q$ such
    that $\Amc_q, \Omc \models \bigsqcap M_1(y')$. Since $\Omc$ is in normal
    form, there is a set of concept names $M$ such that $\Amc_q, \Omc \models
    \bigsqcap M(y)$ and $\Omc \models \bigsqcap M \sqsubseteq \exists R_1.
    \bigsqcap M_1(y)$. By \Omc-saturation of $q$, $\Amc_{q'}, \Omc \models
    \bigsqcap M(y)$ and therefore $\Amc_{q'}, \Omc \models \exists R_1.
    \bigsqcap M_1(y)$. Since $q'$ is a subset of $q$, $y
    \rightsquigarrow^{R_1}_{q', \Omc} M_1$ and thus $y R_1 M_1 \dots R_n M_n
    \in \Delta^{\Umc_{q', \Omc}}$.
    
    Let $A(y) \in q$. If $h(y) \in \mn{var}(q')$, then $A(h(y)) \in \Umc_{q',
    \Omc}$ by \Omc-saturation of $q$.  If $h(y)$ is a trace, then, by
    connectedness it is a trace below a variable of $q'$ and thus $A(h(y)) \in
    \Umc_{q', \Omc}$.

    Let $r(y, y') \in q$.  Again, by connectedness of $q$, $h(y)$ and $h(y')$
    must be variables of $q'$ or traces starting with variables of $q'$.  If
    both $h(y)$ and $h(y')$ are variables, then
    $r(h(y), h(y')) \in \Umc_{q', \Omc}$, since $h(y), h(y') \notin
    \mn{var}(q_{x})$.
    If one or both of
    $h(y)$ and $h(y')$ are a trace, then since the trace-subtrees
    are identical, also $r(h(x_1), h(x_2)) \in \Umc_{q', \Omc}$.

    Thus, $h$ is a homomorphism from $q$ to $\Umc_{q', \Omc}$ as required.
\end{proof}

\section{Proofs for Section~\ref{sec:prelims}}

We describe how to convert a \DLRF-ontology \Omc into a
\DLRF-ontology $\Omc'$ in normal form. We use $\Cmf(\Omc)$ to
denote the set of all concepts that occur on the right-hand side of a
concept inclusion in \Omc. Note that $\Cmf(\Omc)$ is closed under
sub-concepts.  We introduce a fresh concept name $X_C$ for every
complex concept $C \in \Cmf(\Omc)$, and set 
$X_A=A$ for concept names $A\in\Cmf(\Omc)$.
%
The ontology $\Omc'$ consists of the
following concept and role inclusions:
  \begin{itemize}

  \item all role inclusions from $\Omc$;

  \item $C \sqsubseteq X_D$ for every 
    $C \sqsubseteq D \in \Omc$;

  \item $X_{D_1\sqcap D_2} \sqsubseteq X_{D_i}$, for 
    every $D_1\sqcap D_2\in \Cmf(\Omc)$ and $i\in\{1,2\}$;

  \item $X_{\exists R.C}\sqsubseteq \exists R.X_C$, for
    every $\exists R.C\in \Cmf(\Omc)$;

%
%


%

\end{itemize}
Clearly, $\Omc'$ can be computed in polynomial time. Moreover, it is
easy to verify that for all $C \in \Cmf(\Omc)$, we have
$\Omc' \models X_{C} \sqsubseteq C$.
Regarding
the relationship between \Omc and $\Omc'$, we observe the following
consequences of the definition of $\Omc'$.
\begin{lemma}
    \label{lem:normalform}
    ~\\[-5mm]
    \begin{enumerate}


      \item $\Omc' \models \Omc$, that is, every model of $\Omc'$ is a
        model of $\Omc$;

      \item every model $\Imc$ of $\Omc$ can be extended
        into a model $\Imc'$ of $\Omc'$ by starting with $\Imc'=\Imc$
        and then setting for every complex concept $C \in \Cmf(\Omc)$,
	$X_C^{\Imc'}= C^\Imc$.
%
%
%
        
    \end{enumerate} \end{lemma}
  Lemma~\ref{lem:normalform} essentially says that $\Omc'$ is a
  conservative extension of $\Omc$, but is slightly stronger in 
  also making precise how exactly a model of $\Omc$ can be extended
  to a model of $\Omc'$.
  
\section{Proofs for Section~\ref{sect:frontier}}

\lemfrontiernormalformR*
\noindent
\begin{proof}\ Let \Omc be a \DLR ontology and let $\Omc'$ be the
  result of converting $\Omc$ into normal form as described before
  Lemma~\ref{lem:normalform}. Moreover, let $q(x)$ be an ELIQ and
  $\Fmc'$ a frontier of $q$ w.r.t.\ $\Omc'$. Let $\Fmc$ be obtained from
  $\Fmc'$ by including all ELIQs that can be obtained by taking an ELIQ
  $p \in \Fmc'$ and then doing the following:
    \begin{itemize}

    \item for every atom $X_{C}(x)$, $C\in\Cmf(\Omc)$, remove that
      atom and add a variable disjoint
      copy of $C$ viewed as an ELIQ, gluing the root to $x$;

%

    \end{itemize}
    To prove that $\Fmc$ is a frontier of $q$ w.r.t.\ \Omc, we show
    that the three conditions from the definition of frontiers are satisfied:
    \begin{enumerate} 

    \item $q \subseteq_\Omc q_F$ for all $q_F \in \Fmc$.

      Assume to the contrary that there is a $q_F(x)\in \Fmc$ with
      $q\not\subseteq_\Omc q_F$. Then, there is an ABox \Amc and an
      individual $a\in\mn{ind}(\Amc)$ such that
      $\Amc,\Omc\models q(a)$, but $\Amc,\Omc\not\models q_F(a)$.  We
      may assume that concept names $X_{C}$ do not occur in \Amc as they are
      not used in \Omc, $q$, and $q_F$.  From $\Omc' \models \Omc$, we
      obtain $\Amc,\Omc'\models q(a)$.  Since
      $\Amc,\Omc\not\models q_F(a)$, there is a model $\Imc$ of \Amc
      and \Omc such that $\Imc\not\models q_F(a)$. Let $\Imc'$ be the
      extension of \Imc according to Point~2 of
      Lemma~\ref{lem:normalform}. Then $\Imc'$ is a model of $\Omc'$
      and \Amc. Now, let $q^0_F(x)\in \Fmc'$ be the query from which
      $q_F(x)$ was obtained in the construction of \Fmc. By
      construction of $q_F$ and of $\Imc'$,
      $\Imc'\not\models q^0_F(a)$, so
      $\Amc,\Omc' \not\models q^0_F(a)$. Thus, \Amc witnesses that
      $q\not\subseteq_{\Omc'} q^0_F$, a contradiction to $q^0_F$ being
      in $\Fmc'$.

      \item $q_F \not\subseteq q$ for all $q_F \in \Fmc$.

        Let $q_F(x) \in \Fmc$ and let $q^0_F(x)\in \Fmc'$ be the ELIQ
        from which $q_F(x)$ was obtained during the construction of
        \Fmc. Since
        $q^0_F\not\subseteq_{\Omc'} q$, there is an ABox $\Amc'$
        and an individual $a \in \mn{ind}(\Amc')$ such that
        $\Amc', \Omc' \models q^F_0(a)$, but
        $\Amc', \Omc' \not\models q(a)$. Let the ABox $\Amc$
        be obtained by starting with $\Amc'$ and 
	adding $C(b)$, for each concept assertion
	$X_{C}(b)\in \Amc'$.
    Here, the addition of $C(b)$, with $C$ an \ELI-concept, is defined
    as expected: view $C(b)$ as a tree-shaped ABox $\Amc_{C(b)}$
    that uses only fresh individual names, and then add this ABox gluing
    its root to $b$.

    We aim to show $\Amc,\Omc \models q_F(a)$ and
    $\Amc,\Omc \not\models q(a)$, witnessing $q_F \not\subseteq_\Omc q$ as
    required. 

    For the former, assume to the contrary that
    $\Amc,\Omc \not\models q_F(a)$. Then there is a model \Imc of \Amc
    and \Omc with $\Imc \not\models q_F(a)$. Let $\Imc'$ be the
    extension of \Imc according to Point~2 of
    Lemma~\ref{lem:normalform}. Then $\Imc'$ is a model of $\Omc'$
    and, by construction of $\Imc'$ and of $\Amc$ from $\Amc'$, also a
    model of $\Amc'$. Moreover, $\Imc \not\models q_F(a)$ implies
    $\Imc' \not\models q^0_F(a)$ by construction of $q_F$ and of
    $\Imc'$. This contradicts $\Amc', \Omc' \models q^F_0(a)$.

    It remains to show that $\Amc,\Omc \not\models q(a)$. Since
    $\Amc', \Omc' \not\models q(a)$, there is a model $\Imc$ of $\Omc'$
    and $\Amc'$ such that $\Imc \not\models q(a)$. Since
    $\Omc' \models \Omc$, \Imc is also a model of \Omc. Since
    $\Omc' \models X_{\exists R . C} \sqsubseteq \exists R . C$ for all
    $\exists R.C \in \Cmf(\Omc)$ and due to the construction of
    $\Amc$ from $\Amc'$, $\Imc$ is also a model of $\Amc$.
    Thus, \Imc witnesses $\Amc,\Omc \not\models q(a)$, as required.

  \item For all ELIQs $q'$ with $q \subseteq_\Omc q'
    \not\subseteq_\Omc q$, there is a $q_F \in \Fmc$ with $q_F
    \subseteq_\Omc q'$.

    Let $q'$ be an ELIQ with $q\subseteq_\Omc q'\not\subseteq_{\Omc}
    q$. From $q\subseteq_\Omc q'$, it follows that $q'$ does not use
    the fresh concept names $X_{C}(x)$ in $\Omc'$. Consequently, we obtain from
    Lemma~\ref{lem:normalform} that $q\subseteq_{\Omc'}
    q'\not\subseteq_{\Omc'} q$. There is thus a $q^0_F(x)\in \Fmc'$
    with $q^0_F \subseteq_{\Omc'} q'$. Assume that $q_F \in \Fmc$ was
    obtained from $q^0_F$ in the construction of \Fmc. It suffices to
    show that $q_F \subseteq_\Omc q'$. Assume to the contrary that
    this is not the case. Then there is an ABox \Amc and an individual
    $a \in \mn{ind}(\Amc)$ such that $\Amc,\Omc \models q_F(a)$, but
    $\Amc,\Omc\not \models q'(a)$. We can proceed as in Point~1 above
    to show that $\Amc,\Omc' \models q_F(a)$ and $\Amc,\Omc'\not
    \models q'(a)$, in contradiction to $q^0_F \subseteq_{\Omc'} q'$.
        
    \end{enumerate} %

    It remains to prove the second part of the Lemma: every
    $\Omc$-minimal ELIQ $q$ is also $\Omc'$-minimal. Suppose that
    $q$ is not $\Omc'$-minimal, that is, there is a variable
    $x$ such that $q\equiv_{\Omc'}q'$ where $q'=q|_{\mn{var}(q)\setminus\{x\}}$.
    Clearly, we have $q\subseteq_\Omc q'$ as $q'\subseteq q$.
    Moreover, since $q'\subseteq_{\Omc'} q$ and Point~1 of
    Lemma~\ref{lem:normalform}, we have $q'\subseteq_\Omc q$. Hence,
    $q\equiv_{\Omc} q'$ and $q$ is not $\Omc$-minimal.
\end{proof}

\lemfrontiermainh*
\noindent
\begin{proof}\
We show that $\Fmc$ fulfills the three conditions of frontiers. For
Condition~1, let $p(x_0)$ be a query from $\Fmc$. Then, since $q$ is
satisfiable w.r.t.~\Omc, so is $p$.
Hence it suffices to show $p(x_0) \to (\Umc_{q, \Omc}, x_0)$ by
Lemma~\ref{lem:hom-inclusion}. 

We extend the mapping $\cdot^\downarrow$ to be defined on all variables of $p$
by considering the yet unmapped variables added in Step~2A of the
construction.
Let $z$ be such a fresh variable added for $x \in \mn{var}(p)$, roles $R$,
$S$ and concept name $A$.
Then $x^\downarrow \rightsquigarrow^R_{q, \Omc} A$ and by construction
of $\Umc_{q, \Omc}$, there is a trace $x^\downarrow R A \in \Delta^{\Umc_{q, \Omc}}$. Set $z^\downarrow = x^\downarrow R A$.
Now $\cdot^\downarrow$ is defined on all variables of $p$ and, by 
construction of $p$, it is a homomorphism from $p$ to $\Umc_{q, \Omc}$ with
$x_0^\downarrow = x_0$ as required.

\medskip
We start the proof of the second condition of frontiers with the following
claim:

\smallskip
\noindent
\textit{Claim~1.} $p\not\subseteq_\Omc q_x$ for all $x \in
\mn{var}(q)$ and $p(x) \in \Fmc_0(x)$.

\smallskip
\noindent
\textit{Proof of Claim~1.}
We show the claim by induction on the codepth of $x$ in $q$, matching the
inductive construction of $\Fmc_0$.
In the induction start, $x$ has codepth $0$. Then, by definition of codepth,
there is no $R(x, y) \in q$ that is
directed away from $x_0$ and all $p \in \Fmc_0(x)$ must be obtained by dropping a
concept atom.

Let $p(x)$ be a query from $\Fmc_0(x)$ that is obtained by dropping the
concept atom $A(x) \in q$.
Then, by choice of $A(x)$, there is no $B(x) \in p$ with $\Omc \models B
\sqsubseteq A$ and no $R(x, x') \in p$ with $\Omc \models \exists R \sqsubseteq
A$. Hence, $A(x) \in q_x$ and $A(x) \notin \Umc_{p, \Omc}$, therefore $p
\not \subseteq_\Omc q_x$.

In the induction step, let $x$ have codepth $>0$, let $p(x)$ be a query from
$\Fmc_0(x)$ and assume that the claim holds for all variables with smaller
codepth.
Let $\cdot^\downarrow$ be the extension of the original $\cdot^\downarrow$ for
$p$ to a homomorphism from $\Umc_{p, \Omc}$ to $\Umc_{q_x, \Omc}$ that exists
by Lemma~\ref{lem:extend-hom}.
If $p$ is obtained by dropping a concept atom, then the same argument as in
the induction start yields $p \not\subseteq_\Omc q_x$.
If $p$ is obtained by generalizing a subquery attached to a role atom $R(x, y) \in q_x$, assume for
contradiction that there is a homomorphism $h$ from $q_x$ to $\Umc_{p, \Omc}$
with $h(x) = x$. From $h$ we construct a homomorphism $h'$ from $q$ to
$\Umc_{q, \Omc}$ with $h'(x_0) = x_0$ 
by setting $h'(z) = h(z)^\downarrow$ for all $z \in \mn{var}(q_y)$ and $h'(z) = z$
for all $z \notin \mn{var}(q_y)$.
Since $h$ must map $y$ to an $R$-successor of $x$ in $\Umc_{p, \Omc}$,
we may distinguish the following cases.
\begin{itemize}
\item $h(y)$ is a $z \in \mn{var}(p)$ with $z^\downarrow \neq y$.

    Then, by
    definition of $\cdot^\downarrow$, $z^\downarrow = z$, and therefore, by
    construction of $h'$, $h'(y) = h'(z) = z$.
    Hence $h'$ is a non-injective homomorphism from $q$ to $\Umc_{q, \Omc}$
    with $h'(x_0) = x_0$, contradicting \Omc-minimality or \Omc-saturatedness
    of $q$ by Lemma~\ref{lem:minimal-injective-hom}.

\item $h(y)$ is a trace $h(x) S A \in \Delta^{\Umc_{p, \Omc}}$ for some role $S$ with
    $\Omc \models S \sqsubseteq R$ and concept name $A$.
    
    Then, if $h'(y)$ is also a trace, there must be a $y' \in \mn{var}(q)$ with
    $y' \notin \mn{img}(h')$, again contradicting \Omc-minimality or
    \Omc-saturatedness of $q$ by Lemma~\ref{lem:minimal-injective-hom}.

    If $h'(y)$ is not a trace, but a successor $y'$ of $x$ with $y' \neq y$,
    then $h'(y') = h'(y)$, again contradicting \Omc-minimality of $q$ by
    Lemma~\ref{lem:minimal-injective-hom}.

    If $h'(y) = y$ and there is a $y' \in \mn{var}(q_y)$ with $h'(y') = x$,
    then $h'(y') = h'(x)$, again contradicting $\Omc$-minimality by
    Lemma~\ref{lem:minimal-injective-hom}.

    If $h'(y) = y$ and there is no $y' \in \mn{var}(q_y)$ with $h'(y') = x$,
    then we show a contradiction to \Omc-minimality $q$ by constructing a
    homomorphism $h''$ from $q$ to $\Umc_{q', \Omc}$ with $h''(x_0) = x_0$
    where $q'$ is the restriction of $q$ to $\mn{var}(q) \setminus\{y\}$.
    Note that by construction of $h'$, $h'(z) = y$ implies $z = y$.

    Since $h(x)SA^\downarrow = y$, there is no trace $xSA \in \Delta^{\Umc_{q,
    \Omc}}$. But, since $h(x)SA \in \Delta^{\Umc_{p, \Omc}}$ it must be that
    $h(x) \rightsquigarrow^S_{p, \Omc} A$ and thus $\Amc_p, \Omc \models
    \exists R.A(h(x))$ and $\Amc_q, \Omc \models \exists R.A(x)$. However,
    $x \not\rightsquigarrow^S_{p, \Omc} A$ because $\Omc \models R \equiv S$,
    $R(x, y) \in q$ and $\Amc_q, \Omc \models A(y)$.

    Since $R(x, y) \notin q'$ and by \Omc-saturation of $q$ and normal form of
    $\Omc$, it follows that $x \rightsquigarrow^S_{q', \Omc} A$ and therefore
    there is a trace $x S A \in \Delta^{\Umc_{q', \Omc}}$.

    Construct $h''$ by setting $h''(z) = h'(z)$ for all $z \in \mn{var}(q)
    \setminus \mn{var}(q_y)$ and $h''(z) = x S A R_2 A_2 \dots R_n A_n$ for
    all $z \in \mn{var}(q_y)$ if $h(z) = x S A R_2 A_2 \dots R_n A_n$.

\item $h(y)$ is the root $y'$ of a $p' \in \Fmc_0(y)$ that was added in
  Point~3 of generalizing a subquery.
  
  Then, by the induction hypothesis, $q_y \not\to \Umc_{p', \Omc}$ for
  all $p' \in \Fmc_0(y)$. We argue that, consequently, $h$ cannot map
  $q_y$ entirely into the subtree below $y'$ in $\Umc_{p, \Omc}$. To
  show this, it clearly suffices to argue that this subtree is
  isomorphic to $\Umc_{p', \Omc}$. In fact, this is the case by definition of
  universal models, \Omc-saturation of $p$ and since \Omc is in normal form,
  unless there is a concept name $A$ with $\Omc \models \exists R^-
  \sqsubseteq A$ such that $A(y') \in \Umc_{p, \Omc}$, but $A(y') \notin
  \Umc_{p', \Omc}$.
  Because of \Omc-saturatedness, this implies $A(y') \in p$ and
  $A(y') \notin p'$. Since the construction of $\Fmc_0$ never adds
  concept names to an ELIQ, this implies that
  $A({y'}^\downarrow) \in q$.  Thus, $A({y'}^\downarrow)$ was dropped
  during the construction of $p'$ from $q_{y'^\downarrow}$.
  This may only happen by dropping a concept atom. However,
  $R(x,y) \in q$ and   $\Omc \models \exists R^- \sqsubseteq A$
  contradicts Condition (b)~of dropping a concept atom .

  We have thus shown that
  $h$ cannot map $q_y$ entirely into the subtree below
  $y' \in \Umc_{p', \Omc}$. Consequently, there must be a $y'' \in \mn{var}(q_y)$
  with $h(y'') = x$.  This yields $h'(y') = h'(x) = x$ contradicting
  \Omc-minimality of $q$ by Lemma~\ref{lem:minimal-injective-hom}.
\end{itemize}
Note that $h(y)$ cannot be a variable introduced in Point~4 of generalizing a
subquery, as there only $S$-successors of $x$ are introduced that
satisfy $\Omc \not\models S \sqsubseteq R$.
This completes the proof of Claim~1.

\smallskip
We continue by using Claim~1 to show that $p \not\subseteq_\Omc q$ for all $p
\in \Fmc$.
Let $p$ be a query from \Fmc and assume for contradiction that $p
\subseteq_\Omc q$. Then, there is a homomorphism $h$ from $q$ to $\Umc_{p,
\Omc}$ with $h(x_0) = x_0$. 
Let $\cdot^\downarrow$ be the extension of the original $\cdot^\downarrow$ for
$p$ to a homomorphism from $\Umc_{p, \Omc}$ to $\Umc_{q, \Omc}$ which exists
by Lemma~\ref{lem:extend-hom}.
We compose $h$ and $\cdot^\downarrow$ to construct a
homomorphism $h'$ from $q$ to $\Umc_{q, \Omc}$
with $h'(x_0) = x_0$.
By Claim~1, there is no homomorphism that maps $q$ entirely into $\Umc_{p',
\Omc}$ for any $p' \in \Fmc_0(x_0)$. Hence, there must be an $x \in \mn{var}(q)$
such that $h(x)$ is a fresh variable added in the compensation
step. By definition of that step and since $q$ is connected, we may
distinguish the following two cases:
\begin{itemize}

\item 
$h(x)$ is a fresh variable added in Step~2A. 
Then by definition of $\cdot^\downarrow$, $h'(x)$ is a trace, contradicting
\Omc-minimality of $q$ by Lemma~\ref{lem:minimal-injective-hom}.

\item $h(x)$ is a fresh variable $z$ added in Step~2B for the
role atom $S(y, y') \in p$ with $z^\downarrow = y^\downarrow$. 

Then, since $q$
is connected, there must be a predecessor $x'$ of $x$ with $h(x') = y$. Hence
$h'(x) = h'(x') = y^\downarrow$, contradicting \Omc-minimality of $q$ by
Lemma~\ref{lem:minimal-injective-hom}.

\end{itemize} 
This completes the proof of Condition~2 of frontiers. 

\medskip

It remains to show that Condition~3 of frontiers is satisfied.  Let
$q'(x_0)$ be an ELIQ that is satisfiable w.r.t.\ \Omc such that
$q \subseteq_\Omc q' \not \subseteq_\Omc q$. We may assume w.l.o.g.\
that $q'$ is \Omc-saturated.  

There is a homomorphism $g$ from $q'$ to $\Umc_{q, \Omc}$ with $g(x_0) = x_0$.
We have to show that there is a $p \in \Fmc$ with
$p \subseteq_\Omc q'$. To do this, we construct in three steps a
homomorphism $h$ from $q'$ to $\Umc_{p, \Omc}$ with $h(x_0) = x_0$ for
some $p \in \Fmc$.  During all steps, we maintain the
invariant
\begin{equation*}  \label{eqn:h-h}
    h(z)^\downarrow = g(z) \tag{$*$}
\end{equation*}
for all variables $z \in \mn{var}(q')$ with $h(z)$ defined
and $\cdot^\downarrow$ the extension of the original $\cdot^\downarrow$ for $p$ to a homomorphism from $\Umc_{p, \Omc}$ to $\Umc_{q, \Omc}$.
In the first step of the construction, we define $h$ for an initial
segment of~$q'$.

Let $U \subseteq \mn{var}(q')$ be the smallest set of variables
(w.r.t.~$\subseteq$) of $q'$ such that $x_0 \in U$ and $R(x, y) \in q'$ with $x
\in U$, $g(y) \in \mn{var}(q)$, and $S(g(x), g(y)) \in q$ directed away from
$x_0$ implies $y \in U$.  Let $q^U$ be the
restriction of $q'$ to the variables in $U$. 

\smallskip
\noindent
\textit{Claim~2.}  For all $x \in U$ with
$q^U_x \not \subseteq_\Omc q_{g(x)}$, there is a $p \in \Fmc_0(g(x))$
and a homomorphism $h'$ from $q^U_x$ to
$\Umc_{p, \Omc}$ that satisfies \eqref{eqn:h-h}.

\smallskip
\noindent
\textit{Proof of Claim~2.}
Let $y = g(x)$.
We show Claim~2 by induction on the codepth of $x$ in $q^U$.  In the induction
start, $x$ has codepth $0$. We distinguish the following cases:
\begin{itemize}

\item There is an $R(y, y') \in q_y$.

  Then let $p \in \Fmc_0(y)$ be constructed by generalizing the subquery attached to the role atom
  $R(y, y')$ and set $h'(x) = y$. Since $q$ is \Omc-saturated,
  $A(y)\in \Umc_{q,\Omc}$ implies   $A(y)\in \Umc_{p,\Omc}$.
  This and  $y = g(x)$ implies that $h'$ is a homomorphism.

\item There is no $R(y, y') \in q_y$.

  Then $q^U_x \not \subseteq_\Omc q_{y}$ implies that there is an $A(y) \in q_y$ with
  $A(x) \notin \Umc_{q^U_x, \Omc}$, and we must even find an $A$ with
  these properties such that there is no $B(y) \in q_y$ with
  $\Omc \models B \sqsubseteq A$ and
  $\Omc \not\models A \sqsubseteq B$. This implies that Property~(a)
  of dropping concept atoms is satisfied. Property~(b) is satisfied
  since there is no $R(y, y') \in q_y$ and thus we may construct
  $p \in \Fmc_0(y)$ by dropping the concept atom $A(y)$. Set
  $h'(x) = y$.

\end{itemize}
In the induction step, let $x$ have codepth $>0$ in $q^U$ and assume that the
claim holds for all variables of smaller codepth. From $q^U_x \not
\subseteq_\Omc q_{y}$, it follows that $q_{y} \not\to (\Umc_{q^U_{x},
    \Omc}, x)$. We distinguish the following
cases:
\begin{itemize}

\item There is an $R(y, y') \in q_y$ such that $q_{y'} \not\to (\Umc_{q^U_{x'},
    \Omc}, x')$ for all $S(x, x') \in q^U_x$ with $\Omc \models S \sqsubseteq R$.
  
Let $p \in \Fmc_0(y)$ be constructed by
generalizing the subquery attached to the role atom $R(y, y')$.
We construct the homomorphism $h'$ from $q^U_x$ to $\Umc_{p, \Omc}$ by starting with
$h'(x) = y$ and continuing to map all successors of~$x$.
Let $S(x, x') \in q^U_x$.

If $g(x') \neq y'$, then extend $h'$ to the subtree below $x'$ 
by setting $h'(z) = g(z)$ for all $z \in \mn{var}(q_{x'})$.

If $g(x') = y'$ and $\Omc \models S \equiv R$, then, by the induction hypothesis,
there is a $p' \in \Fmc_0(y')$ and a homomorphism $h''$ from $q_{x'}$ to $\Umc_{p', \Omc}$ with $h''(x') = y'$.
Extend $h'$ to the variables in $q'_{x'}$ by mapping $q'_{x'}$ according to $h''$ to the copy of
$p'$ that was attached to $y$ in Point~3 of generalizing a subquery.

If $g(x') = y'$ and $\Omc \not\models S \sqsubseteq R$, then extend $h'$ to the
variables in $q'_{x'}$ by mapping $q'_{x'}$ according to $g$ to the copy of $q_{y'}$
that was added in Point~4 of generalizing the subquery attached with the role $S$.

\item For every $R(y, y') \in q_y$, $q_{y'} \to (\Umc_{q^U_{x'},
    \Omc}, x')$ for some $S(x, x') \in q^U_x$ with $\Omc \models S \sqsubseteq R$.

  Then there is an $A(y) \in q_y$ with
  $A(x) \notin \Umc_{q^U_x, \Omc}$ and we
  must even find an $A$ with these properties such that there is
  no $B(y) \in q_y$ with $\Omc \models B \sqsubseteq A$ and
  \mbox{$\Omc \not\models A \sqsubseteq B$}. Thus, Property~(a) of dropping
  concept atoms is satisfied. To show that Property~(b) is also
  satisfied, we have to argue that there is no $R(y, y') \in q_y$ with
  $\Omc \models \exists R \sqsubseteq A$. But if there is such
  an $R(y,y') \in q_y$, then $q_{y'} \to (\Umc_{q^U_{x'},
    \Omc}, x')$ for some $S(x, x') \in q^U_x$ with $\Omc \models S \sqsubseteq R$. This implies 
  $A(x) \in \Umc_{q^U_x, \Omc}$, a contradiction.
  
We may thus construct $p \in \Fmc_0(y)$ 
by dropping the concept atom $A(y)$. Set $h'(x') = g(x')$ for all $x' \in
\mn{var}(q^U_x)$.

\end{itemize} 
This completes the proof of Claim~2.

\smallskip By Claim~2, there is a $p' \in \Fmc_0(x_0)$ such that
$q^U \to \Umc_{p', \Omc}$. Let $p \in \Fmc$ be the query that was
obtained by applying the compensation step to $p'$. Then clearly also
$q^U \to \Umc_{p, \Omc}$. Define $h$ for all variables in $U$
according to the homomorphism that witnesses this.
Let $\cdot^\downarrow$ be the extension of the original $\cdot^\downarrow$ for
$p$ to a homomorphism from $\Umc_{p, \Omc}$ to $\Umc_{q, \Omc}$ which exists
by Lemma~\ref{lem:extend-hom}.

\smallskip

We continue with the second step of the construction of~$h$ which
covers subtrees of $q'$ that are connected to the initial segment $q^U$
and whose root is mapped by $g$ to traces of $\Umc_{q,\Omc}$ (and
as opposed to a variable from
$\mn{var}(q)$).  Consider all atoms $R(x, x') \in q'$ with $h(x)$
defined, $h(x')$ undefined and $g(x') \notin \mn{var}(q)$.
Before extending $h$ to $q'_{x'}$, we first show that there is an atom $S(h(x), z)
\in p$ with $\Omc \models S \sqsubseteq R$, added in Step~2A.

Since $g(x') \notin \mn{var}(q)$, $g(x')$ must be a trace $g(x) S A \in
\Delta^{\Umc_{q, \Omc}}$ for some concept name $A$ and role $S$ with $\Omc
\models S \sqsubseteq R$. Hence, $g(x) \rightsquigarrow^S_{q, \Omc}
A$.

We aim to show that Step~2A of compensation is applicable. To thus
end, take any concept name
$B$ such that $\Omc
\models \exists R
\sqsubseteq B$. We have to show that $B(h(x)) \in p$. Assume to the
contrary that $B(h(x)) \notin p$. Then, since $q$ is $\Omc$-saturated,
$B(g(x)) \in q$ and $p$ must be the result of dropping the concept atom $B(g(x))$.
However, since $q'$ is \Omc-saturated and $R(x, x') \in q'$, $B(x) \in q'$ and this contradicts Claim~2, since $x \in U$.

Hence, Step~2A adds the atoms $R(h(x), z), B(z), S(z', z)$ with $z$
and $z'$ fresh variables and adds a disjoint copy $\widehat q$ of $q$,
gluing the copy of $h(x)^\downarrow$  in $\widehat q$ to $z'$.  Extend $h$ to the
variables in $q'_{x'}$ by setting
$h(\hat x) = z R_2 M_2 \dots R_n M_n$ if
$g(\hat x) = g(x) S A R_2 M_2 \dots R_n M_n$ for all $\hat x$ in the
subtree below $x'$.  If there is an $x'' \in \mn{var}(q'_{x'})$ with
$g(x'') = g(x)$, set $h(x'') = z'$ and continue mapping the subtree
below $x''$ into the attached copy $\widehat q$ of $q$ according to $g$.

\smallskip

In the third and final step of the construction of $h$, we consider
the remaining subtrees of $q'$.  Let $R(x, x') \in q'$ be directed
away from $x_0$ with $h(x)$ defined and $h(x')$ undefined. 

Then $h(x)$ was defined in the first step of the construction of $h$, and thus
$x \in U$. As $h(x')$ was not defined in the second step $g(x') \in
\mn{var}(q)$.
Therefore, since $x' \notin U$, $R(g(x), g(x')) \in \Umc_{q, \Omc}$ must be directed towards
$x_0$. This implies that $x$ is not the root of $q'$ and that there is an atom
$T(x'', x) \in q'$ directed away from $x_0$ with $g(x'') = g(x')$. From $x \in
U$ it follows that $x'' \in U$ and therefore $h(x'')$ and $h(x)$ were defined
in the first step of the construction of $h$.

Hence, there is an atom $S(h(x''), h(x)) \in p$ directed away from $x_0$ with
$\Omc \models S \sqsubseteq T$ that was not added in Step~2A. Since by
\eqref{eqn:h-h} $h(x'')^\downarrow = g(x'') = g(x')$ and $h(x)^\downarrow =
g(x)$, $\Amc_q, \Omc \models R^-(h(x'')^\downarrow, h(x)^\downarrow)$

Therefore, Step~2B added an atom $R^-(z, h(x))$ to $p$, $z$ a fresh variable,
and glued a copy $\widehat q$ of $q$ to $z$.  Set $h(x') = z$ and extend $h$
to the entire subtree $q'_{x'}$ by mapping all variables into the attached
copy of $q$ according to $g$.

This completes the construction of $h$ and the proof that Condition 3
is satisfied.
\end{proof}

\lemfrontiersize*

\noindent \begin{proof}\ In order to reduce notational clutter, we
  introduce some abbreviations used throughout the proof. 
  \begin{itemize}

    \item $s=|\mn{sig}(q)|$ denotes the number of concept and role
      names used in $q$;

    \item $o=||\Omc||$ denotes the size 
      of \Omc;

    \item for an ELIQ $p$, $n_p=|\mn{var}(p)|$ denotes the number of
      variables in $p$;

    \item for a set $Q$ of queries, $n_Q$ denotes $\sum_{p\in Q} n_p$.

  \end{itemize}
  We assume without loss of generality that $s$ and $o$ are at least
  one.

  We start with analyzing the size of the queries in $\Fmc_0(x)$ that
  are obtained as the result of the generalization step. 

  \smallskip\noindent\textit{Claim.} For every $x\in
  \mn{var}(q)$, 
  \[\displaystyle n_{\Fmc_0(x)} \leq s
  \cdot o \cdot n_{q_x}^3.\]

  \smallskip\noindent\textit{Proof of the claim.} The proof is by
  induction on the codepth of $x$ in $q$.  For the base case, consider
  a variable $x$ of codepth $0$ in $q$, that is, a leaf. In this case,
  only Step~(A) is applicable, and it adds at most $s$ queries to
  $\Fmc_0(x)$, each with a single variable. 

  For the inductive step, consider a variable $x$ of codepth greater
  than $0$. We partition $\Fmc_0(x)$ into $\Fmc_0^{A}(x)$ and $\Fmc_0^{B}(x)$,
  that is, the queries that are obtained by dropping a concept atom in
  Step~(A) and the queries that are obtained by generalizing a subquery in
  Step~(B), respectively, and analyze them separately, starting with
  $\Fmc_0^{A}(x)$. Clearly, every $p\in \Fmc_0^A(x)$ uses $n_{q_x}$
  variables and there are at most $s$ queries in $\Fmc_0^A$. Thus, we
  have
  \[n_{\Fmc_0^A(x)} \leq s\cdot n_{q_x}.\]
  Next, we analyze $\Fmc_0^{B}(x)$. Each query in $\Fmc_0^{B}(x)$ is
  obtained by first picking, in Point~1, an atom $R(x,y)$ in $q_x$. Then, in Point~3, we add
  $\sum_{p\in \Fmc_0(y)} n_p$ variables and, in Point~4, we add at
  most $o$ copies of $q_y$. Thus, we obtain
  \[n_{\Fmc_0^B(x)}\leq \sum_{R(x,y)\in q_x} (n_{q_x}+ n_{\Fmc_0(y)} +
  o\cdot n_{q_y}).\]
  Plugging in the
  induction hypothesis, we obtain 
  \begin{equation}\label{eq:fb}
    n_{\Fmc_0^B(x)}\leq \sum_{R(x,y)\in
    q_x} (n_{q_x}+s\cdot o\cdot  n_{q_y}^3 + o\cdot n_{q_y}).
  \end{equation}
  We simplify the right-hand side of~\eqref{eq:fb} by making the
  following observations:
  \begin{itemize}

    \item $\sum_{R(x,y)\in q_x} n_{q_x}\leq n_{q_x}\cdot (n_{q_x}-1)$, 

    \item $\sum_{R(x,y)\in q_x} n_{q_y} = n_{q_x}-1$,
      and

    \item $\sum_{R(x,y)\in q_x} n_{q_y}^3\leq \left(\sum_{R(x,y)\in
      q_x} n_{q_y}\right)^3=(n_{q_x}-1)^3$. Here, the inequality is an
      application of the general inequality $\sum_{i}a_i^3\leq
      \left(\sum_{i}a_i\right)^3$, for every sequence of non-negative
      numbers $a_1,\ldots,a_k$.

  \end{itemize}
  Using these observations, Inequality~\eqref{eq:fb} can be simplified to:  
  \begin{align*}
    n_{\Fmc_0^B(x)} & \leq n_{q_x}\cdot (n_{q_x}-1) + s\cdot o\cdot (n_{q_x}-1)^3  + o\cdot
    (n_{q_x}-1)\\
    & \leq s\cdot o\cdot \left(n_{q_x}\cdot (n_{q_x}-1)+ 
    (n_{q_x}-1)^3 +
    (n_{q_x}-1) \right)\\
    & = s\cdot o\cdot \left( n_{q_x}^2 + (n_{q_{x}}-1)^3 - 1 \right).
   \end{align*}
  Overall, we get 
  \begin{align*}
    n_{\Fmc_0(x)} & = n_{\Fmc_0^A(x)} + n_{\Fmc_0^B(x)} \\
    & \leq s\cdot n_{q_x} + s\cdot o\cdot
    \left(n_{q_x}^2 + (n_{q_{x}}-1)^3 - 1\right)
    \\
    & \leq s\cdot o\cdot \left(n_{q_x} + n_{q_x}^2 + (n_{q_{x}}-1)^3 - 1\right) \\
    & \leq s\cdot o\cdot n_{q_x}^3.
  \end{align*}
  In the last inequality, we used that $z^3\geq z + z^2 + (z-1)^3-1$, for
  all real numbers $z$. This finishes the proof of the claim.

  \medskip We analyze now the compensation Step~2, in which the
  queries in $\Fmc_0(x_0)$ are further extended. We denote with $\Fmc_1$
  the result of applying Step~2A to $\Fmc_{0}(x_0)$. In Step~2A, we
  add at most one variable and a copy of $q$ for every variable in
  $\Fmc_0(x_0)$ and every choice of a concept name $A$ and role names
  $S$ and $R$ that occurs in \Omc. Therefore, we add at most $(1+n_q)\cdot
  n_{\Fmc_0(x_0)}\cdot o^3$ variables in total.  Using the claim, we get 
  \begin{align*}
    n_{\Fmc_1} & \leq n_{\Fmc_0(x_0)} + (1+n_q)\cdot
  n_{\Fmc_0(x_0)}\cdot o^3 \\& 
  \leq s\cdot o\cdot n_q^3 \cdot \left(1+(1+n_q)\cdot o^3\right).
  \end{align*}

  In Step~2B, we add at most one copy of $q$ for every role atom in some query
  in $\Fmc_1$ and every role name in $\Omc$. Hence,
  \begin{align*}
    n_{\Fmc} & \leq n_{\Fmc_1} + n_{\Fmc_1}\cdot n_q\cdot o \\
    & = \left(
s\cdot o\cdot n_q^3 \cdot \left(1+(1+n_q)\cdot o^3\right)
    \right)
  \cdot (1+n_q\cdot o),
  \end{align*}
  which is polynomial in $||q||$ and $||\Omc||$. Moreover, the computation of
  $\Fmc$ can be carried out in polynomial time since all the involved queries
  are of polynomial size and consequences of \Omc can be decided in polynomial
  time.
\end{proof}

\thmlowerfrontier*

\noindent
\begin{proof}\ For $n\geq 1$, let 
\begin{align*}
  q_n(x) &= A_1(x) \wedge A'_1(x) \wedge \cdots \wedge A_n(x)
  \wedge A'_n(x) \\[1mm]
  \Omc_n &= \{ A_i \sqcap A_i' \sqsubseteq A_1 \sqcap A_1' \sqcap
\cdots \sqcap A_n \sqcap A_n' \mid 1 \leq i \leq n \}.
\end{align*}
Suppose $\Fmc$ is a frontier of $q_n$ w.r.t.\ $\Omc_n$. Let
$p$ be any query that contains for each $i$ with $1\leq i\leq n$
either $A_i(x)$ or $A_i'(x)$. It suffices to show that $p\in
\Fmc$. 

Clearly, $q_n\subseteq_{\Omc_n} p\not\subseteq_{\Omc_n} q_n$ and thus Point~3 of the definition
of frontiers implies that there is a $p'\in \Fmc$ with
$p'\subseteq_\Omc p$. We distinguish cases: 
\begin{itemize}

  \item $p'$ contains the atoms $A_i(x),A_i'(x)$ for some $i$. But then
    $p'\equiv_{\Omc_n} q_n$ and $p'$ cannot be in \Fmc by Point~2 of
    the definition of frontiers, a contradiction. 

  \item $p'$ does not contain both atoms $A_i(x),A_i'(x)$ for any $i$.
    But then the ontology does not have an effect on the containment
    $p'\subseteq_{\Omc_n} p$ and hence every $A_i(x), A_i'(x)$ that occurs in
    $p$ must occur in $p'$.
    As $p'$ does not contain the atoms $A_i(x),A_i'(x)$ for any $i$, we
    actually have $p'=p$, which was to be shown.

\end{itemize}
\end{proof}

\section{Proofs for Section~\ref{sect:dllitef}}

A \emph{CQ-frontier} for an ELIQ $q$ w.r.t.\ \Omc is a finite set of
unary CQs that
satisfies Properties~1--3 of Definition~\ref{def:frontier}. Note that
every frontier is a CQ-frontier, but not vice versa. 

\thmnofinitefrontier*

\noindent
\begin{proof}\
    Let $q(x)= A(x)$ and
    \[
        \Omc = \{\ A \sqsubseteq \exists r,\quad
                  \exists r^- \sqsubseteq \exists r,\quad
                  \exists r \sqsubseteq \exists s,\quad
                  \mn{func}(r^-)\ \}.
    \]
    The universal model $\Umc_{q, \Omc}$ of $\Amc_q$ and $\Omc$ is an infinite
    $r$-path in which every point has a single $s$-successor.
   
    Suppose, for the sake of showing a contradiction, that \Fmc is a
    CQ-frontier of $q$ w.r.t.\ \Omc. We can assume w.l.o.g.\ that all queries
    in $\Fmc$ are satisfiable w.r.t.\ \Omc, especially that they satisfy
    $\mn{func}(r^-)$.
    Since $\Fmc$ is finite, there is an $n \geq 1$ such that 
    $|\mn{var}(p)| < n$, for all $p \in \Fmc$.
    Consider the following ELIQ $q'$:
    \begin{align*}
      q'(x_1) ={ } & r(x_1, x_2), \dots, r(x_{n - 1}, x_n), \\
                      & s(x_n, y), s(x'_n, y), \\
                      & r(x_1', x_2'), \dots, r(x_{n - 1}', x_n'), A(x_1').
    \end{align*}
    Note that $q'
    \not\subseteq_\Omc q \subseteq_\Omc q'$ and that $q'$ satisfies
    $\mn{func}(r^-)$.

    By Property~3 of frontiers, there is a query $p(z) \in \Fmc$ such
    that $p \subseteq_\Omc q'$. By
    Lemma~\ref{lem:hom-inclusion}, there is a homomorphism $h$ from $q'$ to $\Umc_{p,
    \Omc}$ with $h(x_1) = z$. We distinguish cases.

    Suppose first that $h(x_i) \in \mn{var}(p)$ for all $i$ with
    $1 \leq i \leq n$, then by the choice of $n$ there must be $1 \leq
    i < j \leq n$ such that $h(x_i) = h(x_j)$. Since $q'$ contains a
    directed $r$-path from $x_i$ to $x_j$ and $\Umc_{p,\Omc}$ does not
    contain edges between variables that are not part of $p$, this
    implies
    that $p$ must contain an $r$-cycle. Thus, $q \not \subseteq_\Omc
    p$, violating Property~1 of frontiers.
    
    Suppose now that $h(x_i) \notin \mn{var}(p)$ for some $i$ with \mbox{$1
    \leq i \leq n$}, that is, $h(x_i)$ is a trace starting with some $y
    \in \mn{var}(p)$. Since $q'$ is an ELIQ, there is a $j < i$ such
    that $h(x_j) = y$ and $h(x_{j+1}),\ldots,h(x_i)\notin\mn{var}(p)$.
    The structure of $q'$ and the structure of the anonymous part in
    universal models of \Omc imply that $h(x_j')=h(x_j)$.

    We now show that $h(x_1)=h(x_1')$. If $j=1$, we are done. If
    $j>1$, there are atoms $r(x_{j-1},x_j)$ and $r(x_{j-1}',x_j')$ in
    $q'$. Since $h$ is a homomorphism, $h(x_j)=h(x_{j}')$, and $p$
    satisfies $\mn{func}(r^-)$, we obtain $h(x_{j - 1}) = h(x_{j -
    1}')$. Repeating this argument yields $h(x_1) = h(x'_1)$ as
    required. Since $h(x_1)=z$, we also have $h(x'_1) = z$. Since $h$
    is a homomorphism and $A(x_1')\in q'$, we have $A(z)\in p$ and
    thus $p \subseteq_\Omc q$, violating Property~2 of frontiers.
%
%
\end{proof}

\lemfrontiernormalformF*

\noindent
\begin{proof}
  \ The proof of the Lemma is the same as the proof of
  Lemma~\ref{lem:frontier-normal-form-R}, except for the verification
  that the constructed set \Fmc satisfies Property~2 of frontiers. So
  we detail this here. 
  
  \smallskip\noindent\textit{Claim. } $q_F\not\subseteq q$, for all $q_F\in
  \Fmc$.

  \smallskip\noindent\textit{Proof of the claim.} Let $q_F(x) \in \Fmc$ and let
  $q^0_F(x)\in \Fmc'$ be the ELIQ from which $q_F(x)$ was obtained
  during the construction of \Fmc. Since $q^0_F\not\subseteq_{\Omc'}
  q$, there is an ABox $\Amc'$ and an individual $a \in
  \mn{ind}(\Amc')$ such that $\Amc', \Omc' \models q^F_0(a)$, but
  $\Amc', \Omc' \not\models q(a)$. As in the proof of
  Lemma~\ref{lem:frontier-normal-form-R}, the idea is to obtain an ABox $\Amc$ by
  starting with $\Amc'$ and adding $\Amc_{C(b)}$, for each concept assertion
  $X_{C}(b)\in \Amc'$.  However, the addition of $\Amc_{C(b)}$, with $C$ an
  \ELI-concept has to respect the functionality assertions of \Omc. In
  order to achieve this, we define the
  \emph{addition of a tree-shaped ABox $\Bmc$ with root $b_0$ to $\Amc$ at
  $a$} inductively on the structure of $\Bmc$ as follows:
  \begin{enumerate}

    \item for all $A(b_0)\in \Bmc$, add $A(a)$ to \Amc;

    \item for all $R(b_0,b')\in \Bmc$, let $\Bmc'$ be the sub-ABox of
      $\Bmc$ rooted at $b'$ and 

      \begin{enumerate}

	\item if $\mn{func}(R)\in \Omc$ and there is an atom
	  $R(a,a')$ in \Amc, then add $\Bmc'$ to $\Amc$ at $a'$;

	\item otherwise, add an atom $R(a,a')$ for a fresh individual
	  $a'$ and add $\Bmc'$ to \Amc at $a'$. 

      \end{enumerate}

  \end{enumerate}
%
%
%
%
%
%
%
%
  Importantly, there can only be one atom $R(b,b')$ in the first case
  for the existential restriction, if $\Amc$ satisfies the
  functionality assertions in \Omc. It should be clear that the
  resulting ABox also satisfies the functionality assertions if
  $\Amc$ does. 

  We thus obtain the ABox $\Amc$ by
  starting with $\Amc = \Amc'$ and adding $\Amc_{C(b)}$ to $\Amc$ at
  $b$, for each concept assertion $X_{C}(b)\in \Amc'$.
  We aim to show $\Amc,\Omc \models q_F(a)$ and
  $\Amc,\Omc \not\models q(a)$, witnessing $q_F \not\subseteq_\Omc q$ as
  required. 

  For the former, assume to the contrary that
  $\Amc,\Omc \not\models q_F(a)$. Then there is a model \Imc of \Amc
  and \Omc with $\Imc \not\models q_F(a)$. Let $\Imc'$ be the
  extension of \Imc according to Point~2 of
  Lemma~\ref{lem:normalform}. Then $\Imc'$ is a model of $\Omc'$
  and, by construction of $\Imc'$ and of $\Amc$ from $\Amc'$, also a
  model of $\Amc'$. Moreover, $\Imc \not\models q_F(a)$ implies
  $\Imc' \not\models q^0_F(a)$ by construction of $q_F$ and of
  $\Imc'$. This contradicts $\Amc', \Omc' \models q^F_0(a)$.

  It remains to show that $\Amc,\Omc \not\models q(a)$. Since
  $\Amc', \Omc' \not\models q(a)$, there is a model $\Imc$ of $\Omc'$
  and $\Amc'$ such that $\Imc \not\models q(a)$. Since
  $\Omc' \models \Omc$, \Imc is also a model of \Omc. Since
  $\Omc' \models X_{\exists R . C} \sqsubseteq \exists R . C$ for all
  $\exists R.C \in \Cmf(\Omc)$ and due to the construction of
  $\Amc$ from $\Amc'$, $\Imc$ is also a model of $\Amc$.
  Thus, \Imc witnesses $\Amc,\Omc \not\models q(a)$, as required. 

  \smallskip This
  finishes the proof of the claim and of the lemma.
\end{proof}

\lemfrontiermainf*

\noindent
\begin{proof}\
We show that $\Fmc$ fulfills the three conditions of frontiers. For
Condition~1, let $p(x_0)$ be a query from $\Fmc$ and let $p_0(x_0) \in
\Fmc_0(x_0)$ be the query that was used to construct $p$ by applying the
compensation step. First we observe that $p_0$ is satisfiable w.r.t.~\Omc, since
$q$ is satisfiable w.r.t.~\Omc and neither the dropping of a concept
atom nor the generalizing of
a subquery introduces any violations of functionality assertions in \Omc.
Next, let $R(x, z) \in p$ be an atom that was added during Step~2A of the
compensation step. Then either $\mn{func}(R) \notin \Omc$ or there is no atom
$R(x, z') \in p$ and therefore Step~2A introduces no atoms that violate any
functionality assertions.
Similarly Step~2B ensures that all added atoms to not violate functionality
assertions in $\Omc$.
Therefore $p$ is also satisfiable w.r.t.\ \Omc.
Hence it suffices to show $p(x_0) \to (\Umc_{q, \Omc}, x_0)$ by
Lemma~\ref{lem:hom-inclusion}. 

We extend the mapping $\cdot^\downarrow$ to be defined on all variables of $p$
by considering the yet unmapped variables added in Step 2A and Point~(iii) of
Step~2B.
Let $u$ be a fresh variable with $u^\downarrow$ undefined that was added
because there is a $x \in \mn{var}(p)$, a role $R$ and a set $M$ of concept
names such that $x^\downarrow \rightsquigarrow^R_{q, \Omc}~M$.
Then, by construction of $\Umc_{q, \Omc}$ there is a trace $x^\downarrow R M
\in \Delta^{\Umc_{q, \Omc}}$. Set $u^\downarrow = x^\downarrow R M$.
Now $\cdot^\downarrow$ is defined on all variables of $p$ and, by 
construction of $p$, it is a homomorphism from $p$ to $\Umc_{q, \Omc}$ with
$x_0^\downarrow = x_0$ as required.

\medskip
We start the proof of the second condition of frontiers with the following
claim:

\smallskip
\noindent
\textit{Claim~1.} $p\not\subseteq_\Omc q_x$ for all $x \in
\mn{var}(q)$ and $p(x) \in \Fmc_0(x)$.

\smallskip
\noindent
\textit{Proof of Claim~1.}
We show the claim by induction on the codepth of $x$ in $q$, matching the
inductive construction of $\Fmc_0$.
In the induction start, $x$ has codepth $0$. Then, by definition of codepth,
there is no $R(x, y) \in q$ that is
directed away from $x_0$ and all $p \in \Fmc_0(x)$ are obtained by dropping a
concept atom.

Let $p(x)$ be a query from $\Fmc_0(x)$ that is obtained by dropping the
concept atom $A(x) \in q$.
Then, by choice of $A(x)$, there is no $B(x) \in p$ with $\Omc \models B
\sqsubseteq A$ and no $R(x, x') \in p$ with $\Omc \models \exists R \sqsubseteq
A$. Hence $A(x) \in q_x$, but $A(x) \notin \Umc_{p, \Omc}$ and therefore $p
\not \subseteq_\Omc q_x$.

In the induction step, let $x$ have codepth $>0$, let $p(x)$ be a query from
$\Fmc_0(x)$ and assume that the claim holds for all variables with smaller
codepth.
Let $\cdot^\downarrow$ be the extension of the original $\cdot^\downarrow$ for $p$ to a homomorphism from $\Umc_{p, \Omc}$ to $\Umc_{q_x, \Omc}$, which exists by Lemma~\ref{lem:extend-hom}.
If $p$ is obtained by dropping a concept atom, then the same argument as in
the induction start yields $p \not\subseteq_\Omc q_x$.
If $p$ is obtained by generalizing the subquery attached to a role atom $R(x, y) \in q$, assume for
contradiction that there is a homomorphism $h$ from $q_x$ to $\Umc_{p, \Omc}$
with $h(x) = x$. From $h$ we construct a homomorphism $h'$ from $q$ to
$\Umc_{q, \Omc}$ with $h'(x_0) = x_0$ 
by setting $h'(z) = h(z)^\downarrow$ for all $z \in \mn{var}(q_y)$ and $h'(z) = z$
for all $z \notin \mn{var}(q_y)$.
Since $h$ must map $y$ to a $R$-successor of $x$ in $\Umc_{p, \Omc}$,
we may distinguish the following cases.
\begin{itemize}

\item $h(y)$ is a variable $z \in \mn{var}(p)$ with $z^\downarrow \neq y$.  Then, by
    definition of $\cdot^\downarrow$, $z^\downarrow = z$, and therefore, by
    construction of $h'$, $h'(y) = h'(z) = z$.
    Hence, $h'$ is a non-injective homomorphism from $q$ to $\Umc_{q, \Omc}$
    with $h'(x_0) = x_0$, contradicting \Omc-minimality or \Omc-saturatedness
    of $q$ by Lemma~\ref{lem:minimal-injective-hom}.

\item $h(y)$ is a trace $h(x) R M \in \Umc_{p, \Omc}$
    for some set $M$ of concept names.
    
    If $h'(y)$ is also a trace, then there must be a $y' \in \mn{var}(q)$ with
    $y' \notin \mn{img}(h')$, again contradicting \Omc-minimality or
    \Omc-saturatedness of $q$ by Lemma~\ref{lem:minimal-injective-hom}.

    If $h'(y)$ is not a trace, but a different successor $y'$ of $x$, then
    $h'(y') = h'(y)$, again contradicting $\Omc$-minimality of $q$ by
    Lemma~\ref{lem:minimal-injective-hom}.

    If $h'(y) = y$ and there is no $y' \in \mn{var}(q_y)$ with $h'(y') = x$,
    then we show a contradiction to \Omc-minimality $q$ by constructing a
    homomorphism $h''$ from $q$ to $\Umc_{q', \Omc}$ with $h''(x_0) = x_0$
    where $q'$ is the restriction of $q$ to $\mn{var}(q) \setminus\{y\}$.
    Note that by construction of $h'$, $h'(z) = y$ implies $z = y$.

    Since $h(x)RM^\downarrow = y$, there is no trace $xRM \in \Delta^{\Umc_{q,
    \Omc}}$. But, since $h(x)RM \in \Delta^{\Umc_{p, \Omc}}$ it must be that
    $h(x) \rightsquigarrow^S_{p, \Omc} M$ and thus $\Amc_p, \Omc \models
    \exists R.\bigsqcap M(h(x))$ and $\Amc_q, \Omc \models \exists R.\bigsqcap M(x)$. However,
    $x \not\rightsquigarrow^S_{p, \Omc} M$ because 
    $R(x, y) \in q$ and $\Amc_q, \Omc \models \bigsqcap M(y)$.

    Since $R(x, y) \notin q'$ and by \Omc-saturation of $q$ and normal form of
    $\Omc$, it follows that $x \rightsquigarrow^R_{q', \Omc} M$ and therefore
    there is a trace $x R M \in \Delta^{\Umc_{q', \Omc}}$.

    Construct $h''$ by setting $h''(z) = h'(z)$ for all $z \in \mn{var}(q)
    \setminus \mn{var}(q_y)$ and $h''(z) = x R M R_2 M_2 \dots R_n M_n$ for
    all $z \in \mn{var}(q_y)$ if $h(z) = x R M R_2 M_2 \dots R_n M_n$.

\item $h(y)$ is the root $y'$ of a $p' \in \Fmc_0(y)$ that was added in
  Points~3 or~4 of generalizing a subquery.
  
  By the induction hypothesis, $q_y \not\to \Umc_{p', \Omc}$ for
  all $p' \in \Fmc_0(y)$. We argue that, consequently, $h$ cannot map
  $q_y$ entirely into the subtree below $y'$ in $\Umc_{p, \Omc}$. To
  show this, it clearly suffices to argue that this subtree is
  isomorphic to $\Umc_{p', \Omc}$. In fact, this is the case by
  definition of universal models and since \Omc is in normal form,
  unless there is a concept name $A$ with
  $\Omc \models \exists R^- \sqsubseteq A$ such that
  $A(y') \in \Umc_{p, \Omc}$, but $A(y') \notin \Umc_{p', \Omc}$.
  Because of \Omc-saturatedness, this implies $A(y') \in p$ and
  $A(y') \notin p'$. Since the construction of $\Fmc_0$ never adds
  concept names to an ELIQ, this implies that
  $A({y'}^\downarrow) \in q$.  Thus, $A({y'}^\downarrow)$ was dropped
  during the construction of $p'$ from $q_{y'^\downarrow}$.
  This may only happen by dropping a concept atom. However,
  $R(x,y) \in q$ and   $\Omc \models \exists R^- \sqsubseteq A$
  contradicts Condition (b)~of dropping a concept atom.

  We have thus shown that
  $h$ cannot map $q_y$ entirely into the subtree below
  $y' \in \Umc_{p', \Omc}$. Consequently, there must be a $y'' \in \mn{var}(q_y)$
  with $h(y'') = x$.  This yields $h'(y') = h'(x) = x$ contradicting
  \Omc-minimality of $q$ by Lemma~\ref{lem:minimal-injective-hom}.

\end{itemize} 
This completes the proof of Claim~1.

\smallskip
We continue by using Claim~1 to show that $p \not\subseteq_\Omc q$ for all $p
\in \Fmc$.
Let $p$ be a query from \Fmc and assume for contradiction that $p
\subseteq_\Omc q$. Then, there is a homomorphism $h$ from $q$ to $\Umc_{p,
\Omc}$ with $h(x_0) = x_0$. 
Let $\cdot^\downarrow$ be the extension of the original $\cdot^\downarrow$ for
$p$ to a homomorphism from $\Umc_{p, \Omc}$ to $\Umc_{q, \Omc}$, which exists
by Lemma~\ref{lem:extend-hom}.
We compose $h$ and $\cdot^\downarrow$ to construct a
homomorphism $h'$ from $q$ to $\Umc_{q, \Omc}$
with $h'(x_0) = x_0$.
By Claim~1, there is no homomorphism that maps $q$ entirely into $\Umc_{p',
\Omc}$ for any $p' \in \Fmc_0(x_0)$. Hence, there must be an $x \in \mn{var}(q)$
such that $h(x)$ is a fresh variable added in the compensation
step. By definition of that step and since $q$ is connected, we may
distinguish the following cases:
\begin{itemize}

\item 
$h(x)$ is a fresh variable added in Step~2A. 

Then, by definition of $\cdot^\downarrow$, $h'(x)$ is a trace, contradicting
\Omc-minimality of $q$ by Lemma~\ref{lem:minimal-injective-hom}.

\item $h(x)$ is a fresh variable $z$ added in the start of Step~2B for the
role atom $R(y, y') \in p$ with $z^\downarrow = y^\downarrow$. 

Then, since $q$
is connected, there must be a predecessor $x'$ of $x$ with $h(x') = y$. Hence
$h'(x) = h'(x') = y^\downarrow$, contradicting \Omc-minimality of $q$ by
Lemma~\ref{lem:minimal-injective-hom}.

\item $h(x)$ is a fresh variable added in the iterated step of Step~2B.

Then, since $q$ is connected and the step only adds variables to the subtree
below a marked atom , there must be a predecessor $x'$ of $x$ such that
$h(x')$ is a fresh variable added in the start of Step~2B. This leads to the
same contradiction as in the last case.

\end{itemize} 
This completes the proof of Condition~2 of frontiers. 

\medskip

It remains to show that Condition~3 of frontiers is satisfied.  Let
$q'(x_0)$ be an ELIQ that is satisfiable w.r.t.\ \Omc such that
$q \subseteq_\Omc q' \not \subseteq_\Omc q$. We may assume w.l.o.g.\
that $q'$ is \Omc-saturated and that it satisfies all functionality assertions in \Omc.  If, in
fact, $q'$ contains atoms $R(x,y_1), R(x,y_2)$ with $y_1 \neq y_2$ and
$\mn{func}(R) \in \Omc$, then we can identify $y_1$ and $y_2$,
obtaining an ELIQ that is equivalent w.r.t.\ \Omc to the original $q'$.

There is a homomorphism $g$ from $q'$ to $\Umc_{q, \Omc}$ with $g(x_0) = x_0$.
We have to show that there is a $p \in \Fmc$ with
$p \subseteq_\Omc q'$. To do this, we construct in four steps a
homomorphism $h$ from $q'$ to $\Umc_{p, \Omc}$ with $h(x_0) = x_0$ for
some $p \in \Fmc$.  During all steps, we maintain the
invariant
\begin{equation*}  \label{eqn:h}
    h(z)^\downarrow = g(z) \tag{$*$}
\end{equation*}
for all variables $z \in \mn{var}(q')$ with $h(z)$ defined
and $\cdot^\downarrow$ the extension of the original $\cdot^\downarrow$ for $p$ to a homomorphism from $\Umc_{p, \Omc}$ to $\Umc_{q, \Omc}$.
In the first step of the construction, we define $h$ for an initial
segment of~$q'$.

Let $U \subseteq \mn{var}(q')$ be the smallest set of variables
(w.r.t.~$\subseteq$) of $q'$ such that $x_0 \in U$ and, for all $x\in
U$ and $R(x, y) \in q'$ directed away from $x_0$, we have: if $R(g(x),
g(y))$ is an atom in $q$ directed away from $x_0$, then $y \in
U$. Intuitively, $U$ induces the maximal initial segment of $q'$ that is
mapped in a `direction-preserving' way. Let $q^U$ be the
restriction of $q'$ to the variables in $U$. 

\smallskip
\noindent
\textit{Claim~2.}  For all $x \in U$ with
$q^U_x \not \subseteq_\Omc q_{g(x)}$, there is a $p \in \Fmc_0(g(x))$
and a homomorphism $h'$ from $q^U_x$ to
$\Umc_{p, \Omc}$ that satisfies \eqref{eqn:h}.

\smallskip
\noindent
\textit{Proof of Claim~2.}
Let $y = g(x)$.
We show Claim~2 by induction on the codepth of $x$ in $q^U$.  In the induction
start, $x$ has codepth $0$. We distinguish the following cases:
\begin{itemize}

\item There is an $R(y, y') \in q_y$.

  Then let $p \in \Fmc_0(y)$ be constructed by generalizing the subquery attached to  
  $R(y, y')$ and set $h'(x) = y$. Since $q$ is \Omc-saturated,
  $A(y)\in \Umc_{q,\Omc}$ implies   $A(y)\in \Umc_{p,\Omc}$.
  This and  $y = g(x)$ implies that $h'$ is a homomorphism.

\item There is no $R(y, y') \in q_y$.

  Then $q^U_x \not \subseteq_\Omc q_{y}$ implies that there is an $A(y) \in q_y$ with
  $A(x) \notin \Umc_{q^U_x, \Omc}$, and we must even find an $A$ with
  these properties such that there is no $B(y) \in q_y$ with
  $\Omc \models B \sqsubseteq A$ and
  $\Omc \not\models A \sqsubseteq B$. This implies that Property~(a)
  of dropping concept atoms is satisfied. Property~(b) is satisfied
  since there is no $R(y, y') \in q_y$ and thus we may construct
  $p \in \Fmc_0(y)$ by dropping the concept atom $A(y)$. Set
  $h'(x) = y$.

\end{itemize}
In the induction step, let $x$ have codepth $>0$ and assume that the claim
holds for all variables of smaller codepth. From $q^U_x \not
\subseteq_\Omc q_{y}$, it follows that $q_{y} \not\to (\Umc_{q^U_{x},
    \Omc}, x)$.
We distinguish the following cases:
\begin{itemize}

\item There is an $R(y, y') \in q_y$ such that $q_{y'} \not\to (\Umc_{q^U_{x'},
    \Omc}, x')$ for all $R(x, x') \in q^U_x$.
  
First assume $\mn{func}(R) \notin\Omc$. Then let $p \in \Fmc_0(y)$ be constructed by
generalizing the subquery attached to $R(y, y')$.
We construct the homomorphism $h'$ from $q^U_x$ to $\Umc_{p, \Omc}$ by starting with
$h'(x) = y$ and continuing to map all successors of~$x$.
Let $S(x, x') \in q^U_x$. If $g(x') \neq y'$, then extend $h'$ to the subtree below $x'$ 
by setting $h'(z) = g(z)$ for all $z \in \mn{var}(q_{x'})$.
If $g(x') = y'$, then, by the induction hypothesis,
there is a $p' \in \Fmc_0(y')$ and a homomorphism $h''$ from $q_{x'}$ to $\Umc_{p', \Omc}$ with $h''(x') = y'$.
Extend $h'$ to the variables in $q_{x'}$ by mapping $q_{x'}$ according to $h''$ to the copy of
$p'$ that was attached to $y$ in Point~3 of generalizing the subquery attached to $R(y, y')$.

Now assume $\mn{func}(R) \in \Omc$. Then there is at most one
$R(x, x') \in q^U_x$ with $g(x') = y'$.  If there is none, choose an
arbitrary $p \in \Fmc_0(y)$ constructed by generalizing the subquery attached to $R(y, y')$ and
extend $h'$ as above.  If there is a single such $R(x, x')$, then, by
the induction hypothesis, there is a $p' \in \Fmc_0(y')$ and
homomorphism $h''$ from $q^U_{x'}$ to $\Umc_{p', \Omc}$ with
$h''(x') = y'$.  Let $p \in \Fmc_0(y)$ be constructed by generalizing the subquery attached to
$R(y, y')$ and attaching $p'$ in Step~4, then extend $h'$ as above.

\item For every $R(y, y') \in q_y$, $q_{y'} \to (\Umc_{q^U_{x'},
    \Omc}, x')$ for some $R(x, x') \in q^U_x$.

  Then there is an $A(y) \in q_y$ with
  $A(x) \notin \Umc_{q^U_x, \Omc}$ and we
  must even find an $A$ with these properties and such that there is
  no $B(y) \in q_y$ with $\Omc \models B \sqsubseteq A$ and
  $\Omc \not\models A \sqsubseteq B$. Thus, Property~(a) of dropping
  concept atoms is satisfied. To show that Property~(b) is also
  satisfied, we have to argue that there is no $R(y, y') \in q_y$ with
  $\Omc \models \exists R \sqsubseteq A$. But by assumption
  for any such
$R(y,y') \in q_y$ we have $q_{y'} \to (\Umc_{q^U_{x'},
    \Omc}, x')$ for some $R(x, x') \in q^U_x$. This implies 
  $A(x) \in \Umc_{q^U_x, \Omc}$, a contradiction.
    
We may thus construct $p \in \Fmc_0(y)$ 
by dropping the concept atom $A(y)$. Set $h'(x') = g(x')$ for all $x' \in
\mn{var}(q^U_x)$.

\end{itemize} 
This completes the proof of Claim~2.

\smallskip By Claim~2, there is a $p' \in \Fmc_0(x_0)$ such that
$q^U \to \Umc_{p', \Omc}$. Let $p \in \Fmc$ be the query that was
obtained by applying the compensation step to $p'$. Then clearly also
$q^U \to \Umc_{p, \Omc}$. Define $h$ for all variables in $U$
according to the homomorphism that witnesses this
and let $\cdot^\downarrow$ be the extension of the original $\cdot^\downarrow$
for $p$ to a homomorphism from $\Umc_{p, \Omc}$ to $\Umc_{q, \Omc}$.

\smallskip

We continue with the second step of the construction of~$h$ which
covers parts of $q'$ that are connected to the initial segment $q^U$
and which are mapped to traces of $\Umc_{q,\Omc}$ rather than to
$\mn{var}(q)$.  Consider all atoms $R(x, x') \in q'$ with $h(x)$
defined, $h(x')$ undefined and $g(x') \notin \mn{var}(q)$.
Before extending $h$ to $x'$, we first show that there is an atom $R(h(x), z)
\in p$, added in Step~2A.

Since $g(x') \notin \mn{var}(q)$,
$g(x')$ must be a trace $g(x) R M \in \Delta^{\Umc_{q, \Omc}}$ for
some set $M$, hence $g(x) \rightsquigarrow^R_{q, \Omc} M$. 
This implies that there is no $R(g(x), y) \in q$ 
with $\Amc_q, \Omc \models \bigsqcap M(y)$.
Furthermore, assume that there is a concept name $B$ such that $\Omc \exists R
\sqsubseteq B$ but $B(h(x)) \notin p$. Then, since $q$ is $\Omc$-saturated,
$B(g(x)) \in q$ and $p$ must be the result of dropping the concept atom $B(g(x))$.
However, since $q'$ is $\Omc$-saturated, $B(x) \in q'$, contradicting Claim~2,
therefore there is no such concept name.

Hence, Step~2A adds a fresh variable $z \in \mn{var}(p)$ and the atom $R(h(x),
z)$ to $p$ with $z^\downarrow = g(x) R M$.
We extend $h$ to the initial segment of $q'_{x'}$
that is mapped by $g$ into the traces below $g(x')$.
Set $h(\hat x) = z R_2 M_2 \dots R_n M_n$
for all $\hat x$ in this initial segment with 
 $g(\hat x) = g(x) R M R_2 M_2 \dots R_n M_n$.
If we reach a $R(x'', x''') \in q'_{x'}$
directed away from $x_0$ with $g(x'')$ a trace and $g(x''') \in \mn{var}(q)$,
then leave $h(x''')$ undefined. The mapping $h$ will be extended to the subtree
$q'_{x'''}$ in the next steps.
Note that~\eqref{eqn:h} is satisfied.

\smallskip

In the third step of the construction of $h$, we consider all
$R(x, x') \in q'$ directed away from $x_0$ with $h(x)$ defined and
$h(x')$ undefined. Before defining $h(x')$, we first show that
\begin{itemize}

\item[(a)] there is an atom $R(h(x),y) \in p$ directed towards $x_0$ such
  that \mbox{$\mn{func}(R^-) \notin \Omc$}
  and

\item[(b)] $g(x') \in \mn{var}(q)$.
  
\end{itemize}
Distinguish the following cases:
\begin{itemize}

\item $g(x)$ is a trace.

  Then $h(x)$ was defined in the second step. From the fact that
  $h(x')$ was not defined in the second step, it follows that
  $g(x') \in \mn{var}(q)$, as required for (b). Consequently, $g(x)$
  must be of the form $g(x') R^- M \in \Umc_{q, \Omc}$ and $h(x)$ was
  defined in the second step to be a fresh variable added to $p$ in
  Step~2A of its construction and this variable is an $R^-$-successor
  of some variable $y$, that is, $R^-(y, h(x)) \in p$ directed away
  from~$x_0$. We may thus use the inverse of this atom as the desired
  atom $R(h(x),y)$ in~(a).  Since $g(x') R^- M$ is a trace in
  $\Umc_{q, \Omc}$ and due to the syntactic restriction adopted by the
  \DLFsynr ontology \Omc, we further have $\mn{func}(R) \notin \Omc$.

\item $g(x)$ is not a trace, that is, $g(x) \in \mn{var}(q)$.

  Since $h(x')$ has neither been defined in the first nor in the
  second step, we must have $x \in U$, $g(x') \in \mn{var}(q)$ (as
  required for~(b)) and \mbox{$R(g(x), g(x')) \in q$} is directed
  towards $x_0$.  The latter implies that $x$ is not the root of $q'$,
  thus $q'$ contains an atom $S(x'',x)$ directed away from~$x_0$. From
  $x \in U$, it follows by definition of $U$ that $x'' \in U$. Thus
  $h(x'')$ and $h(x')$ were both defined in the first step and, due to
  the formulation of that step, $S(x'',x) \in q'$ directed away from
  $x_0$ implies that $q$ contains the atom $S(g(x''),g(x))$ directed
  away from $x_0$. So $S(g(x''),g(x)) \in q$ is directed away from
  $x_0$ and $R(g(x), g(x')) \in q$ is directed towards $x_0$. Since
  $q$ is a tree, this implies $g(x'')=g(x')$ and $S=R^-$.  We have
  thus shown that $R^-(x'',x)$ and $R(x,x')$ are atoms in $q'$ that
  are both directed away from $x_0$. Since $q'$ satisfies all
  functionality assertions in \Omc, this implies
  $\mn{func}(R) \notin \Omc$.  We use $R(h(x), h(x'')) \in p$ as the
  desired atom $R(h(x),y)$.
  
\end{itemize}
%
%

Consider the inverse $R^-(y,h(x))$ of the atom $R(h(x),y) \in p$ that
exists due to~(a). In the start of Step~2B of the construction of $p$,
the inverse atom $R^-(y,h(x))$ is considered and leads to the
introduction of an atom $R(h(x),y')$, $y'$ a fresh variable with
${y'}^\downarrow=y^\downarrow$.  Set $h(x') = y'$. Note that
$R(h(x),h(x')) \in p$ was marked in Step~2B of the construction of~$p$.

In the final step of the construction of $h$ we define
$h(x)$ for all remaining variables $x$. We do this by repeatedly
choosing atoms $R(x, x') \in q'$ directed away from $x_0$ such that
\begin{enumerate}

\item 
$h(x)$ and
$h(x')$ defined and

\item for all $S(x', x'') \in q'$ directed away from $x_0$, $h(x'')$
  is undefined and there is at least one such $S(x', x'')$.
  
\end{enumerate}
If we choose such an $R(x,x')$ directly after the third step of the
construction of $h$, then $g(x') \in \mn{var}(q)$ due to (b) and
$R(h(x),h(x')) \in p$ was marked in Step~2B of the construction
of~$p$. We implement our extension of $h$ such that these conditions
are always guaranteed when we choose an $R(x, x') \in q'$ that
satisfies Properties~1 and~2 above.

Let $R(x, x') \in q'$ be an atom that satisfies Properties~1 and~2.
First assume that $\mn{func}(R^-) \notin \Omc$. Then processing the
marked atom $R(h(x),h(x'))\in p$ in Step~2B of the construction of $p$
results in a copy $\widehat q$ of $q$ to be added to $p$, with the
copy of ${h(x')}^\downarrow$ in $\widehat q$ glued to $h(x')$.  
Define $h$ for all variables $x'' \in \mn{var}(q'_{x'})$ by setting $h(x'')$ to
be the copy of $g(x'')$ in $\widehat q$ if $g(x'')$ is a variable, or to be the
trace $h(x''') R_1 M_1 \dots R_n M_n$ if $g(x'')$ is the trace $g(x''') R_1
M_1 \dots R_n M_n$.

Now assume that $\mn{func}(R^-) \in \Omc$. Consider each
$S(x', x'') \in q'$ directed away from $x_0$. We distinguish two cases:
\begin{itemize}

\item $g(x'')$ is a trace.

  Since $g(x') \in \mn{var}(q)$, $g(x'')$ must be of the form
  $g(x') S M$. Thus, $g(x') \rightsquigarrow^S_{q, \Omc} M$ and
  by~\eqref{eqn:h}, $h(x')^\downarrow \rightsquigarrow^S_{q, \Omc} M$ 
  and when the marked atom $R(h(x),h(x'))\in p$ is processed in
  Step~2B of the construction of $p$. Thus, Point~(iii) of Step~2B adds
  to $p$ atoms $S(h(x'),u)$, $S^-(u,y')$, and $A(u)$ for every
  $A \in M$. Additionally, processing the marked atom $S^-(u, y')$ attaches a
  copy $\widehat q$ of $q$ to $y'$, since $y^{\prime\downarrow} = g(x')
  \rightsquigarrow^S_{q, \Omc} M$ and therefore either $\mn{func}(S) \notin
  \Omc$ or there is no atom $S(y^{\prime\downarrow}, z) \in q$.

  Extend $h$ by setting $h(x''') = u R_2 M_2 \dots R_n M_n$ if $g(x''') = g(x')
  S M R_2 M_2 \dots R_n M_n$ for all $x''' \in \mn{var}(q'_{x''})$ up until
  $g(x''') = g(x')$. If there is a subtree $q'_{x'''}$ with $x''' \in
  \mn{var}(q'_{x''})$ and $g(x''') = g(x')$, map it to $y'$ and the attached
  $\widehat q$ by setting $h(x''') = y'$ and all $h(z)$ for $z \in
  \mn{var}(q'_{x'''})$ to the copy of $g(z)$ in $\widehat q$, or the trace
  starting in $\widehat q$.

\item $g(x'')$ is not a trace.

  Since $g(x') \in \mn{var}(q)$, this implies
  \mbox{$S(g(x'), g(x'')) \in q$.} It follows from \eqref{eqn:h} that
  $h(x')^\downarrow = g(x')$ and thus
  $S(h(x')^\downarrow, g(x'')) \in q$. When the marked atom
  marked atom $R(h(x),h(x'))\in p$ is processed in Step~2B of the
  construction of $p$, then $S(h(x')^\downarrow, g(x'')) \in q$ is
  thus one of the atoms under consideration in Point~(ii).

  Since $R(x,x') \in q'$ and $S(x',x'') \in q'$ are both directed away
  from $x_0$ and $q'$ satisfies all functionality assertions in \Omc,
  we have $S \neq R$. It follows from \eqref{eqn:h} that
  $h(x)^\downarrow = g(x)$ As a consequence
  $S(h(x')^\downarrow,g(x'')) \neq
  R^-(h(x')^\downarrow,h(x)^\downarrow)$.
  Thus, in Point~(ii) of Step~2B an atom $S(h(x'),z')$ is
  added to $p$ with $z'$ a fresh variable and ${z'}^\downarrow=g(x'')$
  and this atom is marked. Set $h(x'') = z'$ and leave the successors
  of $x''$ in $q'$ to be processed in subsequent iterations of the
  loop in step four of the construction of $h$.

\end{itemize}
This completes the construction of $h$ and the proof of Condition 3.
\end{proof}

\lemfrontiersizeF*

\noindent \begin{proof}\ In order to reduce notational clutter, we
  introduce some abbreviations used throughout the proof. 
  \begin{itemize}

    \item $s=|\mn{sig}(q)|$ denotes the number of concept and role
      names used in $q$;

    \item $o=||\Omc||$ denotes the size of \Omc;

    \item for an ELIQ $p$, $n_p=|\mn{var}(p)|$ denotes the number of
      variables in $p$;

    \item for a set $Q$ of queries, $n_Q$ denotes $\sum_{p\in Q} n_p$.

  \end{itemize}
  We assume without loss of generality that $s$ and $o$ are at least
  one.

  We start with analyzing the size of the queries in $\Fmc_0(x)$ that
  are obtained as the result of the `generalize' step. 

  \smallskip\noindent\textit{Claim.} For every $x\in
  \mn{var}(q)$, we have:
  \begin{enumerate}

    \item $|\Fmc_0(x)|\leq s\cdot n_{q_x}$;

    \item $\displaystyle n_{\Fmc_0(x)} \leq s
  \cdot n_{q_x}^3.$

  \end{enumerate}

  \smallskip\noindent\textit{Proof of the claim.} The proof of both
  points is by
  induction on the codepth of $x$ in $q$. We start with Point~1. 
  For the base case, consider
  a variable $x$ of codepth $0$ in $q$, that is, a leaf. In this case,
  only Step~(A) is applicable, and it adds at most $s$ queries to
  $\Fmc_0(x)$.

  For the inductive step, consider a variable $x$ of codepth greater
  than $0$. We partition $\Fmc_0(x)$ into $\Fmc_0^{A}(x)$ and $\Fmc_0^{B}(x)$,
  that is, the queries that are obtained by dropping a concept atom in
  Step~(A) and the queries that are obtained by generalizing a subquery in
  Step~(B), respectively, and analyze them separately, starting with
  $\Fmc_0^{A}(x)$. Clearly, there are at most $s$ queries in
  $\Fmc_0^A$, that is,  
  \[|\Fmc_0^A(x)|\leq s.\]
  Next, we analyze $\Fmc_0^{B}(x)$. Each query in $\Fmc_0^{B}(x)$ is
  obtained by first picking, in Point~1, an atom $R(x,y)$ in $q_x$. If
  $\mn{func(R)}\notin \Omc$, we add
  1 query to $\Fmc_0^{B}(x)$. Otherwise, we
  add $|\Fmc_0(y)|$ queries (in Point~4). Thus, we obtain
  \begin{align*}
    |\Fmc_0^B(x)| &\leq |\{R(x,y)\in
      q_x\mid \mn{func}(R)\notin \Omc\}| + {}\\
      &\phantom{ {}\leq {} }\sum_{\substack{R(x,y)\in
	q_x,\\\mn{func}(R)\in \Omc}} |\Fmc_0(y)|
  \end{align*}
  Using the fact that $n_{q_y}\geq 1$ and the induction hypothesis, we
  obtain
  \[ |\Fmc_0^B(x)| \leq \sum_{R(x,y)\in q_x} s \cdot n_{q_y} =
  s\cdot \sum_{R(x,y)\in q_x}n_{q_y}.\]
  The above sum can be simplified to $n_{q_x}-1$. Hence, we obtain
  \begin{align*}
    |\Fmc_0(x)| = |\Fmc_0^A(x)| + |\Fmc_0^B(x)| \leq s+s\cdot
    (n_{q_x}-1) = s\cdot n_{q_x}.
  \end{align*}

  \medskip

  We now prove Point~2, again by induction on the codepth of $x$ in
  $q$. For the base case, consider a variable $x$ of codepth $0$ in
  $q$, that is, a leaf. In this case, only Step~(A) is applicable, and
  it adds at most $s$ queries of size $1$ to $\Fmc_0(x)$. 

  For the inductive step, consider a variable $x$ of codepth greater
  than $0$ and the same partition of $\Fmc_0(x)$ into $\Fmc_0^{A}(x)$
  and $\Fmc_0^{B}(x)$ as before. Clearly, every $p\in \Fmc_0^A(x)$ uses $n_{q_x}$
  variables and there are at most $s$ queries in $\Fmc_0^A$. Thus, we
  have
  \[n_{\Fmc_0^A(x)} \leq s\cdot n_{q_x}.\]
  Next, we analyze $\Fmc_0^{B}(x)$. Each query in $\Fmc_0^{B}(x)$ is
  obtained by first picking, in Point~1, an atom $R(x,y)$ in $q_x$. If
  $\mn{func(R)}\notin \Omc$, we add
  $\sum_{p\in \Fmc_0(y)} n_p$ variables (in Point~3). Otherwise, we
  replace $q_y$ with some element of $\Fmc_0(y)$ (in Point~4). Thus, we obtain
  \begin{align*}
    n_{\Fmc_0^B(x)} &\leq \sum_{\substack{R(x,y)\in
      q_x,\\\mn{func}(R)\notin \Omc}} (n_{q_x}+ n_{\Fmc_0(y)})
      + {}\\
      &\phantom{{}\leq{ }}\sum_{\substack{R(x,y)\in
	q_x,\\\mn{func}(R)\in \Omc}} (n_{q_x} \cdot |\Fmc_0(y)| +
	n_{\Fmc_0(y)})\\
      &\leq \sum_{R(x,y)\in
	q_x} (n_{q_x} \cdot |\Fmc_0(y)| +
	n_{\Fmc_0(y)}).
  \end{align*}
  Plugging in the
  induction hypothesis from both Point~1 and~2, we obtain 
  \begin{align}
    n_{\Fmc_0^B(x)} &  \leq \sum_{R(x,y)\in
    q_x} (n_{q_x}\cdot s\cdot n_{q_y}+s\cdot  n_{q_y}^3)\notag \\
  &  = s\cdot n_{q_x}\sum_{R(x,y)\in
  q_x} n_{q_y}+s\sum_{R(x,y) \in q_x} n_{q_y}^3.
    \label{eq:fb2}
  \end{align}
  We simplify the right-hand side of~\eqref{eq:fb2} by making the
  following observations:
  \begin{itemize}

    \item $\sum_{R(x,y)\in q_x} n_{q_y} = n_{q_x}-1$,
      and

    \item $\sum_{R(x,y)\in q_x} n_{q_y}^3\leq \left(\sum_{R(x,y)\in
      q_x} n_{q_y}\right)^3=(n_{q_x}-1)^3$. Here, the inequality is an
      application of the general inequality $\sum_{i}a_i^3\leq
      \left(\sum_{i}a_i\right)^3$, for every sequence of non-negative
      numbers $a_1,\ldots,a_k$.

  \end{itemize}
  Using these observations, Inequality~\eqref{eq:fb2} can be simplified to:  
  \begin{align*}
    n_{\Fmc_0^B(x)} & \leq s\cdot n_{q_x}\cdot (n_{q_x}-1) + s\cdot (n_{q_x}-1)^3\\
    & = s\cdot \left( n_{q_{x}}^3 -2 n_{q_x}^2 + 2n_{q_x}-1  \right).
   \end{align*}
  Overall, we get 
  \begin{align*}
    n_{\Fmc_0(x)} & = n_{\Fmc_0^A(x)} + n_{\Fmc_0^B(x)} \\
    & \leq s\cdot n_{q_x} + s\cdot \left( n_{q_{x}}^3 -2 n_{q_x}^2 + 2n_{q_x}-1  \right)
    \\
    & = s\cdot \left( n_{q_{x}}^3 -2 n_{q_x}^2 + 3n_{q_x}-1  \right)
    \\
    & \leq s\cdot n_{q_x}^3.
  \end{align*}
  In the last inequality, we used that $z^3\geq z^3-2z^2 + 3z-1$, for
  all numbers $z\geq 1$. This finishes the proof of the claim.

  \medskip We analyze now the compensation Step~2, in which the
  queries in $\Fmc_0(x_0)$ are further extended. We let $\Fmc_1$
  denote
  the result of applying Step~2A to $\Fmc_{0}(x_0)$. In Step~2A, we
  add at most one variable per variable in $\Fmc_0(x_0)$ and concept $\exists
  R.B$ that occurs in \Omc. Therefore, we add at most $n_{\Fmc_0(x_0)}\cdot o$
variables.  Using the claim, we get \[n_{\Fmc_1}\leq n_{\Fmc_0(x_0)} +
  n_{\Fmc_0(x_0)}\cdot o\leq s\cdot n_q^3
\cdot (1+o).\]

  We now analyze Step~2B, applied to some query $p\in \Fmc_1$. First
  of all note that the marking proviso ``if $R(x,y)$ is marked
  then $y^\downarrow$ is defined and if $x^\downarrow$ is undefined,
  then $\mn{func}(R^-)\notin \Omc$ or $q$ contains no atom of the form
  $R(y^\downarrow,z)$'' is indeed satisfied. 

  Consider now an atom $R(x,y)\in p$ that was marked in the \emph{Start}
  phase. We distinguish two cases. 

  \begin{itemize}

    \item If $x^\downarrow$ is undefined, then the marking proviso
      implies that $\mn{func}(R^-)\notin \Omc$ or $q$ contains no atom
      of the form $R(y^\downarrow,z)$. In the \emph{Step} phase, we
      just unmark the atom and add a copy of $q$, hence no iteration
      takes place, and the query size increases by $n_q$.

    \item Otherwise, $x^\downarrow$ is defined, and by definition of
      ${\phantom\cdot}^\downarrow$, we have
      $R(x^\downarrow,y^\downarrow)\in q$. Now, the iterative process
      ensures that:

      \begin{itemize}

	\item Whenever an atom $S(y,z')$ is marked in~(ii), then both
	  $y^\downarrow$ and ${z'}^\downarrow$ are defined and
	  $S(y^\downarrow,{z'}^\downarrow)\in q$. Moreover, the
	  condition `$S(y^\downarrow,z)\neq
	  R^-(y^\downarrow,x^\downarrow)$' and the fact that $q$ is an
	  ELIQ ensure that every atom from $q$ is `met' at most once
	  during the entire process.

	\item Whenever an atom $S^-(u,y')$ is marked in~(iii), then
	  $u^\downarrow$ is undefined. Hence, the marking proviso
	  implies that, in the \emph{Step} phase, this atom is
	  unmarked, a copy of $q$ is added, and the iteration stops. 

      \end{itemize}
      
  \end{itemize}
  Overall, we obtain that, per role atom in $p$, the marking process adds
  at most $n_q$ role atoms in Step~(ii), for each such atom and every
  $\exists r.B$ in \Omc one more role atom in Step~(iii), and for each
  introduced variable at most one copy of $q$. All this is polynomial
  in $||q||$ and $||\Omc||$. Moreover, the computation of
  $\Fmc$ can be carried out in polynomial time since all the involved queries
  are of polynomial size and consequences of \Omc can be decided in polynomial
  time.
\end{proof}

\section{Proofs for Section~\ref{sect:uniquechar}}

  \thmunique*

\noindent
\begin{proof}\
  Let \Omc and $q(x)$ be as in the theorem. By
  Theorems~\ref{thm:frontiermain} and~\ref{thm:frontiermainF}, we can compute in polynomial time a
  frontier $\Fmc_q(x)$ for $q$ w.r.t.\ \Omc. Let
  $E^+=\{(\Amc_q,x)\}$ and $E^-=\{(\Amc_p,x)\mid p\in \Fmc_q(x)\}$.
  It is not hard to verify that $q$ fits  $(E^+,E^-)$.
  We show that $(E^+,E^-)$ in fact uniquely characterizes $q$ w.r.t.\
  \Omc.
  
  Let $q'$ be an ELIQ that fits $(E^+,E^-)$. We have
  $q\subseteq_\Omc q'$ since $(\Amc_q,x)$ is a positive example.
  Moreover, since all data examples in $E^-$ are negative examples for
  $q'$, we know that $p\not\subseteq_\Omc q'$ for any
  $p\in\Fmc_q(x)$. By Point~3 of the definition of frontiers, we can
  conclude that $q'\subseteq_\Omc q$. Thus
  $q'\equiv_{\Omc} q$, as required.
\end{proof}

\section{Proofs for Section~\ref{sec:learning}}

We start with showing how to construct a seed CQ in the case that the
ontology \Omc contains no concept disjointness constraints. This is in
fact trivial if \Omc contains no role disjointness constraint either,
as then we can simply use
$$q^0_H(x_0)=\bigwedge_{A \in \Sigma \cap
  \NC} A(x_0) \wedge \bigwedge_{r \in \Sigma \cap
  \NR} r(x_0,x_0).$$
Here and in what follows, for brevity we use $\Sigma$ to denote
$\mn{sig}(\Omc)$.

We consider now the case with role disjointness constraints (but still
without concept disjointness). Let
$\Rbf = \{ r_1,\dots,r_m \}$ be the set of all role names $r\in\Sigma$
such that $\exists r$ is satisfiable w.r.t.\ \Omc. If, for example,
\Omc contains $r \sqsubseteq s$ and $r \sqcap s \sqsubseteq \bot$, then
$\exists r$ is not satisfiable w.r.t.\ \Omc.

Introduce variables $x_0,\dots,x_{2m}$ and let $K_{2m+1}$ be the
$2m+1$-clique that uses these variables as its vertices. It is known
that for each $n \geq 1$, the $n$-clique $K_n$ has at least
$\frac{n-1}{2}$ Hamilton cycles that are pairwise edge-disjoint, see
for instance the survey \cite{Hamilton}. We
thus find in $K_{2m+1}$ Hamilton cycles $P_1,\dots,P_m$ that are
pairwise edge-disjoint. By directing the cycles, we may view each
$P_i$ as a set of  directed edges $(x_j,x_\ell)$. We then set 
$$
\begin{array}{rcl}
  q^0_H(x_0) &=& \displaystyle \bigwedge_{A\in \Sigma\cap\NC,\
                 0 \leq i \leq 2m}  \!\!\!\!\!\!  A(x_i) \\[6mm]
             &&\displaystyle \bigwedge_{(x_i,x_j) \in P_1} \!\!\!\!\!\!
                r_1(x_i,x_j)
                \wedge \cdots \wedge \!\!\!\!\!\!
\bigwedge_{(x_i,x_j) \in P_m} \!\!\!\!\!\!  r_m(x_i,x_j).
\end{array}
$$
Clearly, $q^0_H$ has no multi-edges and thus satisfies all role
disjointness constraints in \Omc. Moreover, every variable has exactly
one $r$-successor and exactly one $r$-predecessor for every role name
$r \in \Rbf$. On the one hand, this implies that all functionality
assertions in \Omc are satisfied. On the other hand, it means that
there is a homomorphism from every target ELIQ $q_T$ to $q^0_H$
because any $q_T$ is required to be satisfiable w.r.t. \Omc and
thus may only use role names from \Rbf.

\medskip

If \Omc contains at least one concept disjointness constraint
$B_1 \sqcap B_2 \sqsubseteq \bot$, then we cannot use the above
$q^0_H$ as it is not satisfiable w.r.t.\ \Omc, but we may obtain a seed query
$q^0_H$ by viewing $B_1 \sqcap B_2$ as an ELIQ $q$ in the obvious way
and posing $q$ as an equivalence query to the oracle. Since the target
query is satisfiable w.r.t.\ \Omc, the oracle is forced to return a
positive counterexample $(\Amc,a)$, that is, a pair $(\Amc,a)$ such
that $\Amc,\Omc \models q_T(a)$ and $\Amc,\Omc \not\models q_H(a)$.
The desired query $q^0_H$ is $(\Amc,a)$ viewed as a CQ with answer
variable $a$.  Note that when learning with equivalence queries, then
in polynomial time learnability, the running time of the learning
algorithm may also polynomially depend, at any given time, on the size
of the largest counterexample returned by the oracle so far. This
condition is satisfied by our algorithm.

\lemlearningnormalform*

\noindent
\begin{proof}\
    We show the lemma by converting a
    learning algorithm $L'$ for ontologies in normal form into a learning
    algorithm $L$ for unrestricted ontologies, relying on the
    normal form described in Lemma~\ref{lem:normalform}. Since $L$ will ask a single
    query for every query asked by $L'$, the lemma follows.

    We start with \DLR. Given a $\DLR$ ontology $\Omc$ and a signature $\Sigma =
    \mn{sig}(\Omc)$ with $\mn{sig}(q_T)\subseteq \Sigma$, algorithm $L$ first computes the
    ontology $\Omc'$ in normal form as per Lemma~\ref{lem:normalform},
    choosing the fresh concept names so that they are not from $\Sigma$.
    It then runs $L'$ on $\Omc'$ and $\Sigma' = \Sigma \cup \mn{sig}(\Omc')$.
    In contrast to $L'$, the oracle still works with the original ontology
    $\Omc$. To ensure that the answers to the queries posed to the oracle are
    correct, $L$ modifies $L'$ as follows.

    Whenever $L'$ asks a membership query $\Amc', \Omc' \models
    q_T(a)$, we may assume that $\Amc'$ satisfies the functionality
    assertions from \Omc, since otherwise the answer is trivially
    ``yes''. Then, $L$
    instead asks the membership query $\Amc, \Omc \models q_T(a)$, where $\Amc$
    is obtained from $\Amc'$ as follows. Start with $\Amc=\Amc'$, and 
    \begin{itemize}

      \item[$(\ast)$] add $C(b)$, for each concept assertion $X_{C}(b)\in \Amc'$.

    \end{itemize}
    Here, the addition of $C(b)$ for an \ELI-concept $C$ to an ABox
    \Bmc is defined as expected in case of $\DLR$ ontologies: View
    $C(b)$ as a tree-shaped ABox $\Amc_{C(b)}$ with root $b$ and
    assume without loss of generality that $b$ is the only individual
    shared by $\Bmc$ and $\Amc_{C(b)}$. Then take the union of \Amc
    and $\Amc_{C(b)}$. 

    By the following claim, the answer to the modified membership
    query coincides with that to the original query.

    \medskip
    \noindent\textit{Claim 1.} $\Amc', \Omc' \models q(a)$ iff $\Amc, \Omc
    \models q(a)$ for all ELIQs $q$ that only use symbols from $\Sigma$.
    
    \medskip

    \noindent\textit{Proof of Claim 1.}
    For ``if'', suppose that $\Amc, \Omc \models q(a)$ and let $\Imc$
    be a model of $\Amc'$ and $\Omc'$. We can assume that
    $\Delta^\Imc$ does not mention any of the individuals that were
    introduced in the construction of \Amc. We will construct a model
    $\Imc'$ of $\Amc$ and \Omc that has a homomorphism $h$ from
    $\Imc'$ to \Imc which is the identity on $\Delta^\Imc$. 
    This clearly suffices since $\Imc'\models q(a)$. 

    The interpretation $\Imc'$ has the following domain:
    \begin{align*}
      \Delta^{\Imc'} = \Delta^\Imc \cup \bigcup_{X_C(b)\in \Amc'}\mn{ind}(\Amc_{C(b)})
    \end{align*}
    In order to define the interpretation of concept and role names,
    observe first that, for every $X_C(b)\in \Amc'$, there is a homomorphism $h_{C(b)}:\Amc_{C(b)},b\to \Imc,b$ since $\Imc$ is a model
    of $\Amc'$ and $\Omc'$, and $\Omc'\models X_{C}\sqsubseteq C$. We combine all these homomorphisms into a mapping
    $h:\Delta^{\Imc'}\to \Delta^\Imc$ by taking
    \[h(c) = \begin{cases} c & \text{if }c\in \Delta^\Imc, \\ 
	h_{C(b)}(c) &\text{if }c\in
	\mn{ind}(\Amc_{C(b)})\setminus\Delta^\Imc.
      \end{cases}
    \] %
    Then, we set
    \begin{align*}
      A^{\Imc'} &= \{d\mid h(d)\in A^\Imc\} \\
      r^{\Imc'} &= \{(d,e)\mid (h(d),h(e)\in r^\Imc\}
    \end{align*}
    It is routine to verify that $\Imc'$ is as required.
 
    For ``only if'', suppose that $\Amc',\Omc'\models q(a)$ and let
    $\Imc$ be a model of $\Amc$ and \Omc. Observe that the model
    $\Imc'$ of $\Omc'$ that can be obtained from \Imc as in
    Lemma~\ref{lem:frontier-normal-form-R} Point~2 coincides with
    $\Imc$ on $\Sigma$ and is additionally a model of $\Amc'$. It
    follows that $\Imc\models q(a)$ as required. This finishes the
    proof of Claim~1.

    \medskip
    In the case of \DLFsynr ontologies, we follow the same strategy.
    However, the addition of $\Amc_{C(b)}$ in $(\ast)$
    has to respect the functionality assertions. In fact, not even
    $\Amc_{C(b)}$ necessarily satisfies the functionality assertions
    in \Omc. We define the
    \emph{addition of a tree-shaped ABox $\Bmc$ with root $b_0$ to $\Amc$ at
    $a$} inductively on the structure of $\Bmc$ as follows:
    \begin{enumerate}

      \item for all $A(b_0)\in \Bmc$, add $A(a)$ to \Amc;

      \item for all $R(b_0,b')\in \Bmc$, let $\Bmc'$ be the sub-ABox of
	$\Bmc$ rooted at $b'$ and 

      \begin{enumerate}

	\item if $\mn{func}(R)\in \Omc$ and there is an atom
	  $R(a,a')$ in \Amc, then add $\Bmc'$ to $\Amc$ at $a'$;

	\item otherwise, add an atom $R(a,a')$ for a fresh individual
	  $a'$ and add $\Bmc'$ to \Amc at $a'$. 

      \end{enumerate}

    \end{enumerate}
    Note that there can only be one atom $R(a,a')$ in Step~2(a) if 
    $\Amc$ satisfies the functionality assertions in \Omc. It should
    be also clear that the resulting ABox also satisfies the functionality
    assertions if $\Amc$ does.

    Now, $\Amc$ is obtained from $\Amc'$ by starting with $\Amc=\Amc'$
    and 
    \begin{itemize}

      \item[$(\ast')$] adding $\Amc_{C(b)}$ to $\Amc$ at $b$, for each
	$X_C(b)\in \Amc'$.

    \end{itemize}

    By the following claim, the answer to the modified membership
    query coincides with that to the original query.

    \medskip
    \noindent\textit{Claim~2.} $\Amc', \Omc' \models q(a)$ iff $\Amc, \Omc
    \models q(a)$ for all ELIQs $q$ that only use symbols from $\Sigma$.
    
    \medskip

    \noindent\textit{Proof of Claim~2.}
    For ``if'', suppose that $\Amc, \Omc \models q(a)$ and let $\Imc$
    be a model of $\Amc'$ and $\Omc'$. We can assume that
    $\Delta^\Imc$ does not mention any of the individuals that were
    introduced in the construction of \Amc. We will construct a model
    $\Imc'$ of $\Amc$ and \Omc that has a homomorphism $h$ from
    $\Imc'$ to \Imc which is the identity on $\Delta^\Imc$. 
    This clearly suffices since $\Imc'\models q(a)$. 

    Observe first that, for every $X_C(b)\in \Amc'$, there is a
    homomorphism $h_{C(b)}:\Amc_{C(b)},b\to \Imc,b$ since $\Imc$ is a
    model of $\Amc'$ and $\Omc'$, and $\Omc'\models X_{C}\sqsubseteq
    C$.

    Let $F$ denote the set of fresh individuals introduced in the
    construction of $\Amc$. Note that for every fresh element there is
    an $X_{C}(b)\in \Amc'$ which `triggered' the addition of $d$ in
    some (possibly later) application of Step~2(b). We associate with
    every $d\in F$ an element $g(d)\in \Delta^{\Imc}$ as follows: 

    \begin{itemize}

      \item if $d$ was introduced in Step~2(b) triggered by
	$X_{C}(b)\in \Amc'$, then set $g(d)=h_{C(b)}(b')$ where
	$b'$ is the element mentioned in Step~2.

    \end{itemize}

    We further associate with every $d\in F$ a tree-shaped
    interpretation $\Imc_d$. 
    Intuitively, $\Imc_d$ is the unraveling of \Imc at $g(d)$, with
    the functionality assertions taken into account. Formally, 
    the domain $\Delta^{\Imc_d}$
    consists of all sequences $a_0R_1a_1\ldots R_na_n$ such that 
    \begin{itemize}

      \item $a_0=g(d)$; 

      \item $a_i\in \Delta^\Imc$, for all $i$ with $0\leq i\leq
	n$;

      \item $(a_i,a_{i+1})\in R_{i+1}^{\Imc}$, for all $i$ with
	$0\leq i<n$;

      \item if $\mn{func}(R_i^-)\in \Omc$, then $R_{i+1}\neq R_i^-$,
	for all  $i$ with
	$0\leq i<n$;

      \item if $\mn{func}(R_1)\in \Omc$, there is no atom
	of shape $R_1(d,d')$ in $\Amc_{C(b)}$, where $X_C(b)\in \Amc'$ triggered
	the addition of $d$.

    \end{itemize}
    The interpretation of concept and role names is as follows:
    \begin{align*}
      A^{\Imc_d} & = \{a_0R_1a_1\ldots R_na_n\in \Delta^{\Imc_d}\mid
      a_n\in A^{\Imc}\} & \text{for all $A\in\mn{N_C}$;} \\
	r^{\Imc_d} & = \{(\pi,\pi r a)\mid \pi r a\in
	  \Delta^{\Imc_d}\}\cup{}\\
	  & \phantom{ {} = {}} \{(\pi r^- a,\pi)\mid \pi r^- a\in
	    \Delta^{\Imc_d}\} & \text{for all $r\in\mn{N_R}$}.
    \end{align*}
    The interpretation $\Imc'$ is then obtained by starting with
    $\Imc' = \Imc\cup \Amc$, and then adding, for every $d\in F$, 
    a copy $\widehat{\Imc_d}$ of $\Imc_d$ and gluing the copy of $g(d)$
    in $\widehat{\Imc_d}$ to $d$.

	  It is routine to verify that $\Imc'$ is as required.

	  \smallskip
	  For ``only if'', suppose that $\Amc',\Omc'\models q(a)$ and let
	  $\Imc$ be a model of $\Amc$ and \Omc. Observe that the model
	  $\Imc'$ of $\Omc'$ that can be obtained from \Imc as in
	  Lemma~\ref{lem:frontier-normal-form-R} Point~2 coincides with
	  $\Imc$ on $\Sigma$ and is additionally a model of $\Amc'$. It
	  follows that $\Imc\models q(a)$ as required. This finishes the
	  proof of Claim~2. 
%
%
%
%
%
%
\end{proof}

We now work towards showing that the algorithm presented in
Section~\ref{sec:learning} indeed learns ELIQs under ontologies
formulated in \DLR or \DLFsynr, in polynomial time. We start with
analyzing the \mn{minimize} subroutine.

\begin{restatable}{lemma}{lemminimize}
  \label{lem:minimize}
  Let $q$ be a unary CQ that is $\Omc$-saturated and satisfiable w.r.t.\
  $\Omc$ such that $q \subseteq q_T$ for the target query $q_T(x_0)$, and let $q'(x_0) = \mn{minimize}(q)$. Then 
  \begin{enumerate}
        \item $q \subseteq_\Omc q'$ and $q' \subseteq_\Omc q_T$;
        \item $|\mn{var}(q')| \leq |\mn{var}(q_T)|$;
        \item $q'$ is \Omc-minimal, connected, and \Omc-saturated.
    \end{enumerate}
\end{restatable}

\noindent
\begin{proof}\ We start with Point~1. We have $q \subseteq_\Omc q'$
  since $q'$ is a subset of $q$. For $q' \subseteq_\Omc q_T$, it
  suffices to observe that \mn{minimize} ensures in each step that
  $\Amc_{q'}, \Omc \models q_T(x_0)$.

  For Point~2, it suffices to show that
  $\mn{var}(q') \subseteq \mn{img}(h)$ for every homomorphism $h$ from
  $q_T$ to $\Umc_{q', \Omc}$ with $h(x_0) = x$. Assume for a
  contradiction that there is a homomorphism $h$ from $q_T$ to
  $\Umc_{q', \Omc}$ with $h(x_0) = x_0$ and a $y \in \mn{var}(q')$
  that is not in $\mn{img}(h)$.  Choose some $r(x,y) \in q$ such that
  the distance from $x_0$ to $x$ is strictly smaller than that from
  $x_0$ to $y$, and let $q^-$ be the restriction of
  $q \setminus \{ r(x,y) \}$ to the atoms that contain only variables
  reachable from $x_0$ in $q \setminus \{ r(x,y) \}$. We argue that
  $h$ is a homomorphism from $q_T$ to $\Umc_{q^-, \Omc}$ which
  witnesses that $\Amc_{q^-}, \Omc \models q_T(x_0)$, in contradiction
  to the construction of $q'$ and $y \in \mn{var}(q')$.

  To see that $h$ is a homomorphism, first note that since $q_T$ is
  connected and $h(x_0)=x_0$, the range of $h$ contains only variables
  from $\mn{var}(q^-)$ and the subtrees below them that (consist of
  traces and) are added in the construction of the universal model. Next
  observe that for all $x_1, x_2 \in \mn{var}(q^-)$, the following
  holds by construction of universal models and since $q$ is
  $\Omc$-saturated:
  \begin{enumerate}
        \item[(a)] $A(x_1) \in \Umc_{q', \Omc}$ iff $A(x_1) \in \Umc_{q^-, \Omc}$;
        \item[(b)] $r(x_1, x_2) \in \Umc_{q', \Omc}$ iff $r(x_1, x_2) \in \Umc_{q^-, \Omc}$.
        \end{enumerate}
        Since $\Omc$ is in normal form and due to (a), it follows from
        the construction of universal models that the subtree in
        $\Umc_{q', \Omc}$ below each
        $x_1 \in \mn{var}(q^-) \setminus \{y\}$ is identical to the
        subtree in $\Umc_{q^-, \Omc}$ below $x_1$. Moreover, the
        subtree in $\Umc_{q', \Omc}$ below $y$ can be obtained from
        the subtree in $\Umc_{q^-, \Omc}$ below $y$ by dropping
        subtrees.  It should thus be clear that, as required, $h$ is a
        homomorphism from $q_T$ to $\Umc_{q^-, \Omc}$.

        For Point~3, we start with \Omc-minimality. Assume for a
        contradiction that $q'$ is not \Omc-minimal, that is, there is
        a homomorphism $h$ from $q'$ to $\Umc_{q'', \Omc}$ with
        $h(x_0) = x_0$ where $q''=q|_{\mn{var}(q) \setminus \{ y \}}$
        for some variable $y \in \mn{var}(q)$. Choose some
        $r(x,y) \in q$ such that the distance from $x_0$ to $x$ is
        strictly smaller than that from $x_0$ to $y$ and let $q^-$ be
        the restriction of $q \setminus \{ r(x,y) \}$ to the atoms
        that contain only variables reachable from $x_0$ in
        $q \setminus \{ r(x,y) \}$. We can show as above that $h$ is a
        homomorphism from $q_T$ to $\Umc_{q^-, \Omc}$ and thus
        $\Amc_{q^-}, \Omc \models q_T(x_0)$, in contradiction to the
        construction of $q'$ and $y \in \mn{var}(q')$.

        Now for connectedness. Assume for a contradiction that
        $q'(x_0)$ is not connected and let $x$ be a variable that is
        in a different maximally connected component of $q'$ than
        $x_0$. Then $x$ is also in a different maximally connected
        component of $\Umc_{q',\Omc}$ than $x_0$.  By Point~1, there
        is a homomorphism $h$ from $q_T$ to $\Umc_{q',\Omc}$ with
        $h(x_0)=x_0$. Since $q_T$ is connected, we must have
        $x \notin\mn{img}(h)$, thus
        $\mn{var}(q') \not\subseteq \mn{img}(h)$. But we have already
        seen in the proof of Point~2 that this is impossible.

        Finally, \Omc-saturatedness of $q'$ is clear given that the
        original CQ $q$ is \Omc-saturated, $q'$ is a subquery of $q$,
        and during the construction of $q'$ we have not removed any
        concept atoms on any of the remaining variables.
\end{proof}
We next turn to the \mn{treeify} subroutine. We start with a
preliminary.
\begin{definition}
    An \emph{\ELI-simulation} from interpretation $\Imc_1$ to interpretation
    $\Imc_2$ is a relation $S \subseteq \Delta^{\Imc_1} \times
    \Delta^{\Imc_2}$ such that for all $(d_1,d_2)\in S$, we have:
    \begin{enumerate}

      \item for all $A \in \NC$: if $A(d_1) \in \Imc_1$, then $A(d_2) \in S$;

      \item for all $r \in \NR$ and $R\in\{r,r^-\}$: if there is some
	$d_1'\in\Delta^{\Imc_1}$ with $R(d_1, d'_1) \in
	\Imc_1$, then there is $d'_2 \in \Delta^{\Imc_2}$ such that
	$(d'_1, d'_2) \in S$ and $R(d_2, d_2') \in \Imc_2$.

    \end{enumerate}
\end{definition}
The following lemma gives an important property of simulations.
The proof is standard and omitted.
\begin{lemma}\label{lem:simulation-implication}
    Let $\Omc$ be a $\DLRF$ ontology, $\Amc_1$, $\Amc_2$ ABoxes and $q(x)$ an  
    ELIQ such that $\Amc_1$, $\Amc_2$, and $q$ are satisfiable w.r.t\ $\Omc$.  
    If there is an \ELI-simulation $S$ from $\Amc_1$ to $\Amc_2$ with $(a_1,  
    a_2) \in S$, then $\Amc_1, \Omc \models q(a_1)$ implies $\Amc_2, \Omc \models q(a_2)$.  
\end{lemma}
\begin{restatable}{lemma}{lemtreeifyiter}
  \label{lem:treeify-iter}
  Let $q(x_0)$ be a unary CQ that is $\Omc$-saturated and satisfiable w.r.t.\
  $\Omc$ such that $q \subseteq_\Omc q_T$ for the target query $q_T(x_0)$.
  Further let  $p_1(x_0),p_2(x_0),\dots$
be the sequence of CQs computed by $\mn{treeify}(q)$. Then for all $i \geq 1$,
    \begin{enumerate}
        \item $p_i \subseteq_\Omc q_T$;
        \item $|\mn{var}(p_{i + 1})| > |\mn{var}(p_i)|$;
        \item $p_i$ is $\Omc$-saturated and satisfiable w.r.t.\
          $\Omc$.
    \end{enumerate}
\end{restatable}
\noindent
\begin{proof}\ 
  We show Point~1 by induction on $i$.  The case $i = 1$ is immediate
  by Point~1 of Lemma~\ref{lem:minimize} since $p_1 = \mn{minimize}(q)$.  Now let
  $i \geq 1$. By the induction hypothesis $p_i \subseteq q_T$ and thus
  $\Amc_{p_i}, \Omc \models q_T(x_0)$.  
  By construction of $p_i'$,
\[
    S = \{ (x, x) \mid x \in \mn{var}(p_i)\} \cup \{ (x, x') \mid x \in \mn{var}(p_i)\}
\]
is an \ELI-simulation from $\Amc_{p_i}$ to $\Amc_{p_i'}$ with
$(x_0, x_0) \in S$.  Therefore, by
Lemma~\ref{lem:simulation-implication},
$\Amc_{p_i'}, \Omc \models q_T(x_0)$ and $p_i' \subseteq_\Omc q_T$.
Hence, by Point~1 of Lemma~\ref{lem:minimize},
$p_{i + 1} \subseteq_\Omc q_T$ for $p_{i + 1} = \mn{minimize}(p_i')$.

For Point~2, define a homomorphism $g$ from $\mn{var}(p_{i + 1})$ to $\mn{var}(p_i)$
by setting $g(x) = x$ for all $x \in \mn{var}(p_i) \cap \mn{var}(p_{i + 1})$
and $g(x') = x$ for all $x' \in \mn{var}(p_{i + 1}) \setminus \mn{var}(p_i)$.
To establish that $|\mn{var}(p_{i + 1})| > |\mn{var}(p_i)|$, we show
the following three claims, proving that $g$ is surjective, but not
injective.  For an injective and surjective function, we use $g^-$ to
denote the inverse of $g$.

\smallskip
\noindent\textit{Claim 1.} $g$ is surjective.

\smallskip
\noindent\textit{Proof of Claim 1.} 
Suppose that $g$ is not surjective.  Then some $y \in \mn{var}(p_i)$
does not occur in the image of $g$. Choose some $r(x,y)\in p_i$ and
define $p_{i}^- = p_i\setminus\{r(x,y)\}$. Clearly, $g$ is still a
homomorphism from $p_{i+1}$ to $p_i^-$.
By Lemma~\ref{lem:extend-hom}, we can extend $g$ to be a
homomorphism from $\Umc_{p_{i + 1}, \Omc}$ to $\Umc_{p^-_i, \Omc}$.
By Point~1, there is a homomorphism $h$ from $q_T$ to
$\Umc_{p_{i + 1}, \Omc}$ with $h(x_0) = x_0$.  Composing $h$ and $g$
yields a homomorphism $g'$ from $q_T$ to $\Umc_{p^-_i, \Omc}$ with
$g'(x_0) = x_0$. Thus, $\Amc_{p_i^-},\Omc \models q_T(x_0)$ which is in
contradiction to the fact that $p_i$ is obtained by applying
\mn{minimize}, and that this operation would replace $p_i$ with
$p_i^-$.

\smallskip
\noindent\textit{Claim 2.} If $g$ is injective, then $r(x, y) \in p_i$ implies
$r(g^-(x), g^-(y)) \in p_{i + 1}$.

\smallskip
\noindent\textit{Proof of Claim 2.} 
Suppose to the contrary that there is an $r(x, y) \in p_i$ with
$r(g^-(x), g^-(y)) \notin p_{i + 1}$. Then $g$ is also a
homomorphism from $p_{i + 1}$ to $p_i \setminus \{r(x, y)\}$ and
using the same composition-of-homomorphisms argument as in the proof
of Claim~1, we find a homomorphism $h$ from $q_T$ to
$\Umc_{p_i \setminus \{r(x, y)\}, \Omc}$ with $h(x_0) = x_0$. Hence
$p_i \setminus \{ r(x, y)\} \subseteq_\Omc q_T$. This contradicts
the fact that $p_i = \mn{minimize}(p'_i)$.

\smallskip
\noindent\textit{Claim 3.} $g$ is not injective.

\smallskip
\noindent\textit{Proof of Claim 3.} 
Let $R_1(x_1, x_2), \dots, R_n(x_n, x_1)$ be a cycle in $p_i$ and $R_n(x_n,
x_1)$ the atom that was chosen in the cycle doubling operation.
Suppose for contradiction that $g$ is injective. The construction of $g$,
together with $g$ being surjective and injective, implies that exactly one of
$x_j, x_j'$ is in $\mn{var}(p_{i + 1})$ for all $j$ with $1 \leq j \leq n$.
Assume that $x_n \in \mn{var}(p_{i + 1})$ (the case $x_n' \in \mn{var}(p_{i +
1})$ is analogous) and thus $g(x_n) = x_n$.

We prove by induction on $j$ that $x_j \notin \mn{var}(p_{i + 1})$ for $1 \leq
j \leq n$, thus obtaining a contradiction to $x_n \in \mn{var}(p_{i + 1})$.
For the induction start, assume to the contrary of what is to be shown that
$x_1 \in \mn{var}(p_{i + 1})$. Then $g(x_1) = x_1$ and $R_n(x_n, x_1) \in p_i$
implies $R_n(x_n, x_1) \in p_{i + 1}$ by Claim~2. 
This contradicts the construction of $p_{i + 1}$ that removes $R(x_n, x_1)$.

For the induction step, let $j \geq 1$. By the induction hypothesis $x_j \notin
\mn{var}(p_{i + 1})$ and thus $x_j' \in \mn{var}(p_{i + 1})$. Assume to the contrary of what is shown that $x_{j + 1} \in \mn{var}(p_{i
+ 1})$. Then $g(x_{j + 1}) = x_{j + 1}$, $g(x_j') = x_j$ and $R_j(x_j, x_{j + 1}) \in p_i$ yield
$R_j(x_j', x_{j + 1}) \in p_{i + 1}$ by Claim~2. 
This contradicts the construction of $p_{i + 1}$.

This completes the proof of Claim~3 and thus Point~2 of the lemma.
\smallskip

We show Point~3 of the lemma by induction on $i$.  In the induction start,
\Omc-saturatedness and satisfiability of $p_1$ w.r.t.\ \Omc follows from
\Omc-saturatedness and satisfiability of $q$ w.r.t.\ \Omc and the fact that
$\mn{minimize}$ preserves those properties.

Now let $p_i$ be \Omc-saturated and satisfiable w.r.t.\ \Omc. Then $p_i'$ is
also \Omc saturated, by construction. Moreover, we can construct a model
$\Imc$ of $\Amc_{p_i'}$ and $\Omc$ by starting with $\Imc = \Amc_{p_i'}$
and attaching the trace subtrees below every $x \in \mn{ind}(\Amc_{p_i})$ in
$\Umc_{\Amc_{p_i}, \Omc}$ to $x$ and $x'$ in $\Imc$, and then adding $(d, e)$
to $r^\Imc$ whenever $(d, e) \in s^\Imc$ and $\Omc \models s
\sqsubseteq r$. The resulting interpretation \Imc is a model of
$\Amc_{p_i'}$ and $\Omc$. In particular, \Imc satisfies all
functionality assertions in \Omc, because every element in
$\Delta^\Imc$ has the same number of $r$-successors and $r$-predecessors as
its corresponding original element in $\Umc_{\Amc_{p_i}, \Omc}$.

Since $\mn{minimize}$ preserves \Omc-saturatedness and satisfiability w.r.t.\
\Omc, $p_{i + 1}$ is therefore also \Omc-saturated and satisfiable w.r.t.\
\Omc.
\end{proof}

Point~2 of Lemma~\ref{lem:treeify-iter} and 
Point~2 of Lemma~\ref{lem:minimize} imply that \mn{treeify} terminates 
and thus eliminates all cycles in $q$ while maintaining 
$q \subseteq_\Omc q_T$.  The next lemma makes this precise.

\begin{restatable}{lemma}{lemtreeterm}
\label{lem:treeterm}
Let 
$q$ be a CQ that is $\Omc$-saturated, satisfiable w.r.t.\ $\Omc$,
and satisfies $q \subseteq_\Omc q_T$. Then $q' = \mn{treeify}(q)$
is an ELIQ. Moreover,  $\mn{treeify}(q)$ runs in time polynomial in
$|\mn{var}(q_T)| + ||q|| + ||\Omc||$.
\end{restatable}

\noindent
\begin{proof}\ Let $p_1, p_2, \dots,$ be the sequence of constructed
  queries.  Recall that for all $i \geq 1$, $p_i$ is the result of
  applying \mn{minimize}. Thus, by Point~2 of Lemma~\ref{lem:minimize},
  $|\mn{var}(p_i)| \leq |\mn{var}(q_T)|$ for all $i \geq i$. But by
  Point~2 of Lemma~\ref{lem:treeify-iter}, 
  $p_{i+1}$ has more variables than $p_i$, for every $i$.
  It follows that the length $n$ of the sequence of queries is at most
  $|\mn{var}(q_T)|$ and thus \mn{treeify} stops at $p_n =
  \mn{treeify}(q)$. It also follows that $p_n$ does not contain a
  cycle. Moreover, $p_n$ is obtained by applying \mn{minimize}
  and thus connected due to Point~3 of Lemma~\ref{lem:minimize}.
  Thus, $p_n$ is an
  ELIQ.

  It remains to argue that the running time is as claimed in the
  lemma. A cycle in $p_i$ can be identified in time polynomial in
  $|\mn{var}(p_i)| \leq |\mn{var}(q_T)|$. Moreover, each call
  $\mn{minimize}(p'_i)$ makes at most
  $||p'_i||$ membership queries. It suffices to note that
  $||p'_i|| \leq 2 \cdot ||p_{i}||$ and that $||p_i|| \leq ||\Omc|| \cdot
  ||q_T||$ since $\mn{var}(p_i) \leq \mn{var}(q_T)$ and $p_i$ uses
  only symbols from \Omc.
\end{proof}
We now analyze the main part of the algorithm. To this end, let
$q_1(x_0),q_2(x_0),\dots$ be the sequences of hypotheses $q_H$ that the
algorithm constructs. The following lemma summarizes their
most important properties.
\begin{restatable}{lemma}{lemalgiter}
\label{lem:alg-iter}
    For all $i \geq 1$,
    \begin{enumerate}
        \item $q_i \subseteq_\Omc q_T$;
        \item $q_i \subseteq_\Omc q_{i + 1}$ and $q_{i + 1} \not\subseteq_\Omc q_i$;
        \item $\mn{var}(q_i) \subseteq \mn{img}(h)$ for every homomorphism $h$
          from $q_{i + 1}$ to $\Umc_{q_i, \Omc}$ with $h(x) = x$.
    \end{enumerate}
\end{restatable}

\noindent
\begin{proof}\ Point~1 is proved by induction on $i$. For $i = 1$,
  $q_1 = \mn{treeify}(q_H^0)$ and $q_H^0 \subseteq_\Omc q_T$ imply
  $q_1 \subseteq_\Omc q_T$ by Point~3 of Lemma~\ref{lem:treeify-iter}.
  For $i > 1$, recall that $q_{i} = \mn{minimize}(q_F)$ for some
  $q_F \subseteq_\Omc q_T$. Thus, $q_{i} \subseteq_\Omc q_T$ follows by
  Point~1 of Lemma~\ref{lem:minimize}.

  Point~2 follows from the fact that $q_{i+1} = \mn{minimize}(q_F)$ for $q_F$
  some element of a frontier of $q_i$ w.r.t.\ \Omc. By definition of frontiers
  and Point~1 of Lemma~\ref{lem:minimize} it follows that $q_i \subseteq_\Omc
  q_{i + 1}$ and $q_{i + 1} \not\subseteq_\Omc q_i$, as required.

  For Point~3, let $h$ be a homomorphism from $q_{i + 1}$ to
  $\Umc_{q_i, \Omc}$ with $h(x_0) = x_0$. Assume to the contrary of
  what is to be shown that some $y \in \mn{var}(q_i)$ does not occur
  in the image of $h$. Choose some $r(x,y) \in q_i$ such that the
  distance from $x_0$ to $x$ is strictly smaller than that from $x_0$
  to $y$ and let $q_i^-$ be the restriction of
  $q_i \setminus \{ r(x,y) \}$ to the atoms that contain only
  variables reachable from $x_0$ in $q_i \setminus \{ r(x,y) \}$.  We
  can argue as in the proof of Lemma~\ref{lem:minimize} that $h$ is a
  homomorphism from $q_{i+1}$ to $\Umc_{q^-_i, \Omc}$.
By
  Lemma~\ref{lem:extend-hom}, we can extend $h$ to be a homomorphism
  from $\Umc_{q_{i + 1}, \Omc}$ to $\Umc_{q^-_i, \Omc}$.  By Point~1,
  there is a homomorphism $g$ from $q_T$ to $\Umc_{q_{i + 1}, \Omc}$
  with $h(x_0) = x_0$.  Composing $g$ and $h$ yields a homomorphism
  $g'$ from $q_T$ to $\Umc_{q^-_i, \Omc}$ with $g'(x_0) = x_0$. Thus
  $\Amc_{q_i^-},\Omc \models q_T(x_0)$ which is in contradiction to the
  fact that $q_i$ is obtained by applying \mn{minimize}, and that this
  operation would replace
  $q_i$ with $q_i^-$.
\end{proof}
It remains to show that the algorithm terminates after polynomially
many steps. 
\begin{restatable}{lemma}{lemalgterm}
    \label{lem:alg-polynom}
    $q_n \equiv q_T$ for some $n \leq p(|\mn{var}(q_T)| + ||\Omc||)$,
    with $p$ a
    polynomial.
\end{restatable}

\noindent
\begin{proof}\ Every ELIQ $q_i$, with $i \geq 1$, is \Omc-saturated
  and satisfiable w.r.t.\ \Omc. This is easy to prove by induction on
  $i$. The induction start follows from the fact that the seed CQ is
  satisfiable w.r.t.\ \Omc and \Omc-saturated, and from Point~3
  of~Lemma~\ref{lem:treeify-iter}. The induction step follows from
  Point~3 of Lemma~\ref{lem:minimize}.

  Point~2 of Lemma~\ref{lem:minimize} thus yields 
  $|\mn{var}(q_i)| \leq |\mn{var}(q_T)|$ for all $i \geq 1$.
  Moreover, Point~3 of Lemma~\ref{lem:alg-iter} implies that
  $|\mn{var}(q_i)| \leq |\mn{var}(q_{i + 1})|$.  Hence, it remains to
  show that the length of any subsequence $q_j, \dots, q_k$ with
  $|\mn{var}(q_j)| = \dots = |\mn{var}(q_k)|$ is bounded by a
  polynomial in $\mn{var}(q_T)+||\Omc||$.

   Let $h_\ell$ for $\ell \in \{j, \dots, k - 1\}$ be the homomorphism from
   $q_{\ell + 1}$ to $\Umc_{q_\ell, \Omc}$ that exists due to
   Point~2 of Lemma~\ref{lem:alg-iter}.
   Since $|\mn{var}(q_\ell)| = |\mn{var}(q_{\ell + 1})|$ and by
   Point~3 of Lemma~\ref{lem:alg-iter}, $h_\ell$ is a
   bijection between $\mn{var}(q_{\ell + 1})$ and $\mn{var}(q_\ell)$.
   Also by Point~2 of Lemma~\ref{lem:alg-iter}, $h^-$ is not a homomorphism from
   $q_\ell$ to $\Umc_{q_{\ell + 1},\Omc}$.

   Therefore, one of the following two cases applies:
   \begin{enumerate}

   \item there is a concept atom $A(x_1) \in q_\ell$ such that
     $A(h_\ell^-(x_1)) \notin \Umc_{q_{\ell + 1},\Omc}$ or

   \item there is a role atom $r(x_1, x_2) \in q_\ell$ such that
     $r(h_\ell^-(x_1), h_\ell^-(x_2)) \notin \Umc_{q_{\ell + 1},\Omc}$.
     
   \end{enumerate}
   We show that each case can occur at most
   polynomially often in $\mn{var}(q_T)+||\Omc||$.

   We start with Case~1. It follows from the fact that $h$ is a
   homomorphism that whenever $A(h_\ell^-(x_1)) \in q_{\ell+1}$, then
   $A(x_1) \in \Umc_{q_\ell,\Omc}$. Since $q_\ell$ is \Omc-saturated,
   this implies $A(x_1) \in q_\ell$. Moreover,
   $A(h_\ell^-(x_1)) \notin \Umc_{q_{\ell + 1},\Omc}$ implies
   $A(h_\ell^-(x_1)) \notin q_{\ell + 1}$.  Consequently, $q_{\ell}$
   contains at least one concept atom more than $q_{\ell + 1}$. Thus,
   Case~1 can occur as most as often as there are concept atoms in
   $q_1$, and this number is bounded by
   $|\mn{var}(q_T)| \cdot ||\Omc||$ since
   $|\mn{var}(q_1)| \leq |\mn{var}(q_T)|$ and $q_1$ may only use
   symbols from \Omc.
   
   In Case~2, consider the unique role atom
   $s(h_\ell^-(x_1),h_\ell^-(x_2)) \in q_{\ell+1}$.  Since $h_\ell$ is
   a homomorphism, $s(x_1,x_2) \in \Umc_{q_\ell,\Omc}$. From
   $r(x_1,x_2) \in q_\ell$ and the construction of universal models,
   it follows that $\Omc \models r \sqsubseteq s$. From
   $r(h_\ell^-(x_1), h_\ell^-(x_2)) \notin \Umc_{q_{\ell + 1},\Omc}$,
   it follows that $\Omc \not\models s \sqsubseteq r$. Thus, Case~2
   can occur at most $||\Omc||$ times for each role atom in $q_1$.
   Since $q_1$ is a tree, $|\mn{var}(q_1)| \leq |\mn{var}(q_T)|$, and
   $q_1$ may only use symbols from \Omc, the number of such atoms
   is bounded by $|\mn{var}(q_T)| \cdot ||\Omc||$.
\end{proof}

\thmlower*

\noindent
\begin{proof}\
    To prove the theorem, we use a proof strategy that is inspired by basic lower
    bound proofs for abstract learning problems due to Angluin
    \cite{DBLP:journals/ml/Angluin87}. 
    Essentially the same proof is given in \cite{FJL-IJCAI21} for a slightly
    different class of ontologies that allows only concept inclusions between
    arbitrary conjunctions of concept names.

    Here, it is convenient to view the oracle as an adversary who maintains a
    set $S$ of candidate target queries that the learner cannot distinguish
    based on the queries made so far.
    We have to choose $S$ and the ontology carefully so that each membership
    query removes only few candidate targets from $S$ and that after a
    polynomial number of queries there is still more than one candidate that
    the learner cannot distinguish.

    For each $n \geq 1$, let
\[
    \Omc_n = \{ A_i \sqcap A_i' \sqsubseteq \bot \mid 1 \leq i \leq n\}
\]
and
$$
\begin{array}{r@{\;}c@{\;}l}
    S_n &=& \{ q(x) = \alpha_1(x) \land \dots \land \alpha_n(x)
            \mid \\[1mm]
  && \qquad \alpha_i \in \{ A_i, A_i'\} \text{ for all $i$ with } 1 \leq i \leq n\}.
\end{array}
$$
Note that $S_n$ is a frontier of $\bot$ w.r.t.\ $\Omc_n$, if only AQ$^\land$
queries using the concept names $A_i$ and $A_i'$ for all $1 \leq i \leq n$, are
considered for Condition~3. Clearly, $S_n$ contains $2^n$ queries.\footnote{In
    fact, it can be shown similar as in the proof of
    Theorem~\ref{thm:lowerfrontier}
that $S_n$ is contained in any frontier
of $\bot$ w.r.t.\ $\Omc_n$. Hence, $\bot$ does not have polynomially
sized frontiers w.r.t.\ disjointness ontologies.}

Assume to the contrary of what is to be shown that AQ$^\land$ queries are
learnable under disjointness ontologies using only polynomially many membership
queries.
Then there exists a learning algorithm and polynomial $p$ such that the number
of membership queries needed to identify a target query $q_T$ is bounded by
$p(n_1, n_2)$, where $n_1$ is the size of $q_T$ and $n_2$ is the size of the
ontology. We choose $n$ such that $2^n > p(r_1(n), r_2(n))$, where $r_1$ is a
polynomial such that every query $q \in S_m$ satisfies $||q|| = r_1(m)$
and $r_2$ is a polynomial such that $r_2(m) > ||\Omc_m||$ for every $m \geq 1$.

Now, consider a membership query posed by the learning algorithm with ABox and
answer individual $(\Amc, a)$. The oracle responds as follows:
\begin{enumerate}
    \item if $\Amc, \Omc_n \models q(a)$ for no $q \in S_n$, then answer \emph{no};
    \item if $\Amc, \Omc_n \models q(a)$ for a single $q \in S_n$, then answer \emph{no} and remove $q$ from $S_n$;
    \item if $\Amc, \Omc_n \models q(a)$ for more than one $q \in S_n$, then answer \emph{yes}.
\end{enumerate}
Note that the third response is consistent since $\Amc$ must then contain
$A_i(a)$ and $A_i'(a)$ for some $i$ and thus $\Amc$ is not satisfiable w.r.t.\
$\Omc_n$. Moreover, the answers are always correct with respect to the updated
set $S_n$. Thus, the learner cannot distinguish the remaining candidate queries
by answers to queries posed so far.

It follows that the learning algorithm removes at most $p(r_1(n),
r_2(n))$ many
queries from $S_n$. By the choice of $n$, at least two candidate concepts remain in
$S_n$ after the algorithm is finished. Thus, the learner cannot distinguish
between them and we have derived a contradiction.
\end{proof}

\thmnolearn*

\noindent
\begin{proof}
  \ We use the same ontology $\Omc$ as in the proof of
  Theorem~\ref{thm:no-frontier}, that is, 
  \[ \Omc = \{\ A \sqsubseteq \exists r,\quad \exists r^- \sqsubseteq
  \exists r,\quad \exists r \sqsubseteq \exists s,\quad
\mn{func}(r^-)\ \}.  \]
To show that ELIQs are not learnable under \Omc using only membership
queries we use an infinite set \Hmc of hypotheses (i.e.,
candidate target queries) that cannot be distinguished
by a finite number of membership queries.  Let
\[\Hmc=\{q^*\}\cup\{q_n\mid n \text{ prime}
  \}
  ,\] where $q^*(x_1) = \{
  A(x_0),r(x_0,x_1),A(x_1)\}$ and
  each $q_n$ is defined as follows: 
%
  %
%
  %
  \begin{align*} 
    q_n(x_1) = \{\ & A(x_0), r(x_0, x_1), r(x_1, x_2), \dots, r(x_{n - 1},
    x_n), \\ 
    & s(x_n, y), s(x'_n, y), \\ 
    & r(x_1', x_2'), \dots,
    r(x_{n - 1}', x_n'), A(x_1')\ \}.  
  \end{align*}
  It is important to note 
  that $q^* \subseteq_\Omc q_n$ for all $n \geq 1$. Intuitively, 
  this makes it impossible for the learner to distinguish between $q^*$
  being the target query and one of the $q_n$ being the target 
  query. If, for example, the learner asks a membership query 
  `$\Amc^*,\Omc \models q_T(a)?$', where $\Amc^*$ is $q^*$ viewed as 
  an ABox, then the oracle will answer `yes' and the learner has not 
  gained any additional information.

  Now, the strategy of the oracle to answer a membership
  query `$\Amc,\Omc \models q_T(a)?$' is as follows:
  \begin{enumerate}

    \item if $\Amc,\Omc\models
      q^*(a)$, then reply ``yes''; 

    \item otherwise, reply ``no'' and remove
      from $\Hmc$ any $q$ that satisfies $\Amc,\Omc\models q(a)$. 

  \end{enumerate}
  An important aspect of this strategy is that, as proved below, only
  finitely many hypotheses~$q$ are removed whenever Case~2 above
  applies. Consequently, after any number of membership queries, 
  the set of remaining hypotheses \Hmc is infinite and
  contains $q^*$.
  The learner can then, however, not distinguish between $q^*$ and
  the remaining hypotheses and thus not identify the
  target query. In particular, 
  the presence of $q^*\in \Hmc$ prevents the learner from simply going 
  through all $q_i\in \Hmc$, asking membership queries with 
  ABoxes that take the form of these queries, and identifying $q_i$
  as the target query when the membership query for $q_i$ succeeds.

  \smallskip
  \noindent
  \textit{Claim.} Let \Amc an ABox and $a \in \mn{ind}(\Amc)$.  If
  $\Amc,\Omc \not \models q^*(a)$, then $\Amc,\Omc \models
  q_n(a)$ for only finitely many primes $n$.

  \smallskip\noindent \textit{Proof of the claim.}  Let \Amc an ABox
  and $a \in \mn{ind}(\Amc)$ such that $\Amc,\Omc \not \models
  q^*(a)$. Then \Amc satisfies
  $\mn{func}(r^-)$. Suppose to the contrary of what we have to show
  that there are infinitely many primes
  $n$ such that $\Amc,\Omc\models q_n(a)$ and let
  $h_n$ be the witnessing homomorphisms from $q_n$ to 
  $\Umc_{\Amc,\Omc}$ with \mbox{$h(x_1)=a$}.
  %
  \begin{itemize}

  \item If $h_n(x_n)\notin\mn{ind}(\Amc)$ for some prime $n$ with
    $\Amc,\Omc\models q_n(a)$, then $h_n(x_n')=h_n(x_n)$ due to the
    tree structure of the non-ABox part of $\Umc_{\Amc,\Omc}$. Since
    $r^-$ is functional, it follows that $h_n(x_j')=h_n(x_j')$ for all
    $j$ with $1\leq j\leq n$. Since $A(x_1')\in q_n$, also
    $A(h_n(x_1'))=A(h_n(x_1))=A(a)\in \Amc$.  Since also
    $A(x_0),r(x_0,x_1)\in q_n$, we have
    $A(h_n(x_0)), r(h_n(x_0),h_n(x_1))\in \Umc_{\Amc,\Omc}$, and thus
    $h_n$ is a homomorphism from $q^*$ to $\Umc_{\Amc,\Omc}$ with
    $h_n(x_1)=a$. Hence, $\Amc,\Omc\models q^*(a)$, a contradiction.

  \item Otherwise, $h_n(x_n)\in\mn{ind}(\Amc)$ for all primes $n$ with
    $\Amc,\Omc\models q_n(a)$.  Since $\Amc$ is finite, there is an
    element $b\in\mn{ind}(\Amc)$ such that $h_m(x_m)=b$ for infinitely
    many primes $m$.  Thus, there is an $r$-path of length $m$ from $a$
    to $b$ in $\Amc$ for infinitely many primes~$m$. Since $\Amc$ is
    finite and satisfies $\mn{func}(r^-)$, this is only
    possible if $a=b$, $r(a,a)\in \Amc$, and $h_m(x_j)=a$ for all
    considered $m$ and all $j$ with $1\leq j\leq m$.  Since also
    $A(x_0),r(x_0,x_1)\in q_n$ and $\Amc$ satisfies
    $\mn{func}(r^-)$, we further have $h_n(x_0)=a$ for all primes $n$
    with $\Amc,\Omc\models q_n(a)$ and $A(a)\in\Amc$. But then
    $\Amc,\Omc\models q^*(a)$, a contradiction.

  \end{itemize}
%
  %
\end{proof}

\end{document}